%% file: bounded.tex
\documentclass[preprint,3p,times,twocolumn]{elsarticle}
\usepackage{lineno}

\usepackage{amssymb}

\input{tex/macros.tex}

\journal{ }

\begin{document}

\begin{frontmatter}


\title{BoundED: Neural Boundary and Edge Detection in 3D Point Clouds via Local Neighborhood Statistics}
\input{tex/authors.tex}
\input{tex/abstract.tex}

\end{frontmatter}



\input{tex/introduction.tex}
\input{tex/related_work.tex}
\input{tex/methodology.tex}
\input{tex/results_and_discussion.tex}
\input{tex/conclusion.tex}
\input{tex/acknowledgements.tex}



\bibliographystyle{elsarticle-num-names}
\bibliography{refs}

\end{document}

%% file: tex/macros.tex
\usepackage{bm}
\usepackage{booktabs}
\usepackage{tikz}
\usetikzlibrary{positioning}
\usetikzlibrary{fit}
\usetikzlibrary{calc}
\PassOptionsToPackage{hyphens}{url}\usepackage{hyperref}

\usepackage{amsmath}

\newcommand{\dotprod}[2]{\langle#1,#2\rangle}

\usepackage{color}
\usepackage[normalem]{ulem}

%% file: tex/authors.tex
\author[aff:bonn]{Lukas Bode\corref{cor1}}
\ead{lbode@cs.uni-bonn.de}
\cortext[cor1]{Corresponding author}

\author[aff:delft]{Michael Weinmann}
\ead{M.Weinmann@tudelft.nl}

\author[aff:bonn]{Reinhard Klein}
\ead{rk@cs.uni-bonn.de}

\affiliation[aff:bonn]{organization={University of Bonn},
            addressline={Friedrich-Hirzebruch-Allee 8},
            city={Bonn},
            postcode={53115},
            country={Germany}}

\affiliation[aff:delft]{organization={Delft University of Technology},
            addressline={Van Mourik Broekmanweg 6},
            city={Delft},
            postcode={2628 XE},
            country={Netherlands}}



%% file: tex/abstract.tex
\begin{abstract}
    Extracting high-level structural information from 3D point clouds is challenging but essential for tasks like urban planning or autonomous driving requiring an advanced understanding of the scene at hand.
    Existing approaches are still not able to produce high-quality results consistently while being fast enough to be deployed in scenarios requiring interactivity.
    We propose to utilize a novel set of features describing the local neighborhood on a per-point basis via first and second order statistics as input for a simple and compact classification network to distinguish between non-edge, sharp-edge, and boundary points in the given data.
    Leveraging this feature embedding enables our algorithm to outperform the state-of-the-art techniques in terms of quality and processing time.
\end{abstract}


%% file: tex/introduction.tex
\begin{figure*}[!ht]
    \centering
    \begin{tikzpicture}[image/.style = {inner sep=0mm, outer sep=0pt},
                        label_top/.style = {anchor=south},
                        label_side/.style = {rotate=90, anchor=south, align=center},
                        node distance = 2mm and 0mm]
        \node[image]                 (teaser)  {\includegraphics[width=1.0\textwidth]{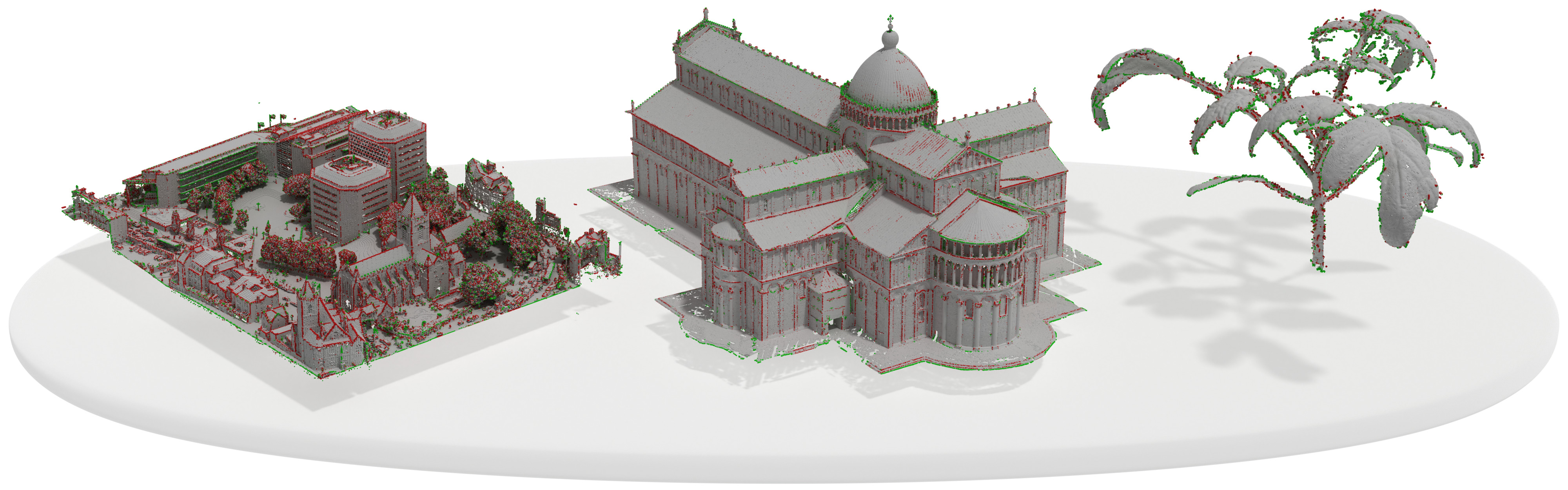}};
    \end{tikzpicture}
    \caption{\label{fig:teaser2}
        Our BoundED approach extracts sharp edges and boundaries from 3D point cloud data purely based on positional data.
        Points classified as sharp-edge are highlighted in red while boundary points are highlighted in green.
    }
\end{figure*}

\section{Introduction}
\label{sec:introduction}

3D point cloud data obtained from terrestrial or airborne laser scanning as well as depth sensors and image-based structure-from-motion have become the prerequisite for numerous applications including geographic information systems, urban planning, indoor modeling for the built environment, autonomous driving, and navigation systems.
However, the sampling of scenes with arbitrary complexity in terms of unstructured data complicates the further processing of the data as e.g. required when extracting characteristic features for navigation or scene interpretation according to object instances and materials.
Edges represent characteristic features that often occur at object borders as well as on surfaces (in the form of ridges or engravings) and linear scene structures like scaffolds and, hence, provide essential information regarding the underlying geometric structures.
However, automatic edge detection in 3D point cloud data remains a challenging task.
Whereas physical edges may not appear as sharp due to damage or cleaning (e.g. stone or plastered buildings, progressively smoothed edges, polished mechanical parts, etc.), there are also limitations inherent to the scanning approaches, especially due to the typically uneven, noisy sampling of the scene, that may result in a slight rounding effect of edges in the reconstruction.
Furthermore, the sharpness, smoothness or roundness of edges also depends on the observation scale.
Therefore, there might be some ambiguity in defining edges, that may require involving further context information.
In addition, with point clouds typically consisting of tens or hundreds of millions of points, efficient operators are required.
Advances in machine learning and the rapidly growing availability of 3D data have led to several supervised learning approaches for concept classification.
Respective approaches include the classification of structures according to semantic categories such as facades, roofs, different forms of vegetation or pole/trunk structures using pointwise hand-crafted geometric descriptors on a single \emph{optimal} scale~\cite{demantke2011dimensionality,weinmann2015semantic,weinmann2015distinctive,hackel2016fast} or multiple scales~\cite{brodu20123d,blomley2017using}, additionally leveraging contextual information~\cite{niemeyer2014contextual,weinmann2015contextual,steinsiek2017semantische,landrieu2017structured}, as well as deep-learning strategies~\cite{guo2020deep,xie2020linking,huang2016point,qi2017pointnet++,boulch2017unstructured,tchapmi2017segcloud,hackel2017isprs,lawin2017deep,landrieu2018large,thomas2019kpconv,li2022vd,mao2022beyond}.
Furthermore, a few works also focused on the individual classification of points according to being or not being on edges based on multi-scale features and a random-forest-based classification~\cite{hackel2016contour}, multi-scale features and a dedicated neural network based edge detection classifier~\cite{himeur2021pcednet}, neural-network-based pointwise distance estimation to the next sharp geometric feature~\cite{matveev2022def}, binary-pattern-based filtering on local topology graphs~\cite{guo2022sglbp}, neural-network-based edge-aware point set consolidation leveraging an edge-aware loss~\cite{yu2018ec}, training two networks based on PointNet++~\cite{qi2017pointnet++} to classify points into corners and edges and subsequently applying non-maximal suppression and inferring feature curves~\cite{wang2020pie}, the learning of multi-scale local shape properties (e.g., normal and curvature)~\cite{guerrero2018pcpnet}, and the computation of a scalar \emph{sharpness} field defined on the underlying Moving Least-Squares surface of the point cloud whose local maxima correspond to sharp edges~\cite{raina2018mls2,raina2019}.
However, extracting high-quality edge and boundary data from a large variety of different 3D point clouds fast enough to eventually be suitable for usage in embedded systems or real-time settings remains an open problem.
In this paper, inspired by the \emph{maximum mean discrepancy} (MMD) operator~\cite{gretton2012kernel} which allows to compare distributions by embedding them in a feature space and comparing the mean of the respective embeddings, we propose to tackle the point classification task by training a network to distinguish between classes based on a feature embedding related to the first and second order statistics of the respective point's neighborhood.
This embedding contains enough information for the classification network to learn the difference between non-edge, sharp-edge, and boundary points while at the same time being well structured and compact, making our solution very fast in terms of processing time.
Various results of our \emph{Boundary and Edge Detection} (BoundED) approach are depicted in Figure~\ref{fig:teaser2}.
Our main contributions can be summarized as follows:
\begin{itemize}
    \item We present a novel set of features for edge and boundary characterization and detection capturing local neighborhood information of point clouds better and being cheaper to compute than state-of-the-art approaches~\cite{himeur2021pcednet}.
    \item We demonstrate the benefits of this novel feature embedding at the example of a modified state-of-the-art neural edge detection network architecture giving better results with an even smaller network.
    \item Our evaluation demonstrates the ability of the proposed features to capture information regarding boundary classification of points in addition to edge classification.
\end{itemize}

%% file: tex/related_work.tex
\section{Related Work}
\label{sec:related_work}

The detection of 3D edges in terms of sharp features, feature contours, or curves within unstructured point cloud data is a challenging task.
Conventional methods include surface mesh reconstruction or graph-based approaches and analyzing local neighborhoods of each individual point based on principal component analysis (PCA).
Thereby, the given connectivity information of a point with respect to its neighbors allows for a faster nearest neighbor search in comparison to unstructured point sets.
However, preserving sharp edges and complex features in a reconstructed model is challenging due to smoothing effects induced by several reconstruction techniques.
Directly extracting edges from unstructured point clouds has been addressed based on computing geometric descriptors per point based on the local covariance characteristics~\cite{gumhold2001feature,gelfand2004shape}.
Respective variants include taking the ratio between the Eigenvalues of the local covariance matrices on a single scale~\cite{merigot2011voronoi,xia2017fast} or different scales~\cite{pauly2003multi,bazazian2015fast}, local slippage analysis to define edges between segments of rotationally and translationally symmetrical shapes such as planes, spheres, and cylinders~\cite{gelfand2004shape}, or directly estimating curvature~\cite{lin2015line,nguyen2018edge}.
Considering multiple scales reduces the susceptibility to noise, but such methods still rely on the suitable specification of a decision threshold.
Non-parametric edge extraction has been achieved via kernel regression~\cite{oztireli2009feature} or Eigenvalue analysis~\cite{bazazian2015fast}.
Others focused on detecting depth-discontinuities based on finding triangles with oblique orientations or finding triangles with long edges~\cite{tang2007comparative} or focusing on high-curvature points given as the extremum of curvatures~\cite{fan1987segmented} or curvature-guided region growing~\cite{rusu2008towards}.
In addition, edge detection has been approached based on normal variation analysis~\cite{che2018multi}, 3D Canny edge detection~\cite{monga19913d}, the combination of normal estimation and graph theory~\cite{yague20133d}, alpha-shapes~\cite{edelsbrunner1994three}, or boundary detection via DBSCAN-based detection and segmentation of 3D planes~\cite{CHEN2022108431}.
Further approaches followed a moving least-squares (MLS) surface reconstruction with the subsequent detection of 3D edges based on a Gaussian map clustering computed within a local neighborhood \cite{demarsin2007detection,weber2010sharp,weber2012sharp,ni2016edge}.
The consideration of higher-order local approximations of non-oriented input gradients in MLS-based reconstruction has been used for the computation of continuous non-oriented gradient fields~\cite{chen2013non}, which allows a better preservation of surface or image structures.
Another possibility to achieve continuously differentiable surfaces consists in exploring the scale-space for MLS~\cite{mellado2012growing}.
Furthermore, a scalar \emph{sharpness} field defined on the underlying Moving Least-Squares surface of the point cloud has been proposed, where local maxima correspond to sharp edges~\cite{raina2018mls2,raina2019}. 
Other approaches include the combination of adaptive reconstruction kernels~\cite{fleishman2005robust} and spline fitting~\cite{daniels2008spline}, the detection of boundary points and internal points as well as the subsequent application of a Fast-Fourier-Transform-based edge reconstruction to avoid the need to define a specific order for polynomial curve fitting~\cite{mineo2019novel}, the use of subspace detection and feature intersection~\cite{fernandes2012general}, mean-shift-based selection of the most distant points with respect to the centroid of their neighborhood~\cite{ahmed2018edge}, the use of locally defined curve set features~\cite{li2017curve}, the intersection of automatically detected planes~\cite{mitropoulou2019automated}, the filtering of potential feature points according to their local topology graph based on binary patterns~\cite{guo2022sglbp}, or RANSAC-based spatial regularization of sharp feature detector responses~\cite{lin2015line}.
In addition, gradient-based edge detection with a subsequent non-maxima suppression and edge linking into linear and smooth structures~\cite{xia2017fast} has been investigated.

Along the rapid progress in machine learning, learning-based approaches have been proposed for classifying individual points as \emph{edge} or \emph{non-edge}.
Besides approaches based on least square regression or support vector machines~\cite{wang2019development} that, however, had not been investigated in a general scenario, this can be achieved by the use of multi-scale features with a random forest based edge classification~\cite{hackel2016contour} or neural network based edge classifier~\cite{himeur2021pcednet}.
Other approaches include the neural network based pointwise distance estimation to the next sharp geometric feature~\cite{matveev2022def}, or neural network based edge-aware point set consolidation~\cite{yu2018ec} and 3D semantic edge detection based on a two-stream fully-convolutional network to jointly perform edge detection and semantic segmentation~\cite{hu2020jsenet}.
A further method~\cite{wang2020pie} trains two neural networks to classify points into corners and edges based on a PointNet++ like architecture~\cite{qi2017pointnet++}.
After a subsequent non-maxima suppression of the classified points and their PointNet++ based clustering, a two-headed PointNet~\cite{qi2017pointnet} generates the final set of curves.
This concatenation of deep networks induces a high computational burden and relies on high resource requirements.
In addition, the learning of multi-scale local shape properties (e.g., normal and curvature)~\cite{guerrero2018pcpnet} and the use of CNNs for adaptive feature extraction from observations in a camera and laser-scanner setup~\cite{XIAO2019111533} have been investigated.
Furthermore, the prediction of part boundaries within a 3D point cloud based on a graph convolutional network has been proposed~\cite{loizou2020learning}.
Further purely on boundary detection focused methods include the initial extraction of the exterior boundary based on neighborhood characteristics and the subsequent analysis regarding whether a point belongs to a hole boundary~\cite{trinh2015hole}, and approaches based on a deep neural network~\cite{tabib2020learning}.

There are also a few image-based approaches that initially convert the 3D point cloud data into images~\cite{lin2015line}.
Subsequently, a line segment detector~\cite{von2008lsd} is used to extract lines in 2D, which are backprojected to the point cloud.
Another approach~\cite{lu2019fast} relies on an initial segmentation of the point cloud into planar regions based on region growing and merging, which is followed by a plane-wise point projection into a 2D image and a final 2D contour extraction and backprojection to get the respective line segment in 3D space.

\begin{figure*}[t]
  \centering
  \includegraphics[width=\linewidth]{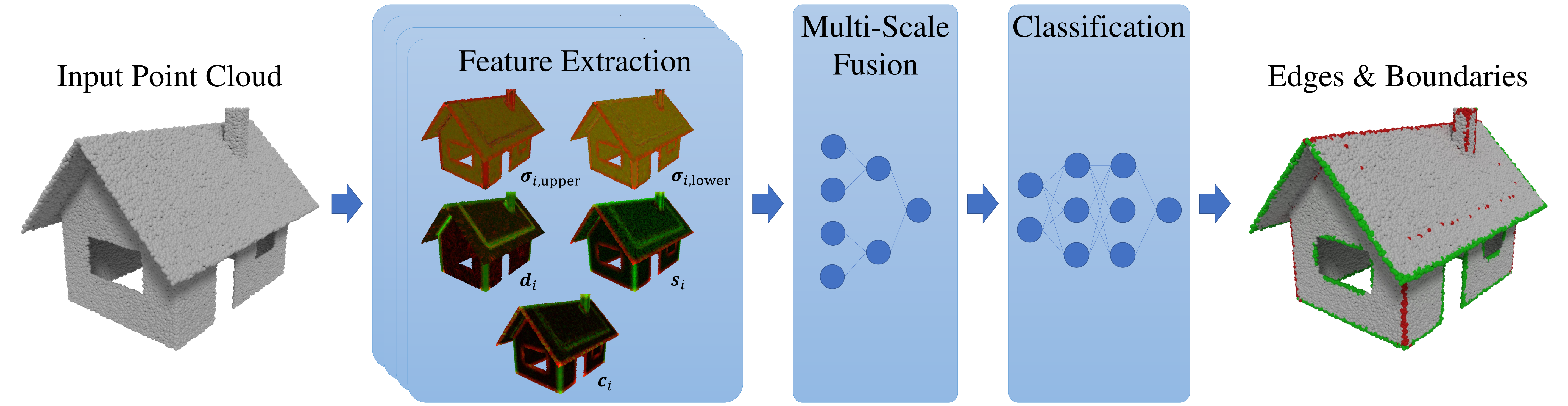}
  \caption{\label{fig:teaser}
      Overview of our BoundED approach:
      Based on an input point cloud, several features describing the local geometry are extracted on multiple scales.
      After pairwise fusion of the features for different scales, we classify the input points by an MLP leveraging the fused features as either non-edge, sharp-edge, or boundary points.
  }
\end{figure*}

With our approach we follow the avenue of neural network based edge and boundary detection within 3D point clouds. We take inspiration from the \emph{maximum mean discrepancy} (MMD) operator~\cite{gretton2012kernel} for the definition of a local feature embedding with respect to first- and second-order statistics of a local point's neighborhood, and we show that this embedding allows robust detection of non-edge, sharp-edge, and boundary points already with a compact network, thereby enabling fast inference times.

%% file: tex/methodology.tex
\section{Methodology}
\label{sec:methodology}

With our approach, that we denote as BoundED, we aim at the robust and fast detection of non-edge, sharp-edge, or boundary points within given point clouds.
For this purpose, we leverage the combination of a local encoding of feature characteristics based on the maximum mean discrepancy operator with respect to the local first- and second-order statistics and their efficient classification based on a compact multi-layer perceptron (MLP) (see Figure~\ref{fig:teaser}).
In the following sections, we provide detailed descriptions regarding these aspects as well as respective implementation details.

\begin{figure*}[t]
    \centering
    \includegraphics[width=\linewidth]{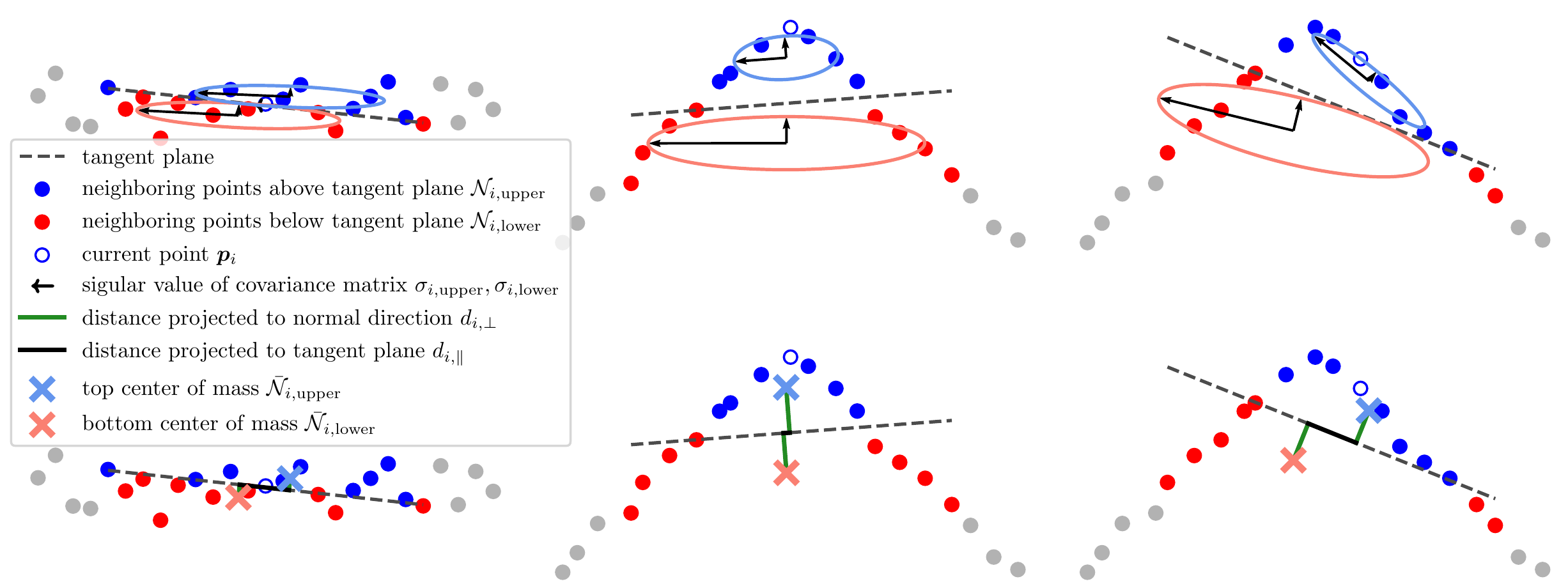}
    \caption{\label{fig:subsets}
        Features extracted from the local neighborhood of a point.
        Points can be classified as non-edge or sharp-edge by analyzing their neighborhood with respect to singular values and means of points above and below a least-squares fitted tangent plane.
        Planar neighborhoods (left) tend to have similar values for $\sigma_{i,\mathrm{upper}}$ and $\sigma_{i,\mathrm{lower}}$ while having low values for $d_{i,\perp}$.
        Sharp-edge neighborhoods (middle) exhibit a larger difference in $\sigma_{i,\mathrm{upper}}$ and $\sigma_{i,\mathrm{lower}}$ as well as large $d_{i,\perp}$.
        In contrast, neighborhoods of points close to sharp-edges (right) have higher $d_{i,\parallel}$ than neighborhoods of points directly on the edge.
    }
\end{figure*}

\subsection{Feature Computation}\label{sec:features}

To compute meaningful features as input for the consecutive neural classification step, we generalize the idea of dividing a set of 3D points into two disjoint subsets and analyzing their respective covariances introduced by \citet{bode2022locally-guided} in the context of image denoising.
%

%
Let $\mathcal{P} = \{\bm{p}_i\}$ for $i=1, \ldots, n$ be the given 3D point cloud consisting of $n$ points.
Using the $k$-nearest neighbors ($k$-NN) operator $\mathrm{NN}_k(\bm{p}, \mathcal{P})$, we extract local neighborhoods $\mathcal{N}_{i,k} = \mathrm{NN}_k(\bm{p}_i, \mathcal{P})$ with $k$ points each.
Throughout the remainder of this section, the neighborhood size $k$ is omitted for notational simplicity.
%

%
For sufficiently dense point clouds in the absence of noise, this neighborhood represents a roughly disc-shaped set of points.
In order to be invariant to the scale and sampling of the given point cloud, the point sets are normalized individually before features can be extracted.
We propose to utilize the covariance matrix $\bm{K}_i = \mathrm{cov}(\mathcal{N}_i)$ for this purpose.
By conducting an SVD of the covariance matrix
\begin{equation}
    \bm{K}_i = \bm{U}_i \bm{\Sigma}_i \bm{V}^T_i
\end{equation}
singular values $\sigma_{i,j} = \bm{\Sigma}_{i,jj}$ can be read from the diagonal entries of the matrix $\bm{\Sigma}_i$.
Without loss of generality, these sigular values are assumed to be sorted in descending order, i.e. $\sigma_{i,1} \geq \sigma_{i,2} \geq \sigma_{i,3}$.
Intuitively, these singular values are directional variances with directions being given by the corresponding Eigenvectors.
Since $\mathcal{N}_i$ is roughly disk shaped, $\sigma_{i,1}$ and $\sigma_{i,2}$ can be seen as variance in direction of the disk's perpendicular semiaxes.
Note, that in general $\sigma_{i,1}$ and $\sigma_{i,2}$ are similar but not equal as the points $\mathcal{N}_i$ will never represent a perfect uniformly sampled disk in practice.
For the purpose of normalization, the neighborhood is centered around the origin according to the neighborhood's center of mass
\begin{align}
    \bar{\mathcal{N}}_{i} &= \frac{1}{|\mathcal{N}_{i}|}\sum_{\bm{p} \in \mathcal{N}_{i}} \bm{p}
\end{align}
and scaled by the average standard deviation along the semiaxes:
\begin{equation}
    \hat{\mathcal{N}}_i = \left\{\frac{2}{\sqrt{\sigma_{i,1}} + \sqrt{\sigma_{i,2}}}\,(\bm{p} - \bar{\mathcal{N}}_{i}) \mid \bm{p} \in \mathcal{N}_i\right\}.
\end{equation}
%

%
Besides normalization of the neighborhood, this SVD and in particular the Eigenvector $\bm{n}_i$ corresponding to $\sigma_{i,3}$ is utilized for further processing as this vector together with the neighborhood's center of mass $\bar{\mathcal{N}}_{i}$ defines a least-squares fitted plane to $\mathcal{N}_i$.
Note that, in contrast to other approaches like e.g. PCEDNet~\citep{himeur2021pcednet}, by using this Eigenvector $\bm{n}_i$ as normal, our BoundED does not rely on any precomputed normals but only on the 3D positions of the points.
We have observed, that the orientation of $\bm{n}_i$ can be unstable near outliers.
Thus, we only consider the $\lfloor k/2 \rfloor$ points closest to $\bar{\mathcal{N}}_i$ for this step.
According to this plane, the neighborhood is partitioned into two disjoint subsets
\begin{align}
    \mathcal{N}_{i,\mathrm{upper}} &= \left\{\bm{p} \in \hat{\mathcal{N}}_i \mid \dotprod{\bm{p}}{\bm{n}_i} \geq 0 \right\} \\
    \mathcal{N}_{i,\mathrm{lower}} &= \left\{\bm{p} \in \hat{\mathcal{N}}_i \mid \dotprod{\bm{p}}{\bm{n}_i} < 0 \right\}.
\end{align}
%

%
As depicted in Figure~\ref{fig:subsets}, an analysis of these provides valuable information regarding local geometry.
We propose to analyze the subset's statistics to capture this information.
In particular, singular values $\sigma_{i,\mathrm{upper},j}$ and $\sigma_{i,\mathrm{lower},j}$ for $j \in \{1, 2, 3\}$ are computed by means of individual SVDs of the covariance matrices of $\mathcal{N}_{i,\mathrm{upper}}$ and $\mathcal{N}_{i,\mathrm{lower}}$ respectively.
Additionally, the distance between the centers of mass of both subsets 
\begin{align}
    \bar{\mathcal{N}}_{i,\mathrm{upper}} &= \frac{1}{|\mathcal{N}_{i,\mathrm{upper}}|}\sum_{\bm{p} \in \mathcal{N}_{i,\mathrm{upper}}} \bm{p} \\
    \bar{\mathcal{N}}_{i,\mathrm{lower}} &= \frac{1}{|\mathcal{N}_{i,\mathrm{lower}}|}\sum_{\bm{p} \in \mathcal{N}_{i,\mathrm{lower}}} \bm{p}
\end{align}
decomposed into perpendicular and tangential components is calculated as
\begin{align}
    d_{i,\perp} &= \dotprod{\bar{\mathcal{N}}_{i,\mathrm{upper}} - \bar{\mathcal{N}}_{i,\mathrm{lower}}}{\bm{n}_i} \\
    d_{i,\parallel} &= \lVert (\bar{\mathcal{N}}_{i,\mathrm{upper}} - \bar{\mathcal{N}}_{i,\mathrm{lower}}) - d_\perp \bm{n}_i \rVert_2.
\end{align}
Intuitively, low values for $d_{i,\perp}$ indicate that the local neighborhood $\mathcal{N}_i$ is near planar and thus the probability for $\bm{p}_i$ being part of a sharp edge is small.
In contrast, high values are found in areas with a high amount of geometric detail or noise.
A large tangential distance $d_{i,\parallel}$ can indicate, that an edge is close-by, but $\bm{p}_i$ may not necessarily be coincident (see Figure~\ref{fig:subsets}).
Furthermore, inspired by~\citet{bendels2006detecting}, to improve detection of outliers and boundaries, the perpendicular and tangential components of the distance between $\bm{p}_i$ and the center of mass of its $k$ nearest neighbors are computed as:
\begin{align}
    s_{i,\perp} &= \dotprod{\bm{p}_i - \bar{\mathcal{N}}_i}{\bm{n}_i} \\
    s_{i,\parallel} &= \lVert (\bm{p}_i - \bar{\mathcal{N}}_i) - s_\perp \bm{n}_i \rVert_2.
\end{align}
While not necessarily always following this observation, points at boundaries tend to have large $s_{i,\parallel}$ and at the same time small $s_{i,\perp}$.
Intuitively, the neighbors of points at boundaries are all on one side which indicates that they are far away from the center of mass of their neighborhood.
If $\bm{p}_i$ is an outlier near a well-defined surface, the corresponding $s_{i,\perp}$ tends to be large.

In summary, the analysis yields the following features:
the singular values $\bm{\sigma}_{i,\cdot} = (\sigma_{i,\cdot,1}, \sigma_{i,\cdot,2}, \sigma_{i,\cdot,3})^T$ of the upper and lower subsets respectively, the perpendicular and tangential distances between the centers of mass of both subsets $\bm{d}_i = (d_{i,\perp}, d_{i,\parallel})^T$, and the perpendicular and tangential distances between the point $\bm{p}_i$ and the center of mass of its neighborhood $\bm{s}_i = (s_{i,\perp}, s_{i,\parallel})^T$.
Thus, we assemble a per-point 10D feature vector according to
\begin{equation}
    \hat{\bm{x}}_i = (\bm{\sigma}_{i,\mathrm{upper}} ,\bm{\sigma}_{i,\mathrm{lower}}, \bm{d}_i, \bm{s}_i)^T.
\end{equation}

\begin{figure}[t]
    \centering
    \includegraphics[width=\linewidth]{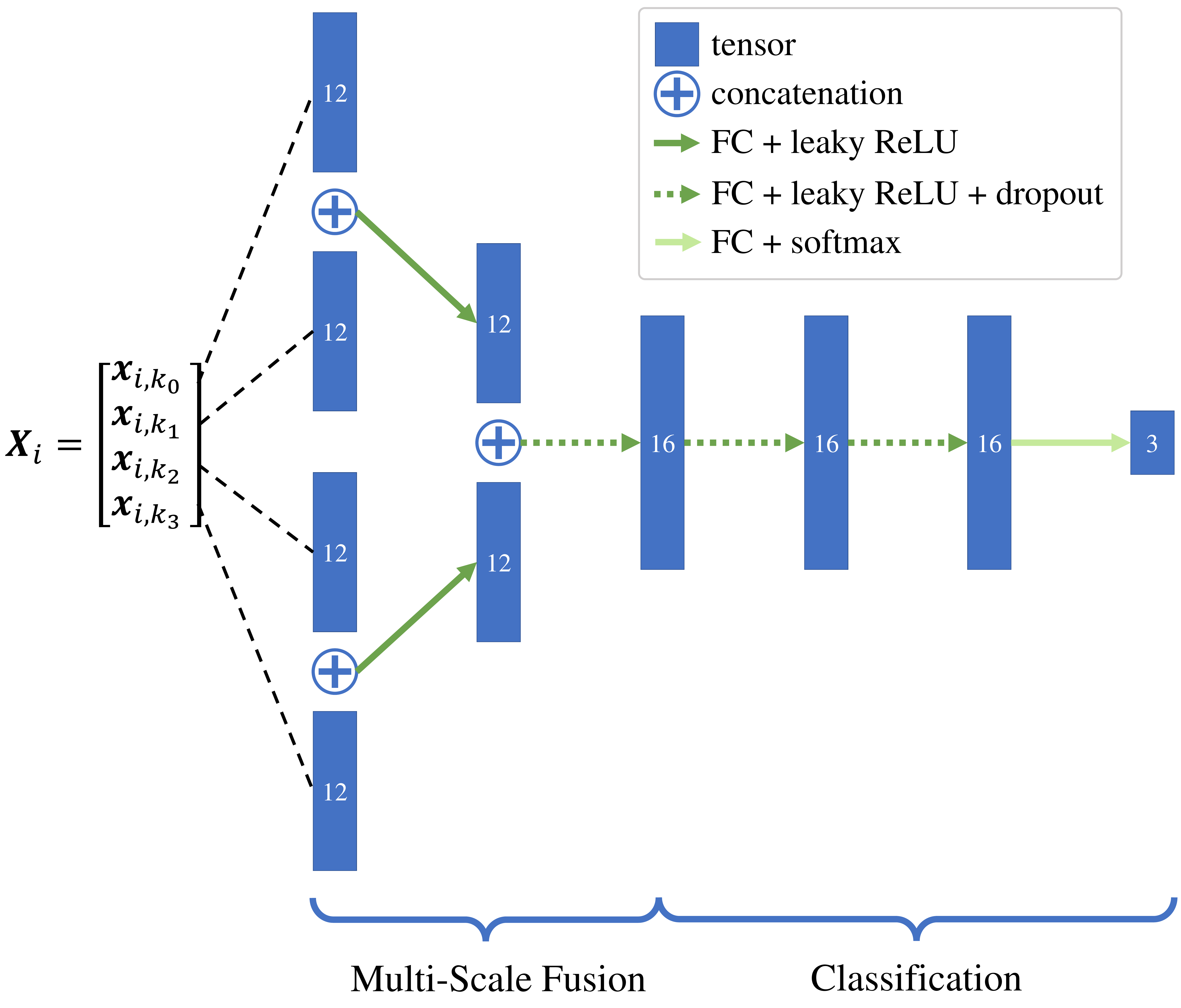}
    \caption{\label{fig:network_architecture}
        Architecture of the multi-scale fusion and classification network consisting of fully connected (FC) layers, leaky rectified linear unit (leaky ReLU) activations, dropout, and softmax function.
        Features computed on multiple scales are combined in a pairwise manner and afterwards processed by an MLP to classify a point as non-edge, sharp-edge, or boundary.
    }
\end{figure}

\subsection{Multi-Scale Feature Embedding}\label{sec:network}

In order to classify points $\bm{p}_i$ of a point cloud $\mathcal{P}$ as non-edge, sharp-edge, or boundary, the per-point data $\bm{X}_i$ is individually processed by a small MLP.
$\bm{X}_i$ relies on computing $\hat{\bm{x}}_{i,k}$ on $m$ different scales $k_0$, \ldots, $k_{(m-1)}$, i.e. choosing neighborhoods containing varying numbers of points $k$, for each point $\bm{p}_i$.
Inspired by the GLS~\citep{mellado2012growing} features utilized by PCEDNet~\citep{himeur2021pcednet}, we add the tangential and perpendicular distances
\begin{align}
    c_{i,k,\perp} &= \dotprod{\bar{\mathcal{N}}_{i,k} - \overline{(\mathcal{N}_{i,k_0} - \mathcal{N}_{i,k})}}{\bm{n}_{i,k_0}} \\
    c_{i,k,\parallel} &= \lVert (\bar{\mathcal{N}}_{i,k} - \overline{(\mathcal{N}_{i,k_0} - \mathcal{N}_{i,k})}) - c_{i,k,\perp} \bm{n}_{i,k_0} \rVert_2.
\end{align}
between the center of mass $\bar{\mathcal{N}_{i,k}}$ of each scale $k$ and the center of mass of points of the largest scale's neighborhood $\mathcal{N}_{i,k_0}$ which are not part of $\mathcal{N}_{i,k}$ as well to each $\hat{\bm{x}}_{i,k}$:
\begin{equation}
    \bm{x}_{i,k} = (\hat{\bm{x}}_{i,k}, \bm{c}_{i,k})^T,
\end{equation}
where $\bm{c}_{i,k} = (c_{i,k,\perp}, c_{i,k,\parallel})^T$.
The complete multi-scale per-point features can be written in matrix form as
\begin{align}
    \bm{X}_i =
    \begin{bmatrix}
        \bm{x}_{i,k_0,1} & \ldots & \bm{x}_{i,k_0,12} \\
        \vdots           & \ddots & \vdots           \\
        \bm{x}_{i,k_{(m-1)},1} & \ldots & \bm{x}_{i,k_{(m-1)},12}
    \end{bmatrix}.
\end{align}
These multi-scale features are fused in a pair-wise manner similarly to PCEDNet~\citep{himeur2021pcednet} as depicted in Figure~\ref{fig:network_architecture}, before being processed by the classification MLP itself.

\subsection{Network Architecture}

For our experiments, we use features computed on four different scales using 128, 64, 32, and 16 neighboring points respectively.
In contrast to PCEDNet, BoundED uses less scales, i.e. 4 instead of 16.
However, to accomodate for the lost network depth due to using less scales, an additional hidden layer is added to the classification MLP, giving the network a total of 1.6k learnable parameters.
For training the network, a focal loss~\citep{lin2017focal} with $\gamma = 2$ is used as training batches are usually very unbalanced due to the small number of edge points compared to non-edge points in most point clouds.
Furthermore, we propose to add dropout~\citep{srivastava2014dropout} with $p=0.5$ to the classification layers to prevent overfitting and facilitate a more stable training process.

\subsection{Implementation Details}
Our algorithm is implemented using PyTorch~\citep{pytorch} for feature extraction as well as the neural network and its training.
For finding the local neighborhood of points, the k-NN implementation of PyTorch3D~\cite{ravi2020pytorch3d} is used.
Due to the point sets $\mathcal{N}_{i,k,\mathrm{upper}}$, $\mathcal{N}_{i,k,\mathrm{lower}}$ containing different numbers of points for different $\bm{p}_i$, we employ masking to efficiently vectorize the task and fully utilize the tremendous computation capabilities of modern GPUs during the feature extraction phase.
The network is trained using the Adam optimizer~\citep{kingma2014adam} with $\beta_1 = 0.9$, $\beta_2 = 0.999$, and learning rate 0.001.
Batch size is set to 16384.
The number of training iterations varies between used datasets and is described in detail in Section~\ref{sec:datasets}.

%% file: tex/results_and_discussion.tex
\section{Results and Discussion}
\label{sec:results_and_discussion}

In the following, the effectiveness of the proposed combination of our novel multi-scale features and our compact classification network is evaluated quantitatively as well as qualitatively on several different datasets.
We focus mostly on the comparison with the state-of-the-art point cloud edge detection network PCEDNet by \citet{himeur2021pcednet} as it is the most relevant previous work due to also being designed to be fast and compact.
Similar to our BoundED approach, they rely on feeding their classification network with multi-scale per-point features allowing for a direct comparison of the used embeddings.
Furthermore, boundary detection capabilities of our network are assessed.
Finally, an experiment on noisy data as well as an ablation study regarding the chosen features and the employed number of scales further validate our results.

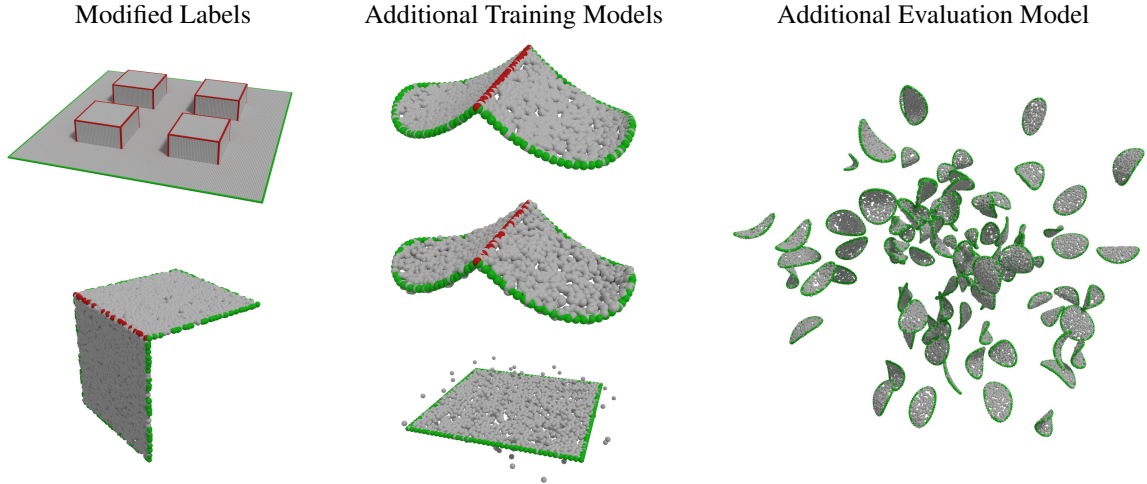
\begin{figure*}[t]
    \centering
    \begin{tikzpicture}[image/.style = {inner sep=0mm, outer sep=0pt},
                        label_top/.style = {anchor=north},
                        label_side/.style = {rotate=90, anchor=south, align=center},
                        node distance = 2mm and 10mm]
        \node[image]                 (left1)                                                              {\includegraphics[trim={0 0 0 100px},clip,width=0.3\linewidth]{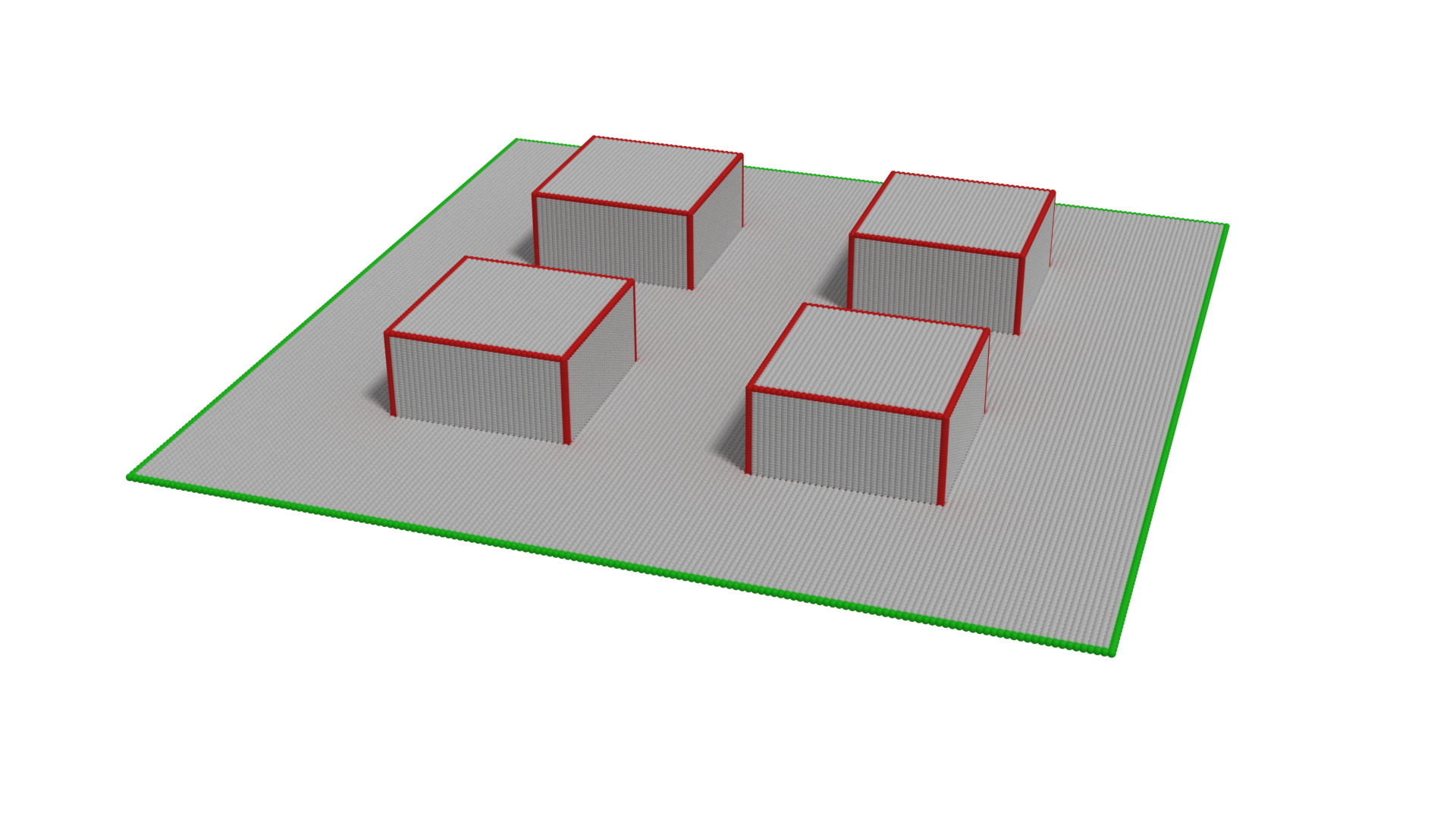}};
        \node[image, below=of left1] (left2)                                                              {\includegraphics[width=0.3\linewidth]{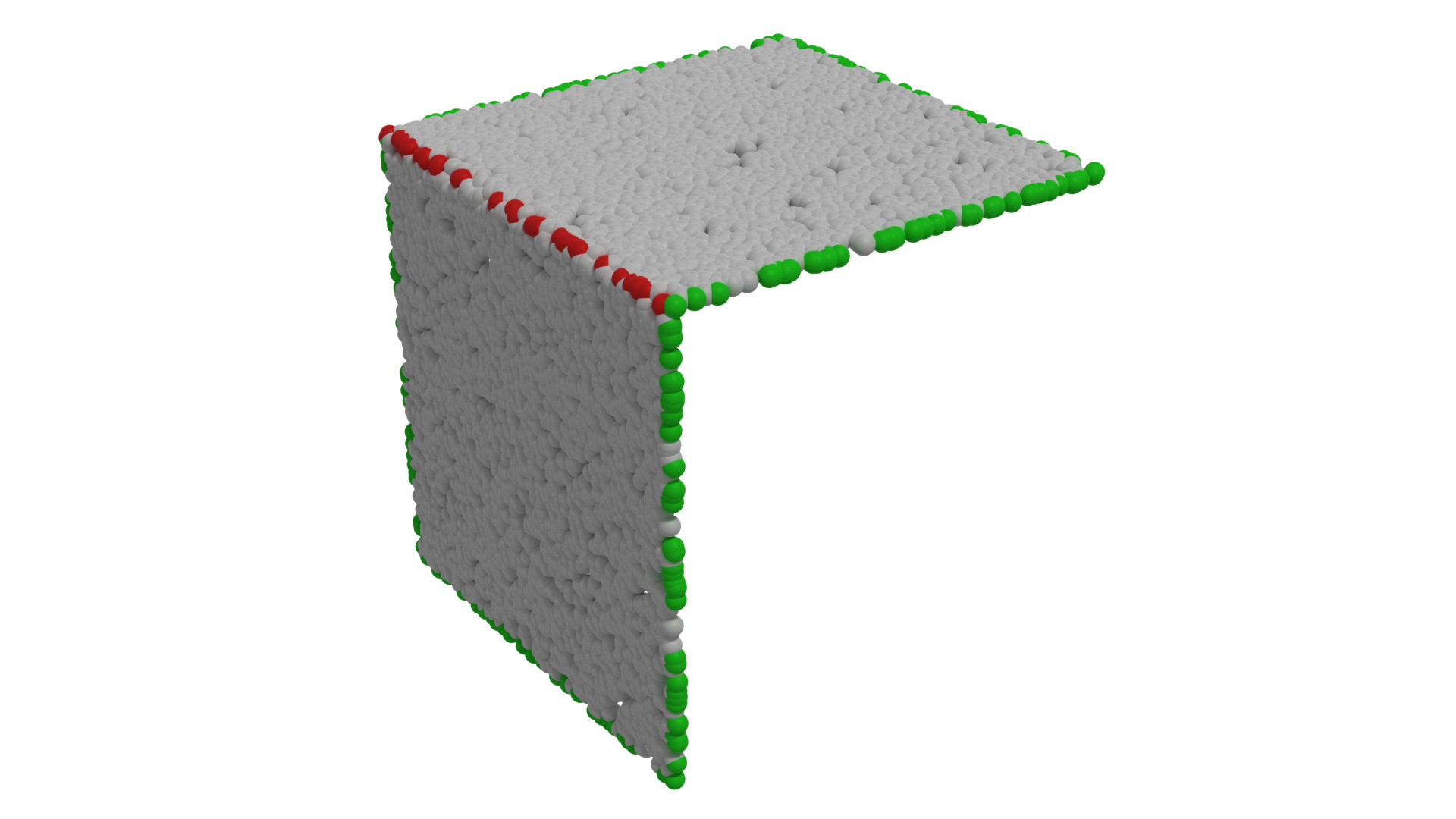}};
        \node[image, anchor=west]    (mid2)   at ($(left1.north east)!0.5!(left2.south east) + (5mm, 0)$) {\includegraphics[width=0.2\linewidth]{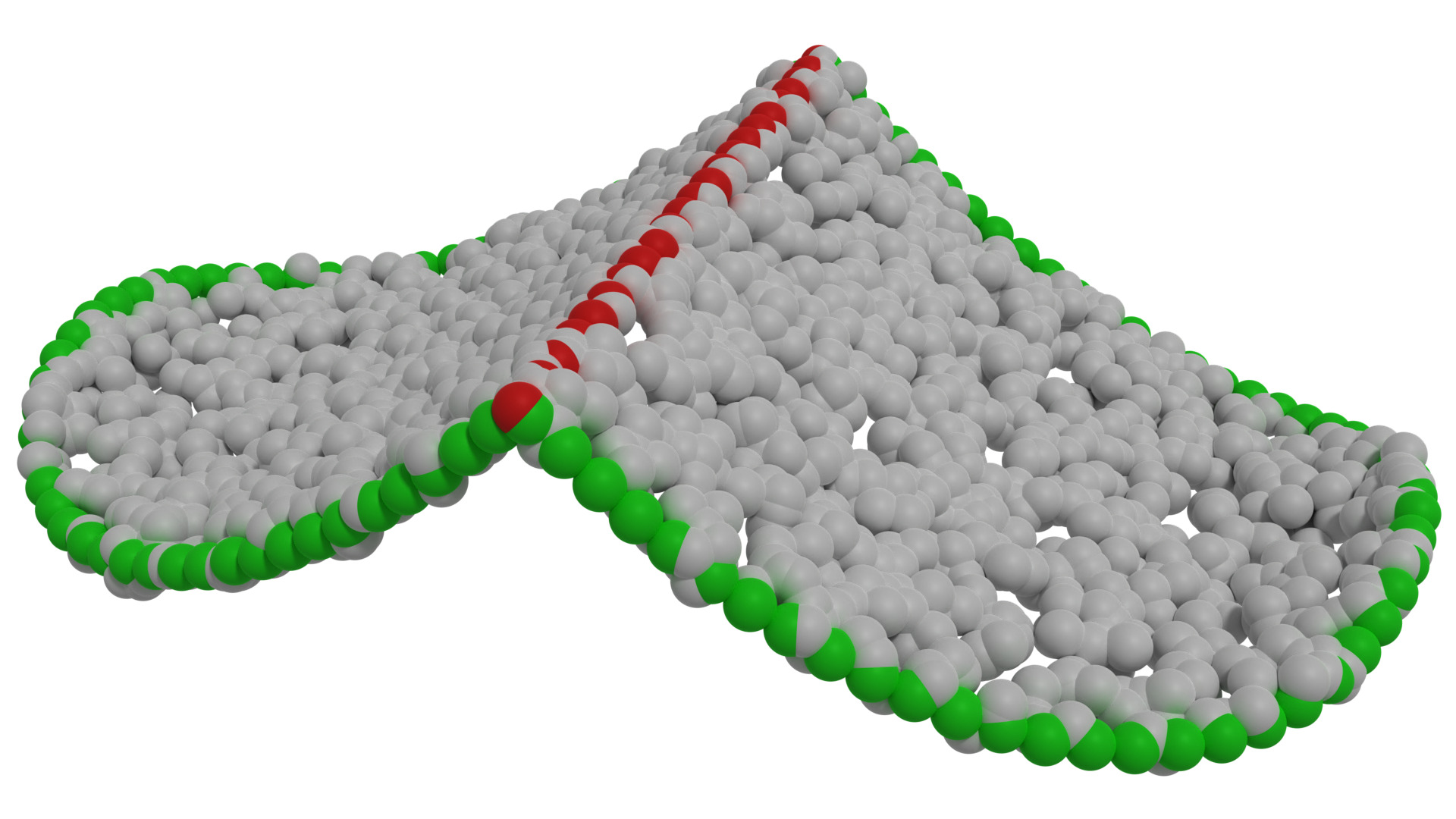}};
        \node[image, above=of mid2]  (mid1)                                                               {\includegraphics[width=0.2\linewidth]{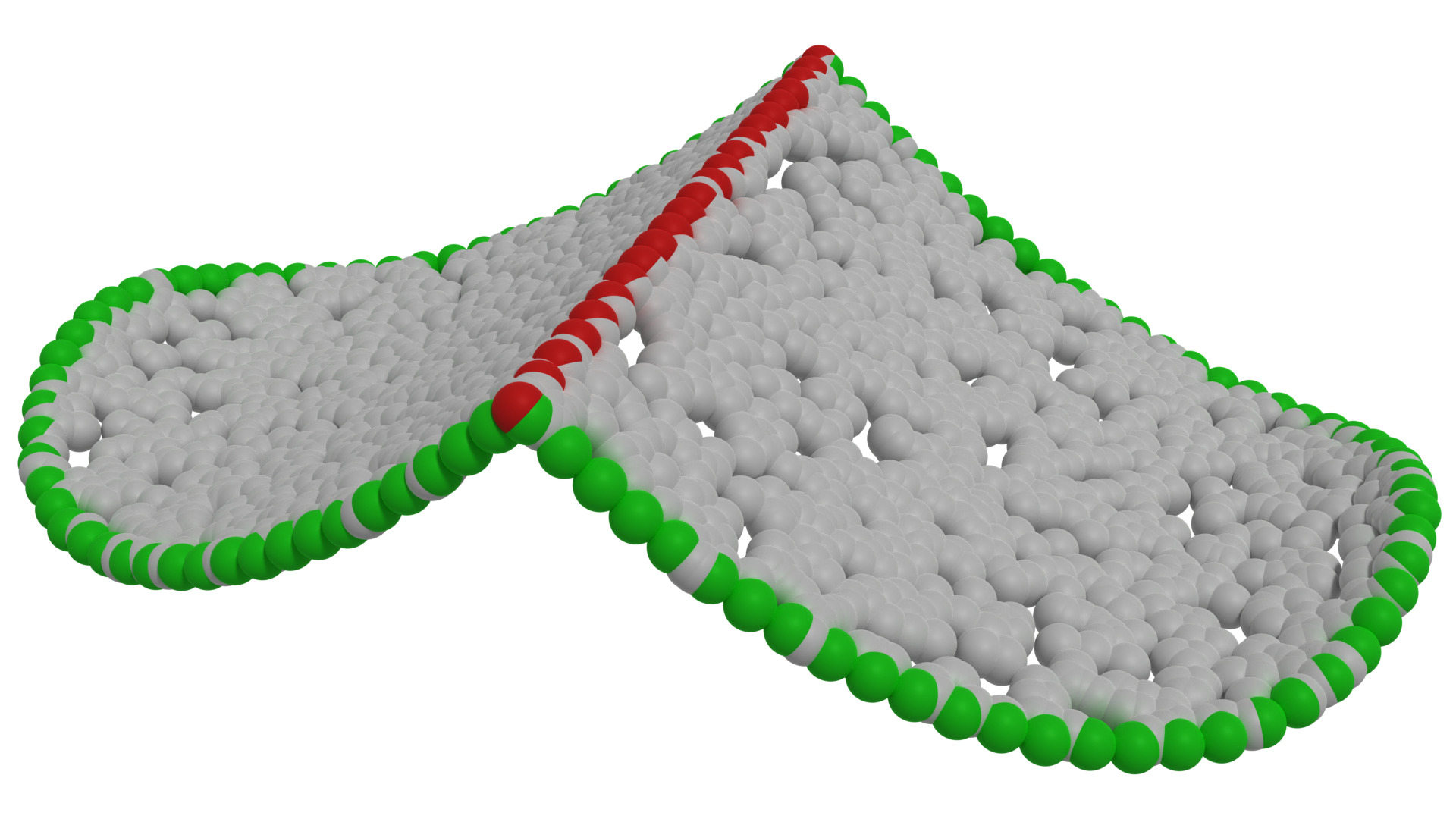}};
        \node[image, below=of mid2]  (mid3)                                                               {\includegraphics[width=0.2\linewidth]{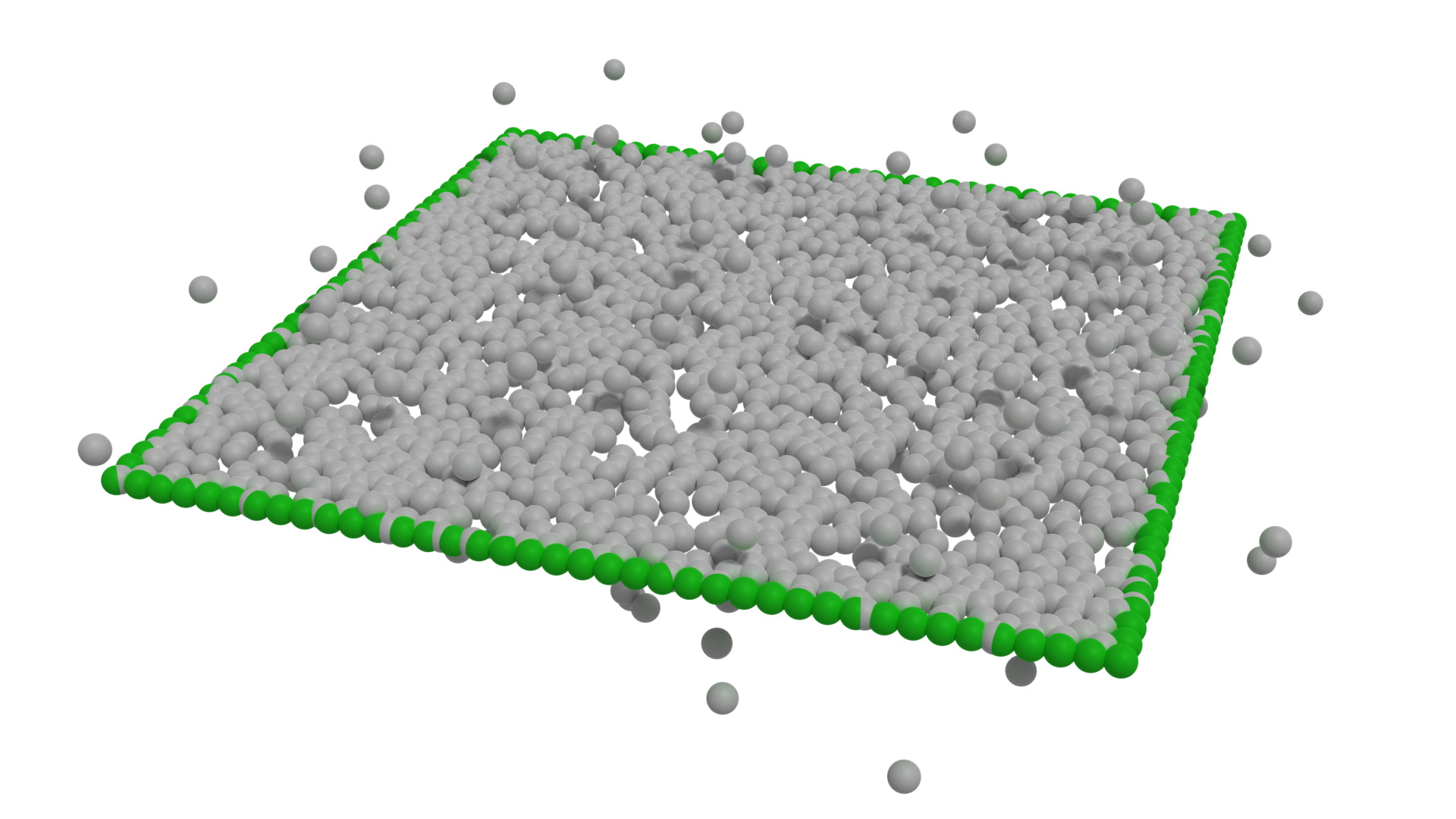}};
        \node[image, right=of mid2]  (right1)                                                             {\includegraphics[trim={250px 0 400px 0},clip,width=0.35\linewidth]{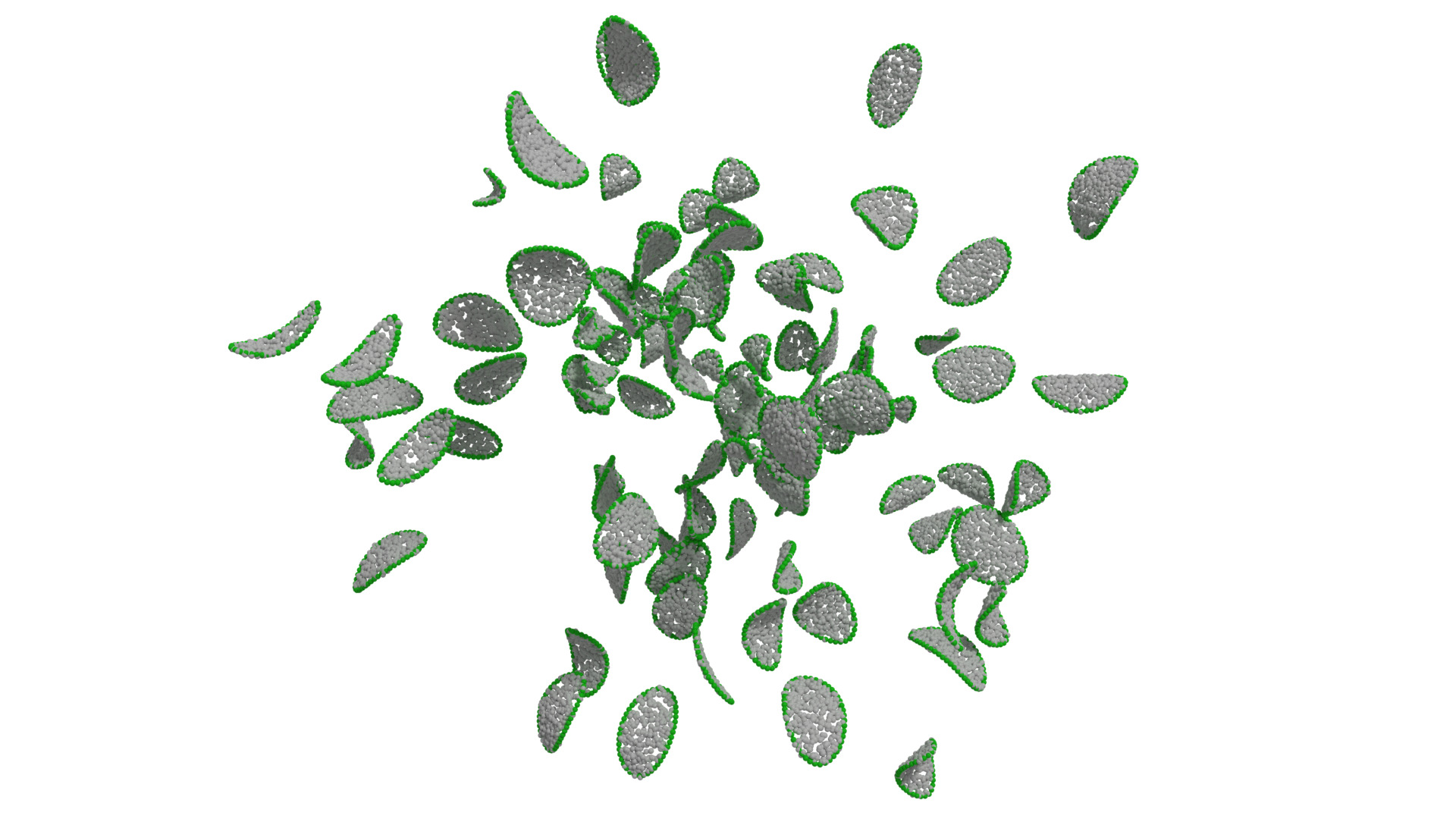}};

        \node[label_top, anchor=south] (l2) at (mid1.north)               {Additional Training Models};
        \node[label_top] (l1) at (left1.north |- l2.north)  {Modified Labels};
        \node[label_top] (l3) at (right1.north |- l2.north) {Additional Evaluation Model};
    \end{tikzpicture}
    \caption{\label{fig:default_plus_plus}
        Adjustments made to the \emph{Default} dataset to facilitate boundary detection.
        The labels of two point clouds (left) are modified to identify already included boundary points correctly.
        Three simple point clouds (middle) are added to the training and validation set to improve the coverage of potential boundary point cases.
        For an additional evaluation we add a further point cloud (right) to ensure a sufficient representation of boundary points in the evaluation data.
    }
\end{figure*}

\subsection{Datasets}\label{sec:datasets}

We train and evaluate our approach on several different datasets and provide comparisons to other point cloud edge detection algorithms.
To allow for a direct comparison with PCEDNet~\citep{himeur2021pcednet}, their \emph{Default} dataset as well as the publicly available \emph{ABC}~\citep{koch2019abc} dataset are used.
\paragraph{Default}
Introduced by \citet{himeur2021pcednet}, this dataset is designed to be as small as possible in order to facilitate very short training times with only a few simple hand-labeled point clouds to train on but still generalize well to arbitrary other point clouds.
It contains 9 point clouds for training as well as 7 different point clouds for evaluation.
To form the validation set, 1000 points are randomly sampled from each class.
Despite containing three different classes of points, i.e. non-edge, sharp-edge, and smooth-edge, originally, this work focuses on non-edge and sharp-edge classification only and therefore treats smooth-edge points as non-edge points in all results.
We train BoundED for 3000 iterations on this dataset.
\paragraph{ABC}
The \emph{ABC} dataset published by \citet{koch2019abc} is a very large collection of CAD models accompanied with triangle meshes and feature annotations among other data.
Point clouds are generated from triangle meshes by simply removing all edges and faces.
A ground truth classification label for each point is extracted by checking whether it is part of any CAD curve flagged as \emph{sharp}.
To ensure a meaningful comparison with the work by \citet{himeur2021pcednet}, we also only use chunk 0000 and exactly the same 200 models for training and 50 models for validation while also using all 7168 point clouds for evaluation.
As \emph{ABC} contains many more points than \emph{Default}, we train our network for 8000 iterations on its training data.

\paragraph{Default++}
As the original \emph{Default} dataset published by \citet{himeur2021pcednet} does not include annotated boundary vertices, which prevents its use for training models for boundary detection tasks, we propose to extend it as shown in Figure~\ref{fig:default_plus_plus} to create the \emph{Default++} dataset.
The original \emph{Default} dataset contains two models containing boundary points not annotated as such.
Thus, the first modification is to add these boundary annotations accordingly.
Furthermore, it is extended by two additional point clouds for training, which were specifically designed to contain clean and noisy curved boundaries of varying radii as these cases are not included in the original \emph{Default} dataset.
Finally, to prevent boundary points from being heavily underrepresented in the evaluation set, we additionally add an evaluation model containing a multitude of boundary situations with varying levels of noise.
Since the class of boundary points in the training set is still much smaller than the classes of non-edge or sharp-edge points, we only add 100 randomly sampled boundary points to the validation set.
The resulting training set contains 279.5k non-edge, 15.7k sharp-edge, and only 0.9k boundary points.
As it is similar in size to the \emph{Default} dataset, we use the same 3000 iterations to train on \emph{Default++}.
\paragraph{Additional Evaluation Data}
To assess the capabilities of the proposed algorithm more thoroughly, we also use publicly available point clouds of 3D scanned buildings and plants.
The \emph{christ\_church}\footnote{Available at: \url{https://sketchfab.com/3d-models/christ-church-and-dublin-city-council-b5f6bcce8ebc44a3b4bbb6b0fef067b3}, accessed on 10/14/2022.} point cloud contains 1.9 million points of the Christ Church Cathedral and its surrounding in Dublin.
Furthermore, the \emph{pisa\_cathedral}\footnote{Available at: \url{https://www.irit.fr/recherches/STORM/MelladoNicolas/category/datasets/}, accessed on 10/22/2022.} point cloud with 2.5 million points scanned by \citet{mellado2015relative} is used as well.
The \emph{station}\footnote{Available at: \url{https://sketchfab.com/3d-models/station-rer-6c636ca4793345e8ae12beb97b7d6359}, accessed on 10/14/2022.} point cloud is an even larger point cloud representing a train station as 12.5 million points which we also use for evaluation.
Finally, we are using point clouds of three different plants scanned by~\citet{conn2017statistical}: An Arabidopsis\footnote{Available at: \url{http://plant3d.navlakhalab.net/shoots/public/view/plant/40}, time point 33, accessed on 10/14/2022.}, a Tobacco\footnote{Available at: \url{http://plant3d.navlakhalab.net/shoots/public/view/plant/20}, time point 30, accessed on 10/14/2022.}, and a Tomato\footnote{Available at: \url{http://plant3d.navlakhalab.net/shoots/public/view/plant/15}, time point 30, accessed on 10/14/2022.} plant with 172k, 1474k, and 226k points respectively.

\subsection{Metrics}

Similarly to the work by \citet{himeur2021pcednet}, we use several metrics for comparison: Precision, Recall, Matthews Correlation Coefficient (MCC), F1 score, Accuracy, and Intersection over Union (IoU, also known as Jaccard index).
Precision evaluates the ratio of true classifications as sharp-edge or boundary to the total number of classified points.
In contrast, Recall measures the ratio of correctly classified sharp-edge or boundary points compared to the true number of such points existing in the processed model.
Precision and Recall are coupled, i.e. Precision increases and Recall decreases if only points exhiting very high confidence are classified and vice versa.
Thus, mainly the other mentioned metrics, which combine Precision and Recall scores in different ways, are used for directly comparing our BoundED technique to related works.

\begin{table*}[t]
    \scriptsize
    \begin{center}
        \begin{tabular}{lcccccc}\toprule
                                                                       & Precision($\uparrow$) & Recall($\uparrow$) & MCC($\uparrow$) & F1($\uparrow$) & Accuracy($\uparrow$) & IoU($\uparrow$) \\ \midrule
            CA (Default)~\citep{bazazian2015fast}                      & 0.184      & \bf{0.891} & 0.332      & 0.305      & 0.753      & 0.178      \\
            CA (ABC)~\citep{bazazian2015fast}                          & 0.183      & 0.357      & 0.188      & 0.242      & 0.863      & 0.138      \\
            FEE (Default)~\citep{cgal:ass-psp-22b,merigot2011voronoi}  & 0.241      & 0.866      & 0.400      & 0.376      & 0.828      & 0.232      \\
            FEE (ABC)~\citep{cgal:ass-psp-22b,merigot2011voronoi}      & 0.060      & 0.961      & -0.021     & 0.113      & 0.082      & 0.060      \\
            PCEDNet-2c (\emph{Default})~\citep{himeur2021pcednet}      & 0.364      & 0.611      & 0.402      & 0.430      & 0.908      & 0.274      \\
            BoundED (Ours) (\emph{Default})                            & \bf{0.365} & 0.595      & \bf{0.423} & \bf{0.453} & \bf{0.912} & \bf{0.293} \\
            BoundED (Ours) (\emph{ABC})                                & 0.248      & 0.589      & 0.328      & 0.348      & 0.869      & 0.210      \\
            \bottomrule
        \end{tabular}
        \caption{\label{tab:quant_eval_default}
            Median scores of edge detection approaches evaluated on the \emph{Default} dataset.
            The dataset used for parameter tuning or training is mentioned in parentheses.
            Data regarding PCEDNet-2c is taken from \citet{himeur2021pcednet}.
        }
    \end{center}
\end{table*}

\begin{table*}[t]
    \scriptsize
    \begin{center}
        \begin{tabular}{lcccccc}\toprule
                                                                       & Precision($\uparrow$) & Recall($\uparrow$) & MCC($\uparrow$) & F1($\uparrow$) & Accuracy($\uparrow$) & IoU($\uparrow$) \\ \midrule
            CA (Default)~\citep{bazazian2015fast}                      & 0.312      & \bf{0.991} & 0.482      & 0.471      & 0.845      & 0.308      \\
            CA (ABC)~\citep{bazazian2015fast}                          & 0.498      & 0.820      & 0.541      & 0.574      & 0.929      & 0.403      \\
            FEE (Default)~\citep{cgal:ass-psp-22b,merigot2011voronoi}  & 0.178      & 0.621      & 0.213      & 0.270      & 0.775      & 0.156      \\
            FEE (ABC)~\citep{cgal:ass-psp-22b,merigot2011voronoi}      & 0.857      & 0.898      & 0.821      & 0.832      & 0.980      & 0.712      \\
            PCEDNet-2c (\emph{Default})~\citep{himeur2021pcednet}      & 0.662      & 0.936      & 0.708      & 0.730      & 0.958      & 0.574      \\
            PCEDNet-2c (\emph{ABC})~\citep{himeur2021pcednet}          & 0.735      & 0.984      & 0.808      & 0.822      & 0.970      & 0.597      \\
            ECNet (\emph{ABC})~\citep{yu2018ec}                        & 0.487      & 0.573      & -          & 0.526      & -          & 0.356      \\
            PIE-NET (\emph{ABC})~\citep{wang2020pie}                   & 0.692      & 0.858      & -          & 0.766      & -          & 0.622      \\
            PCPNet-2c (\emph{ABC})~\citep{guerrero2018pcpnet}          & \bf{0.954} & 0.756      & 0.797      & 0.807      & 0.979      & 0.668      \\
            BoundED-2c (Ours) (\emph{Default})                         & 0.420      & 0.594      & 0.381      & 0.420      & 0.909      & 0.266      \\
            BoundED-2c (Ours) (\emph{ABC})                             & 0.932      & 0.833      & \bf{0.842} & \bf{0.850} & \bf{0.983} & \bf{0.739} \\
            \bottomrule
        \end{tabular}
        \caption{\label{tab:quant_eval_abc}
            Median scores of edge detection approaIn particular, tches evaluated on the \emph{ABC} dataset.
            The dataset used for training is mentioned in parentheses.
            Data regarding PCEDNet-2c, PCPNET-2c is taken from \citet{himeur2021pcednet}.
            Data regarding ECNet, and PIE-NET is taken from \citet{wang2020pie}.
        }
    \end{center}
\end{table*}

\begin{figure*}[t]
    \centering
    \begin{tikzpicture}[image/.style = {inner sep=0mm, outer sep=0pt},
                        label_top/.style = {anchor=south, align=center},
                        label_side/.style = {rotate=90, anchor=south, align=center},
                        node distance = 12mm and 2mm]
        \node[image]               (i11) {\includegraphics[width=0.18\linewidth]{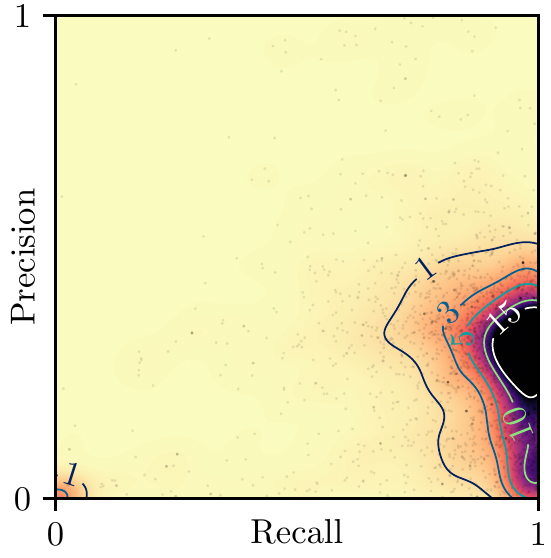}};
        \node[label_top] (l11) at (i11.north) {CA\\(\emph{Default})};
        \node[image, right=of i11] (i12) {\includegraphics[width=0.18\linewidth]{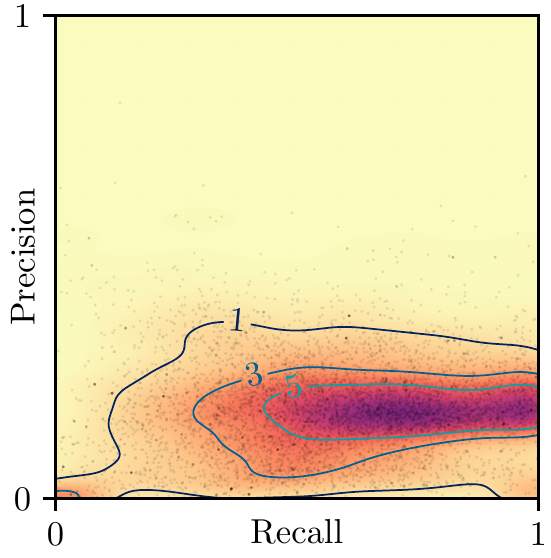}};
        \node[label_top] (l12) at (i12.north) {FEE\\(\emph{Default})};
        \node[image, right=of i12] (i13) {\includegraphics[width=0.18\linewidth]{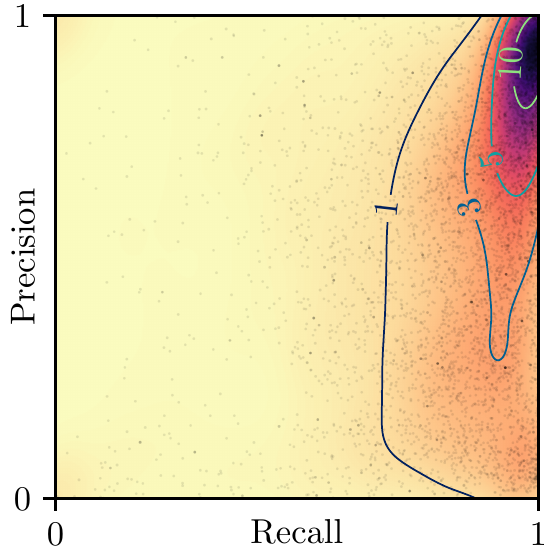}};
        \node[label_top] (l13) at (i13.north) {PCEDNet-2c\\(\emph{Default})};
        \node[image, right=of i13] (i14) {\includegraphics[width=0.18\linewidth]{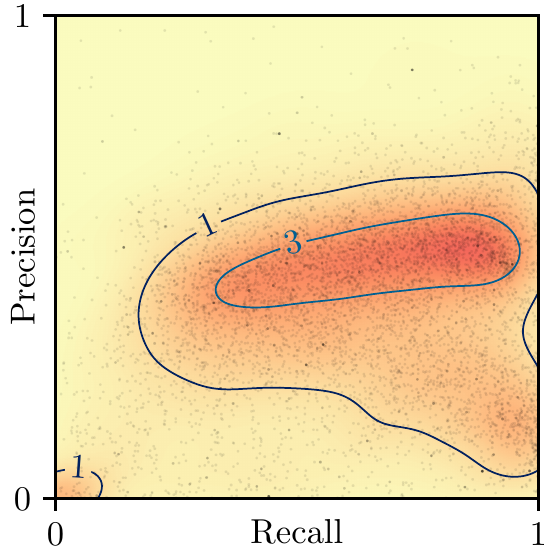}};
        \node[label_top] (l14) at (i14.north) {ECNet\\(\emph{ABC})};
        \node[image, right=of i14] (i15) {\includegraphics[width=0.18\linewidth]{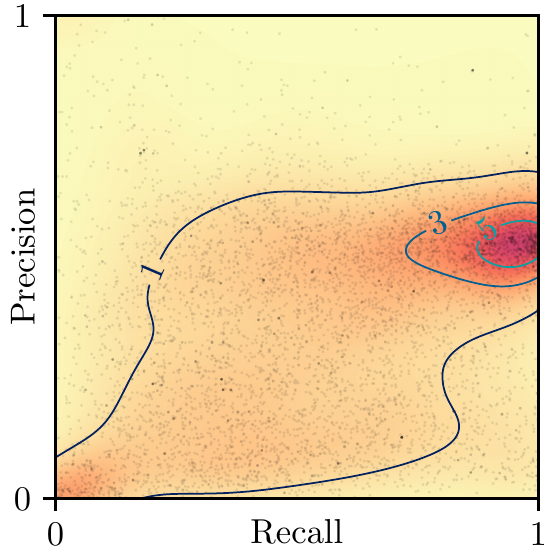}};
        \node[label_top] (l15) at (i15.north) {BoundED (Ours)\\(\emph{Default})};
        \node[image, below=of i11] (i21) {\includegraphics[width=0.18\linewidth]{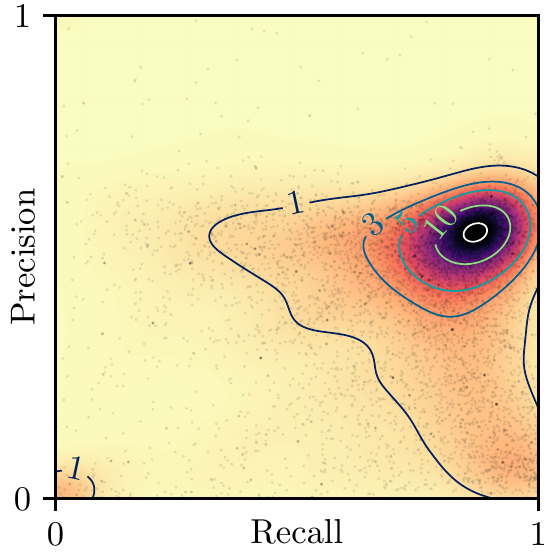}};
        \node[label_top] (l21) at (i21.north) {CA\\(\emph{ABC})};
        \node[image, right=of i21] (i22) {\includegraphics[width=0.18\linewidth]{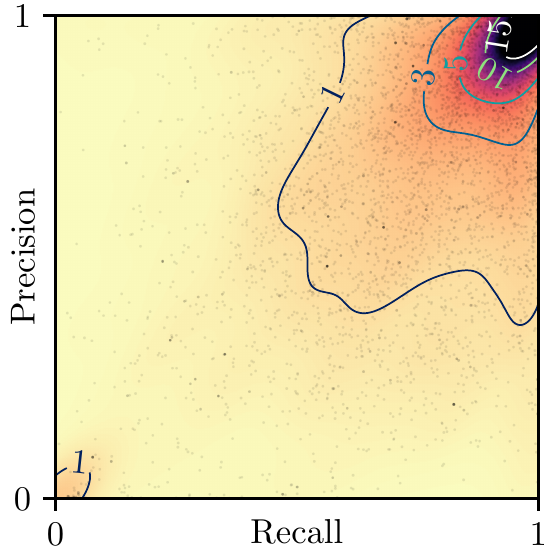}};
        \node[label_top] (l22) at (i22.north) {FEE\\(\emph{ABC})};
        \node[image, right=of i22] (i23) {\includegraphics[width=0.18\linewidth]{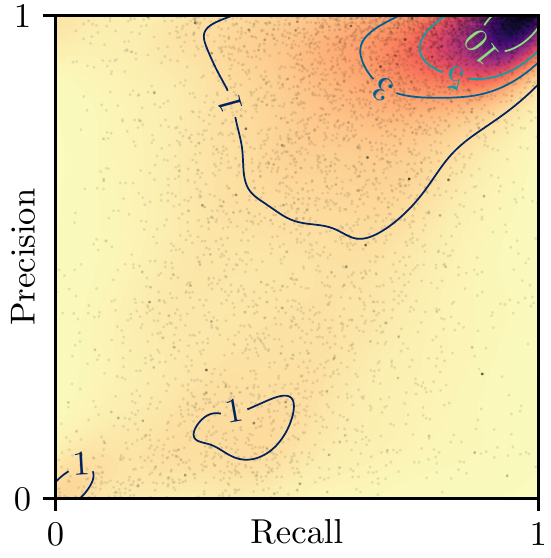}};
        \node[label_top] (l23) at (i23.north) {PCEDNet-2c\\(\emph{ABC})};
        \node[image, right=of i23] (i24) {\includegraphics[width=0.18\linewidth]{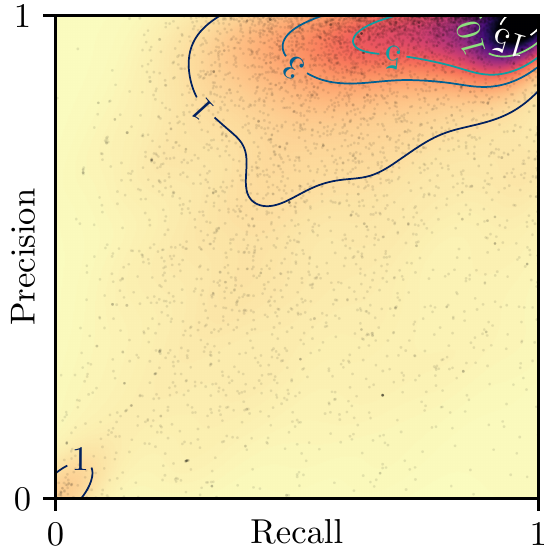}};
        \node[label_top] (l24) at (i24.north) {PCPNet-2c\\(\emph{ABC})};
        \node[image, right=of i24] (i25) {\includegraphics[width=0.18\linewidth]{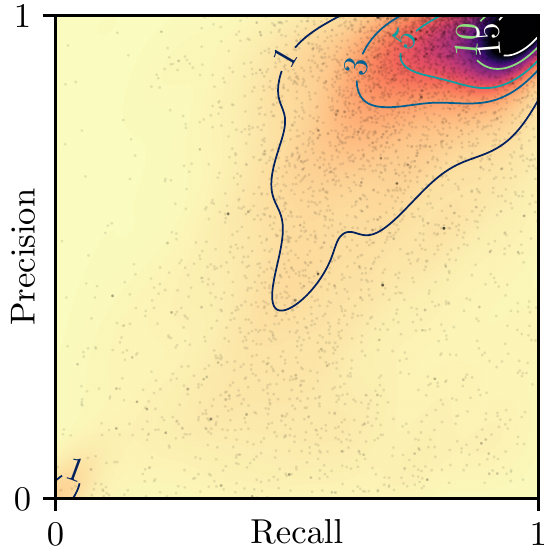}};
        \node[label_top] (l25) at (i25.north) {BoundED (Ours)\\(\emph{ABC})};

        \node[inner sep=0mm, outer sep=0pt, anchor=north west] (colorbar) at ($(i15.north east)+(2mm, 0)$) {\includegraphics[width=0.0566\linewidth]{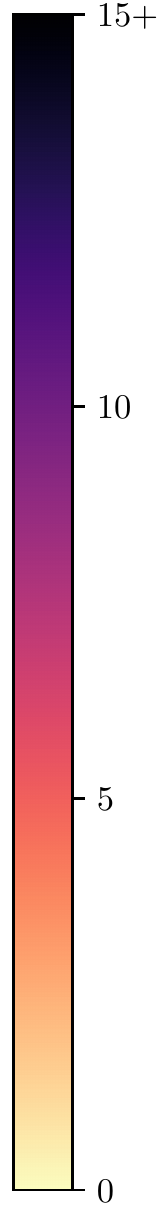}};
    \end{tikzpicture}
    \caption{\label{fig:precision_recall}
        Precision-Recall-plots of most approaches listed in Table~\ref{tab:quant_eval_abc}.
        Every small semi-transparent black dot corresponds to a single point cloud from the \emph{ABC} dataset and its Precision and Recall scores when being processed by the respective approach.
        The background depicts the color-coded local density of points.
    }
\end{figure*}

\subsection{Comparison to Related Work}

Throughout this section, we compare the performance of our work with the performance of several other recent related works for point cloud edge detection: Covariance Analysis (CA)~\citep{bazazian2015fast}, Feature Edges Estimation (FEE)~\citep{cgal:ass-psp-22b,merigot2011voronoi}, ECNet~\citep{yu2018ec}, PIE-NET~\citep{wang2020pie}, PCPNet~\citep{guerrero2018pcpnet}, and PCEDNet~\citep{himeur2021pcednet}.
The postfix \emph{-2c} denotes that the respective algorithm has been trained for classification of two classes only, i.e. non-edge and sharp-edge, despite being originally designed to potentially handle more than two classes. 
For the quantitative evaluation (see Section~\ref{sec:quant_eval}), data reported by \citet{himeur2021pcednet} is used for PCEDNet and PCPNet, while we use the numbers published by \citet{wang2020pie} for ECNet and PIE-Net.
For the two non-learning methods CA and FEE, we use one set of parameters each per dataset finetuned on the dataset's characteristics, i.e. more aggressive thresholding on clean data compared to noisy data:
CA finetuned on \emph{Default} uses 0.025 as threshold, while using 0.08 on \emph{ABC}.
The parameters for FEE are set to $R=0.1$, $r=0.03$ to work well with the \emph{Default} dataset and to $R=0.02$ and $r=0.002$ to yield good results on the \emph{ABC} dataset.
In both cases, we use 0.16 as threshold.
For FEE, we additionally normalize all point clouds to fit inside an axis-aligned unit box as $R$ and $r$ are related to the expected feature size, which varies heavily for the models in the \emph{ABC} dataset.
The PCEDNet results shown for the purpose of qualitative evaluation in Section~\ref{sec:qual_eval} are generated using the publicly available precompiled demo application\footnote{Available at: \url{https://storm-irit.github.io/pcednet-supp/software.html}, accessed on 10/14/2022.}.
We assume only the point positions to be given as input for the algorithm.
Since PCEDNet relies on point normals, these are generated according to the authors' specification using Meshlab~\citep{meshlab}.
To be able to report meaningful numbers for the quantitative evaluation in Section~\ref{sec:quant_eval}, we have done every experiment five times, evaluated the loss function over the validation set, and chose the best result according to this metric.
To ensure practicality of our algorithm, timings are reported for two different hardware configurations:
On the one hand, we use an old consumer-grade Nvidia RTX 2080 Ti GPU with 11GB memory and an AMD Ryzen 3600X CPU with 32GB memory.
On the other hand, we also used the recent enterprise Nvidia A40 GPU with 46GB memory and two AMD EPYC 7313 CPUs with 32 threads each and 512GB memory.
Note, however, that we only used 12 worker threads in the data loader during training for both hardware configurations.
We exclude the IO and network initialization time from the timings listed in this section and focus on reporting the time required by the actual feature extraction as well as network inference instead.

\subsection{Quantitative Comparison}\label{sec:quant_eval}

Tables~\ref{tab:quant_eval_default} and~\ref{tab:quant_eval_abc} show median scores of various commonly used metrics to allow a quantitative comparison of our approach with others.
For all experiments in this section, we are working on the \emph{Default} and \emph{ABC} datasets and aim at distinguishing sharp-edge points from non-edge points.
When training and evaluation are done on the \emph{Default} dataset, our algorithm performs better than all related works in all metrics except for Recall, i.e. BoundED is not able to identify quite as many sharp-edge points as others, but more of those points classified as being a sharp-edge point are actually correctly identified as such.
As we are also using a smaller network in comparison to PCEDNet, this suggests, that our multi-scale features are better at describing the geometry of the local neighborhood in terms of sharp edges than their GLS based features.

\begin{table*}[!t]
    \scriptsize
    \begin{center}
        \begin{tabular}{lcccc}\toprule
                                                                  & \multicolumn{2}{c}{Training} & \multicolumn{2}{c}{Evaluation} \\ \cmidrule(lr){2-3}\cmidrule(lr){4-5}
                                                                  & Preprocessing & Training     & Preprocessing & Classification \\ \midrule
            PCEDNet (\emph{Default})~\citep{himeur2021pcednet}    & 0:19 m        & 2:52 m       & -             & -              \\
            BoundED-2c (Ours) (\emph{Default}, RTX 2080 Ti)       & 0:04 m        & 1:24 m       & 1.0 s         & 0.002 s        \\
            BoundED-2c (Ours) (\emph{Default}, A40)               & 0:04 m        & 1:13 m       & 1.0 s         & 0.004 s        \\ \midrule
            PCEDNet-2c (\emph{ABC})~\citep{himeur2021pcednet}     & 2:11 m        & 20:00 m      & 2:35:00 h     & 0:25:30 h      \\
            BoundED-2c (Ours) (\emph{ABC}, RTX 2080 Ti)           & 1:07 m        & 2:24 m       & 1:39:06 h     & 0:00:03 h      \\
            BoundED-2c (Ours) (\emph{ABC}, A40)                   & 0:59 m        & 3:00 m       & 1:20:17 h     & 0:00:04 h      \\ \midrule
            BoundED (Ours) (\emph{Default++}, RTX 2080 Ti)        & 0:05 m        & 1:22 m       & 1.4 s         & 0.002 s        \\
            BoundED (Ours) (\emph{Default++}, A40)                & 0:04 m        & 1:13 m       & 1.4 s         & 0.008 s        \\
            \bottomrule
        \end{tabular}
        \caption{\label{tab:performance}
            Comparison of time required to calculate the multi-scale features used as network input and training or evaluation time on the training or evaluation data respectively of the dataset in parentheses.
            Timings of our approach are determined on two different hardware configurations: An older consumer grade Nvidia RTX 2080 Ti GPU with 11GB memory and a recent enterprise grade Nvidia A40 GPU with 46GB memory.
            Data regarding PCEDNet and PCEDNet-2c is taken from \citet{himeur2021pcednet}.
        }
    \end{center}
\end{table*}

Also observe, that BoundED trained on \emph{ABC} performs better than CA and FEE finetuned on \emph{ABC} when evaluating on the \emph{Default} dataset.
Both non-learning approaches, i.e. CA and FFE, rely on setting a threshold to distinguish between sharp-edge and non-edge points.
On clean data like the models from the \emph{ABC} dataset, this threshold can be set much more aggressively.
In the presence of noise, this, however, leads to the algorithms not detecting all edges in the case of CA and tremendous overclassification of points as sharp-edge points in the case of FEE.
When being evaluated on \emph{ABC}, BoundED trained on \emph{ABC} once again outperforms all other approaches in terms of MCC, F1, Accuracy, and IoU scores, but PCEDNet loses less effectiveness if being trained on \emph{Default} in comparison to BoundED.
While PCPNet has the highest Precision and CA trained on \emph{Default} exhibits the highest Recall, they are worse in terms of overall classification performance due to having much worse scores in Recall and Precision respectively.
The Precision-Recall-plots shown in Figure~\ref{fig:precision_recall} confirm these observations.
In these diagrams, every point cloud of the \emph{ABC} dataset is depicted as one small semi-transparent black point according to it Precision and Recall scores.
The background color depicts the color-coded local density of points.
The plot for BoundED trained on the \emph{ABC} dataset exhibits the highest density in the top right corner suggesting that the classification results on most models are of high quality, while the peak density for approaches trained or finetuned on \emph{Default} is much lower and the individual points are more evenly distributed over a larger area.
Besides yielding better classification scores across the board, the computation of our features is also cheaper compared to PCEDNet and our multi-scale fusion and classification network has roughly $25\%$ less parameters.
Table~\ref{tab:performance} lists training and evaluation timings for PCEDNet and our approach.
Training in this context consists of the multi-scale feature extraction for the training and validation data of the dataset given in parentheses as well as using this data to train the network.
Similarly, evaluation consists of extracting the features on the evaluation set given in parentheses and classifying all points using the trained network.
While using a powerful GPU accelerates the feature extraction step, the difference for the network training and inference is negligible due to the networks compactness and simplicity.

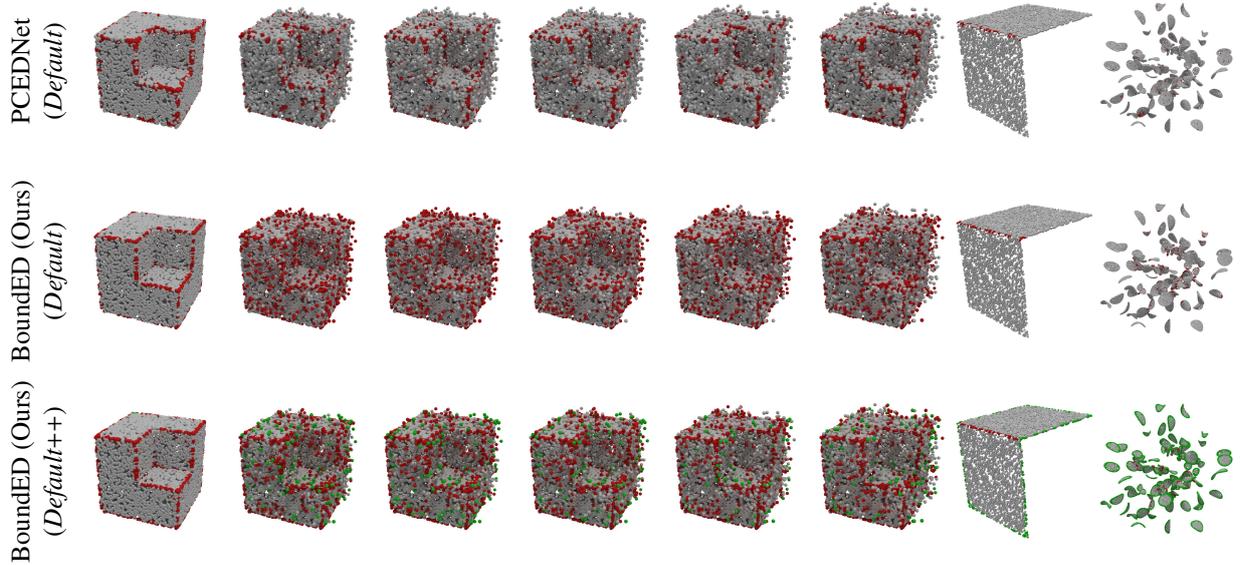
\begin{figure*}[t]
    \centering
    \begin{tikzpicture}[image/.style = {inner sep=0mm, outer sep=0pt},
                        label_top/.style = {anchor=south},
                        label_side/.style = {rotate=90, anchor=south, align=center},
                        node distance = 8mm and 1mm]
        \node[image]               (i11) {\includegraphics[trim={220px 0 220px 0},clip,width=0.11\textwidth]{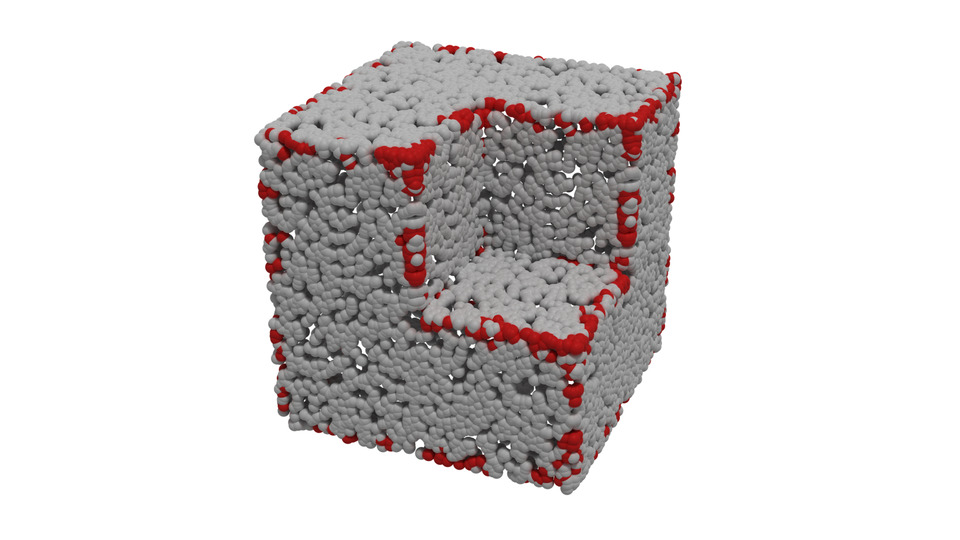}};
        \node[image, right=of i11] (i12) {\includegraphics[trim={220px 0 220px 0},clip,width=0.11\textwidth]{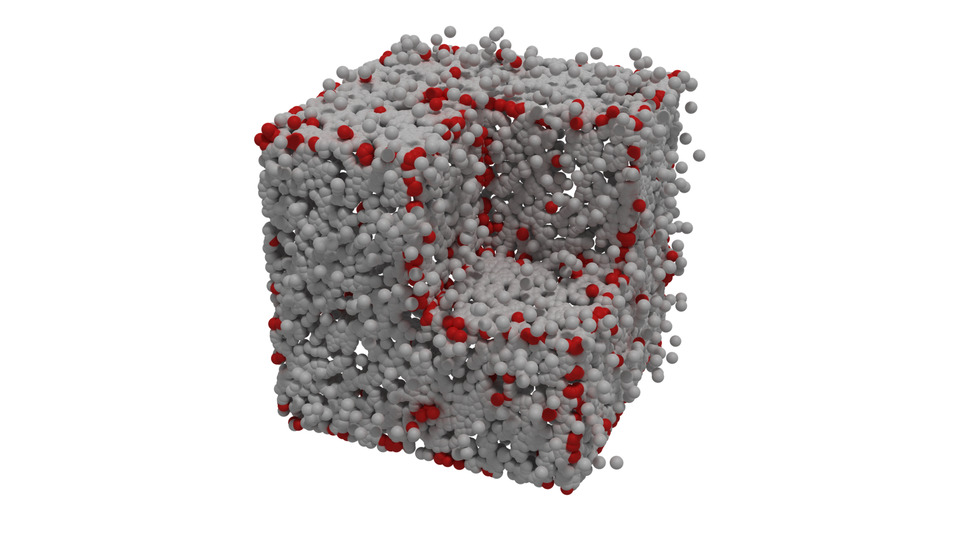}};
        \node[image, right=of i12] (i13) {\includegraphics[trim={220px 0 220px 0},clip,width=0.11\textwidth]{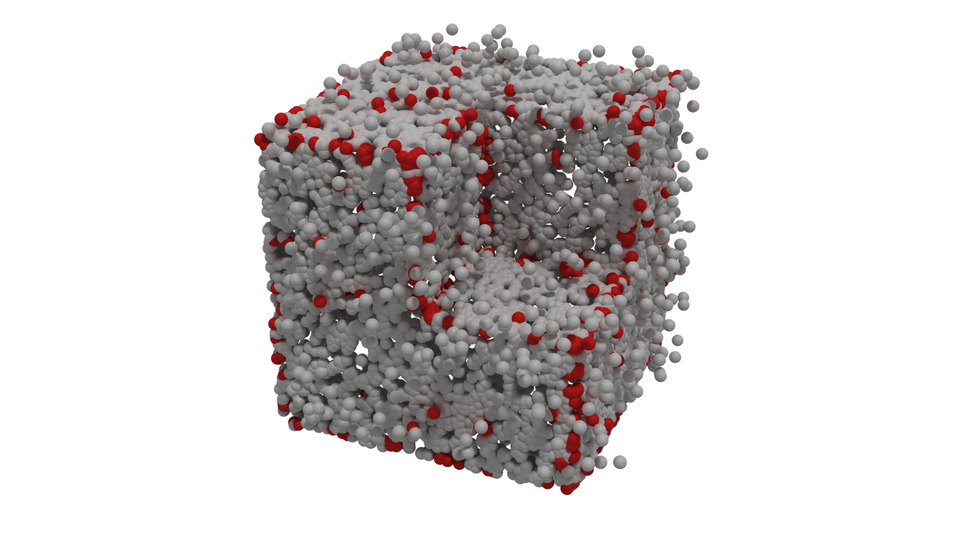}};
        \node[image, right=of i13] (i14) {\includegraphics[trim={220px 0 220px 0},clip,width=0.11\textwidth]{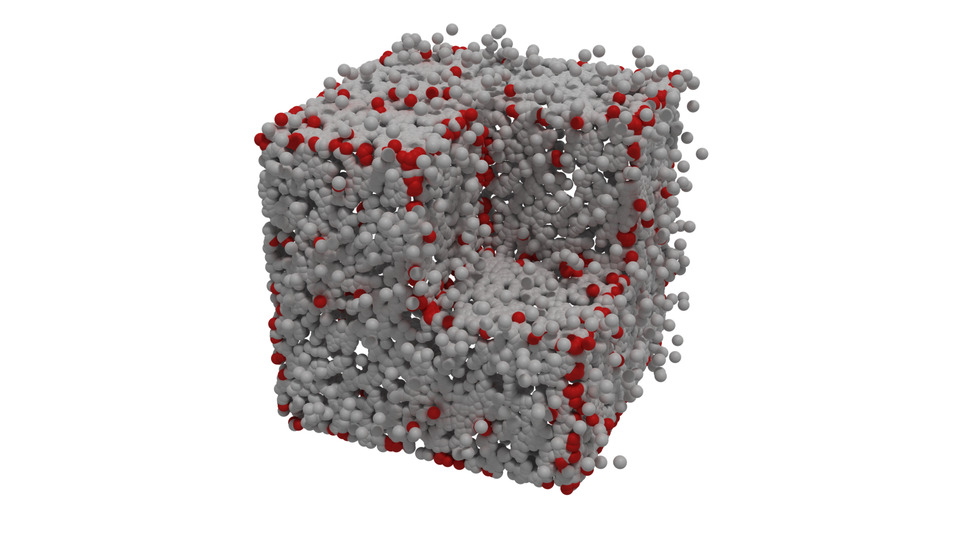}};
        \node[image, right=of i14] (i15) {\includegraphics[trim={220px 0 220px 0},clip,width=0.11\textwidth]{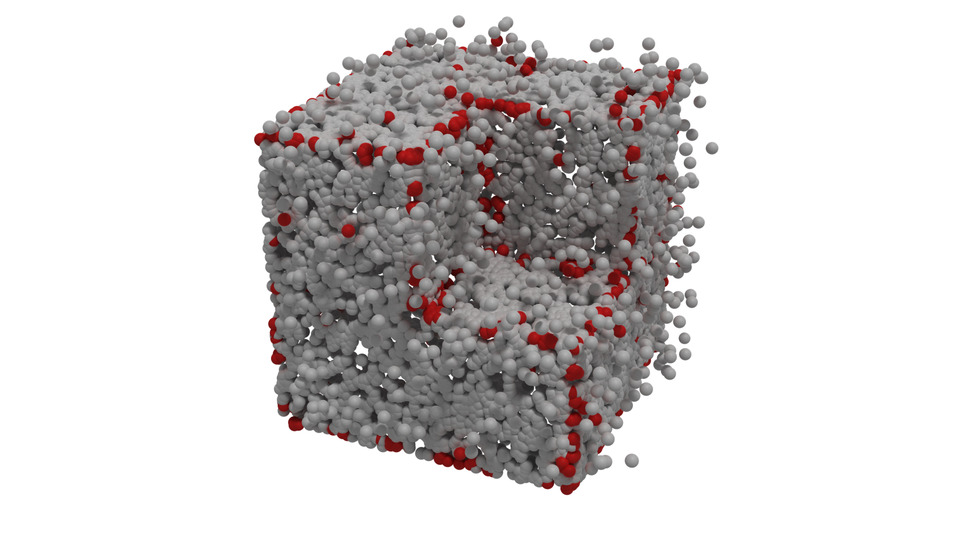}};
        \node[image, right=of i15] (i16) {\includegraphics[trim={220px 0 220px 0},clip,width=0.11\textwidth]{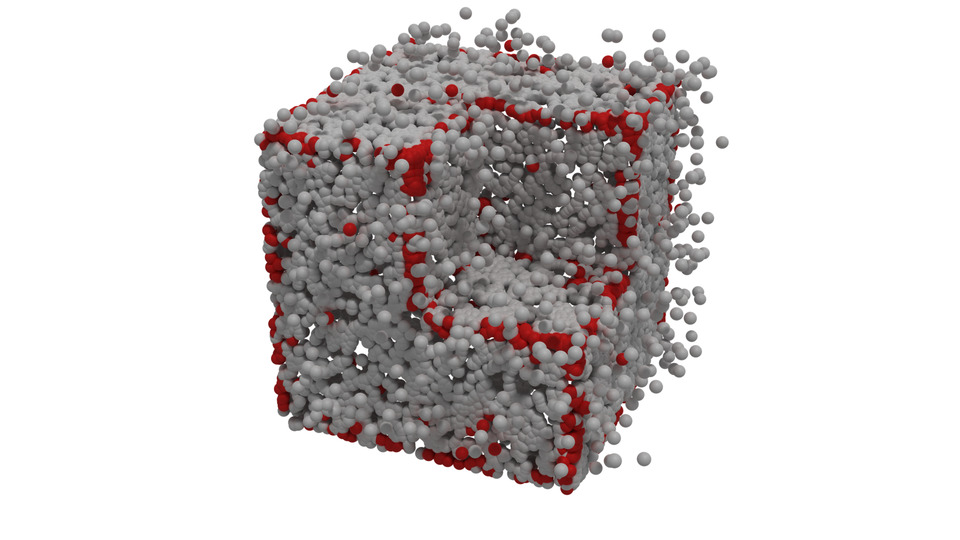}};
        \node[image, right=of i16] (i17) {\includegraphics[trim={220px 0 220px 0},clip,width=0.11\textwidth]{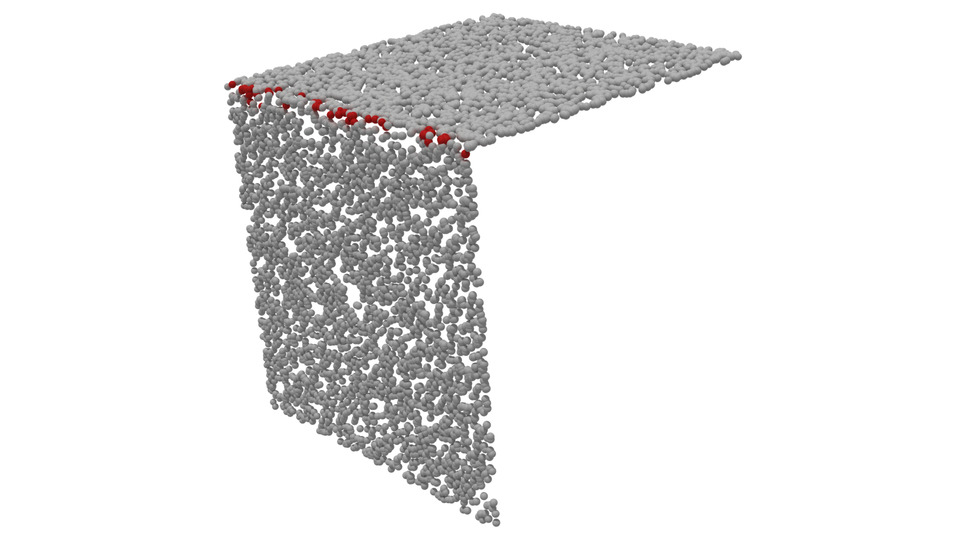}};
        \node[image, right=of i17] (i18) {\includegraphics[trim={220px 0 220px 0},clip,width=0.11\textwidth]{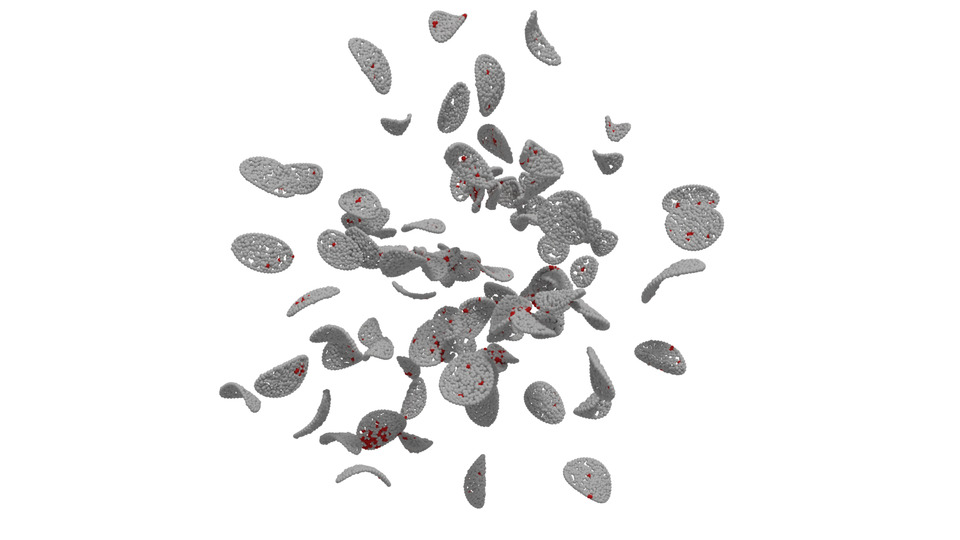}};
        \node[image, below=of i11] (i21) {\includegraphics[trim={220px 0 220px 0},clip,width=0.11\textwidth]{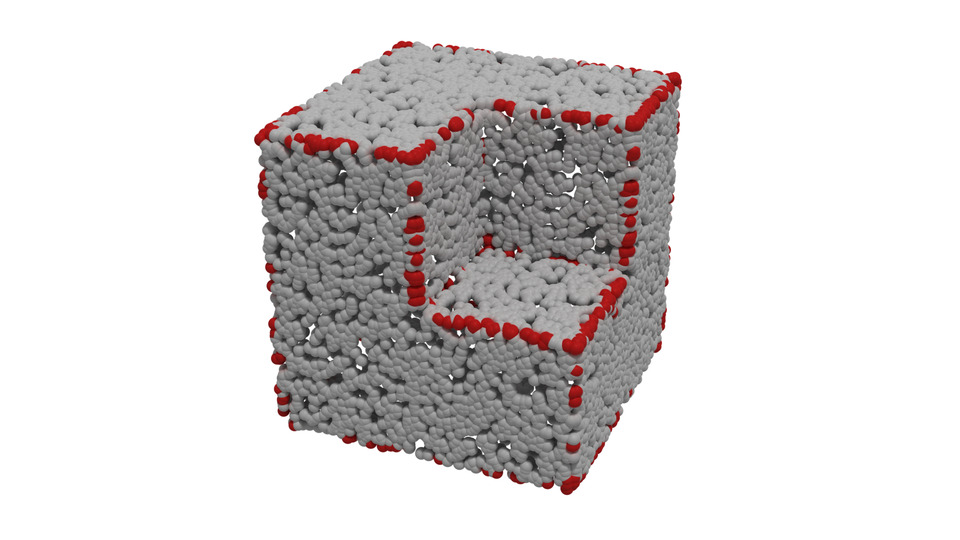}};
        \node[image, right=of i21] (i22) {\includegraphics[trim={220px 0 220px 0},clip,width=0.11\textwidth]{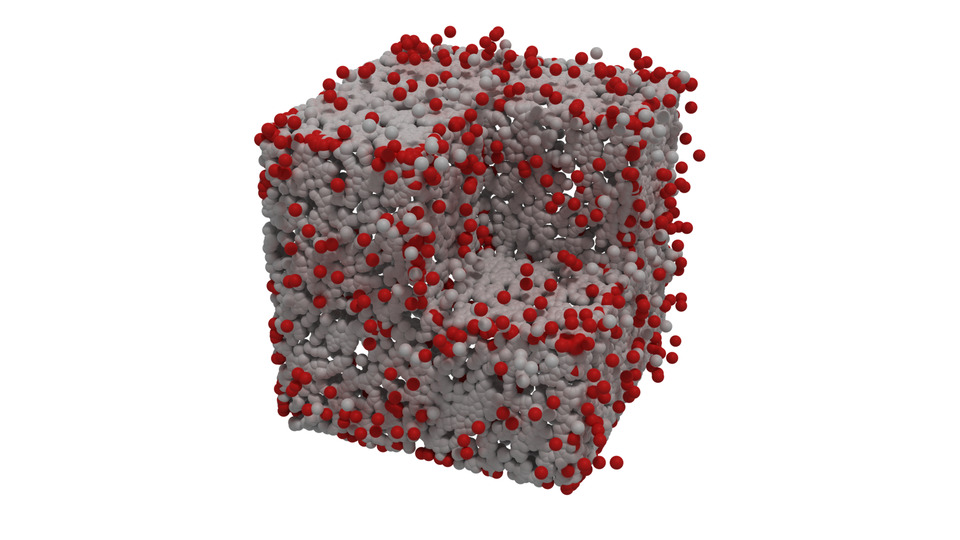}};
        \node[image, right=of i22] (i23) {\includegraphics[trim={220px 0 220px 0},clip,width=0.11\textwidth]{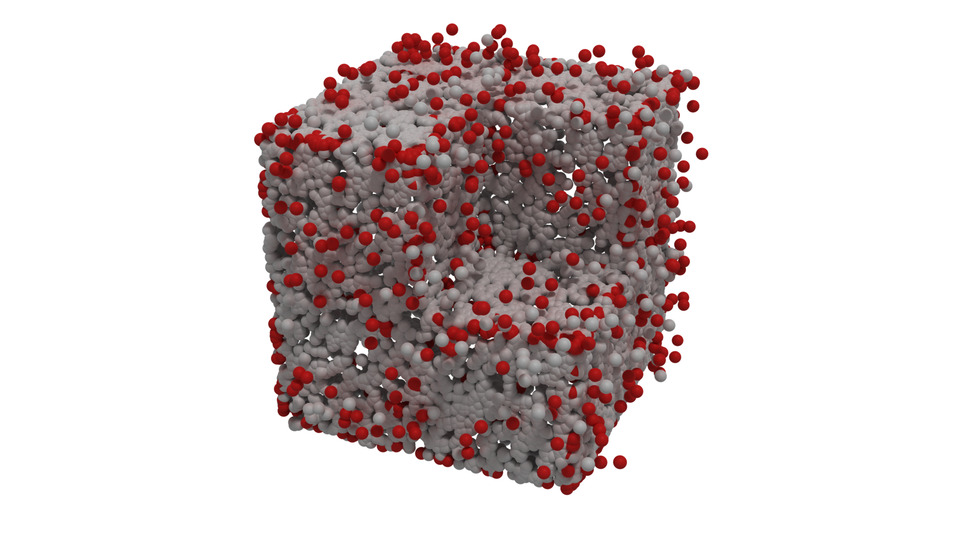}};
        \node[image, right=of i23] (i24) {\includegraphics[trim={220px 0 220px 0},clip,width=0.11\textwidth]{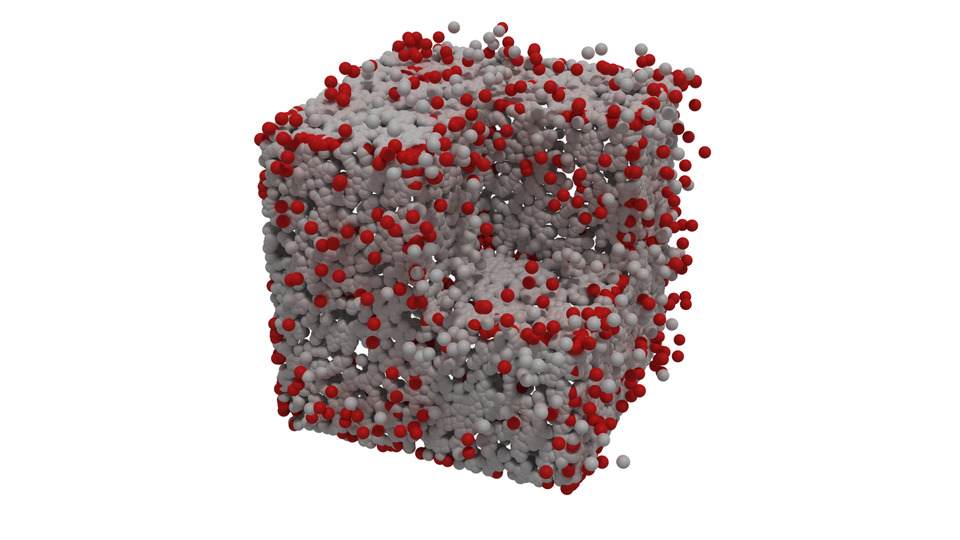}};
        \node[image, right=of i24] (i25) {\includegraphics[trim={220px 0 220px 0},clip,width=0.11\textwidth]{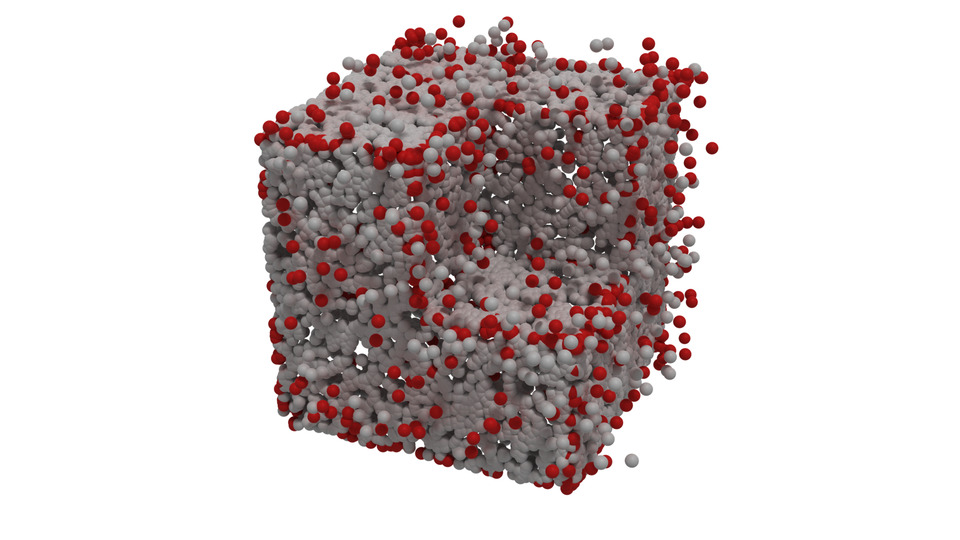}};
        \node[image, right=of i25] (i26) {\includegraphics[trim={220px 0 220px 0},clip,width=0.11\textwidth]{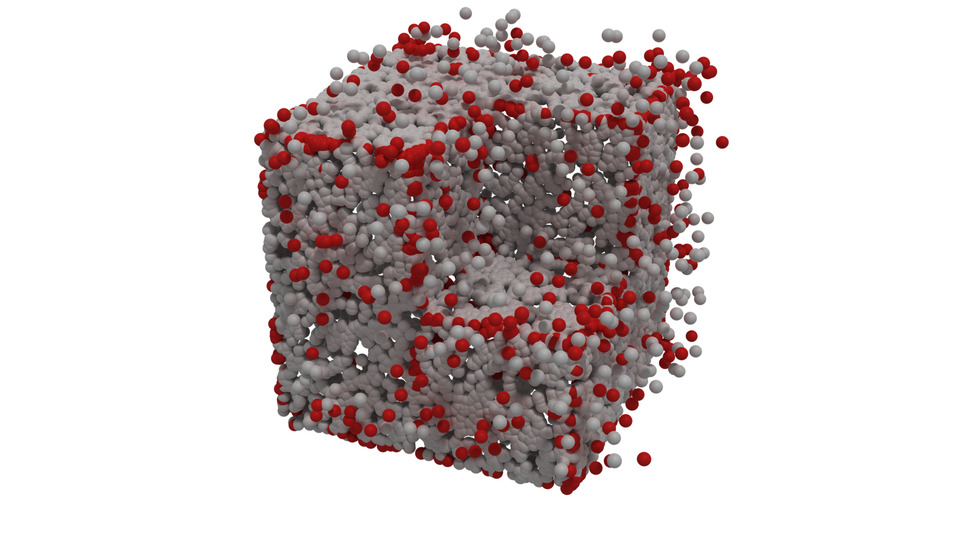}};
        \node[image, right=of i26] (i27) {\includegraphics[trim={220px 0 220px 0},clip,width=0.11\textwidth]{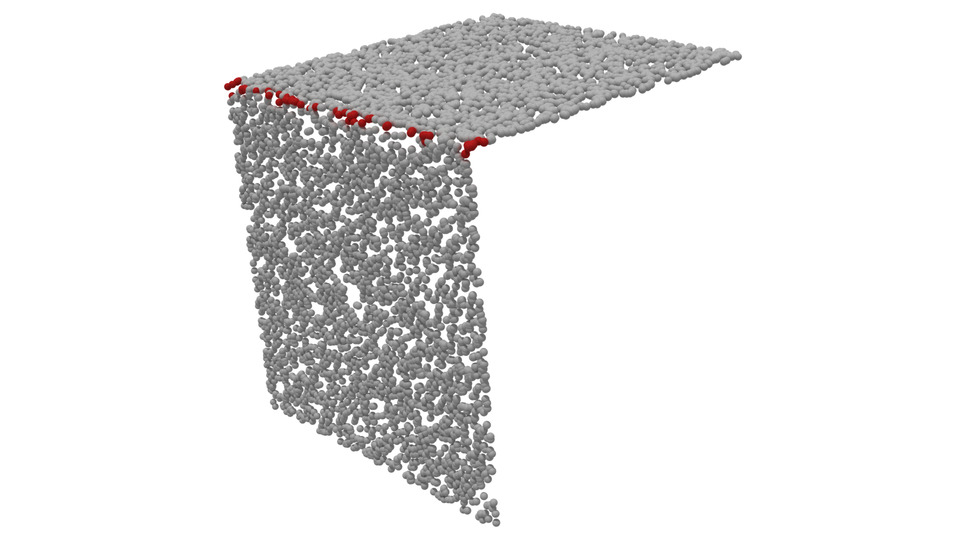}};
        \node[image, right=of i27] (i28) {\includegraphics[trim={220px 0 220px 0},clip,width=0.11\textwidth]{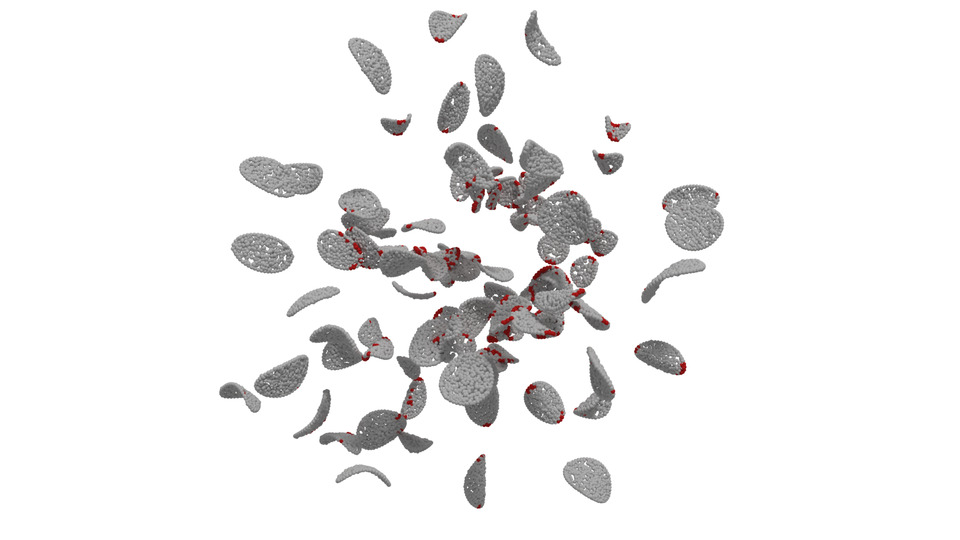}};
        \node[image, below=of i21] (i31) {\includegraphics[trim={220px 0 220px 0},clip,width=0.11\textwidth]{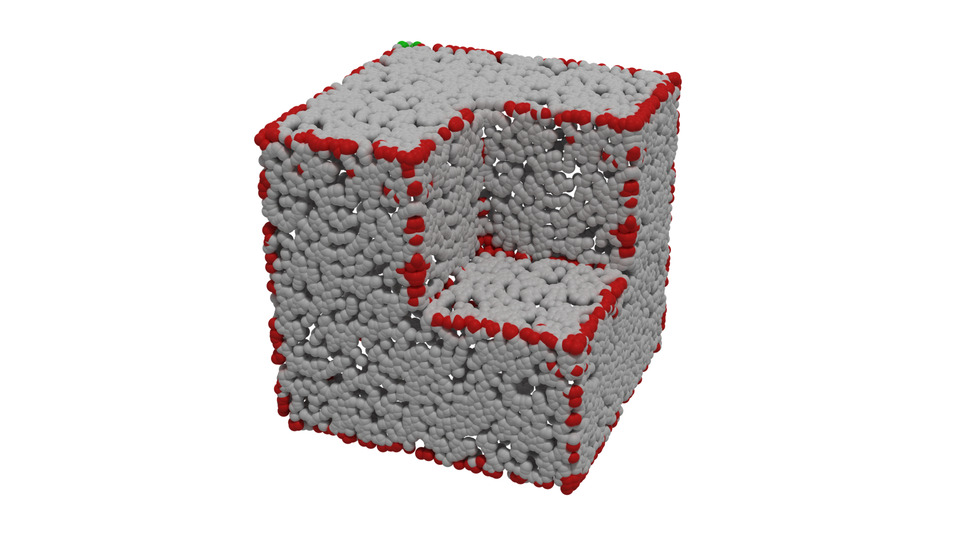}};
        \node[image, right=of i31] (i32) {\includegraphics[trim={220px 0 220px 0},clip,width=0.11\textwidth]{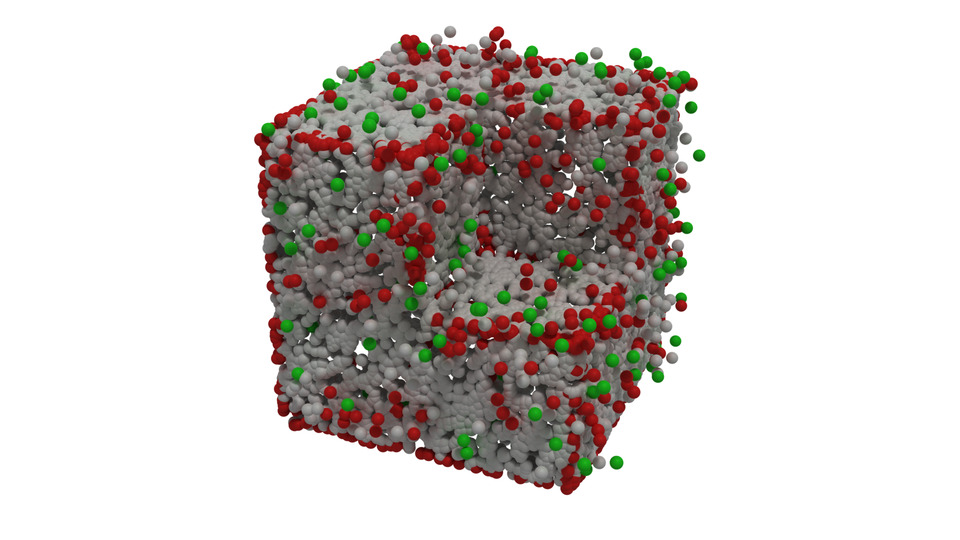}};
        \node[image, right=of i32] (i33) {\includegraphics[trim={220px 0 220px 0},clip,width=0.11\textwidth]{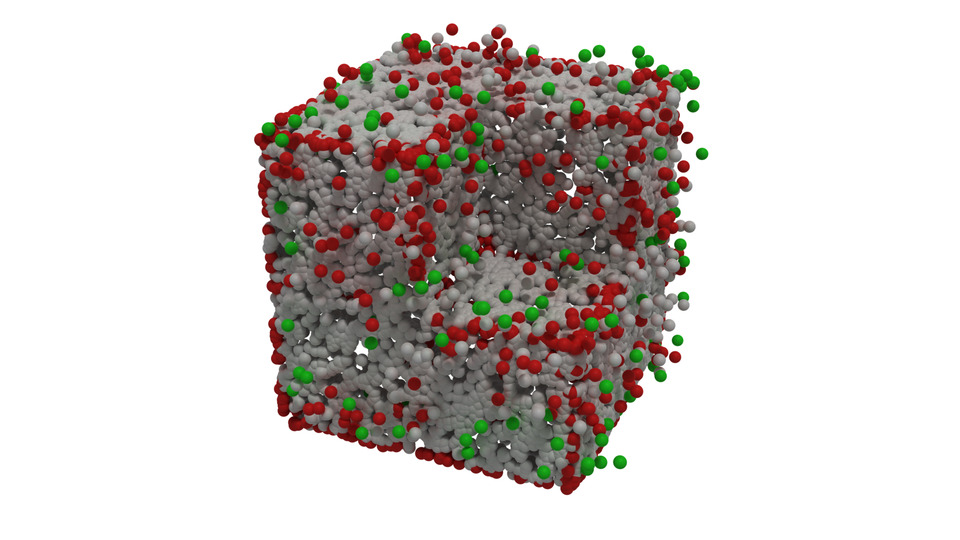}};
        \node[image, right=of i33] (i34) {\includegraphics[trim={220px 0 220px 0},clip,width=0.11\textwidth]{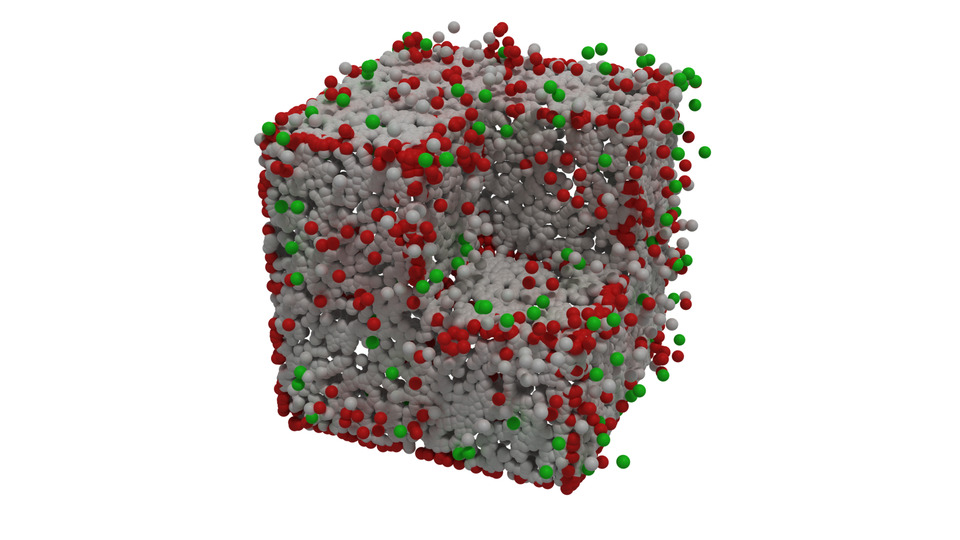}};
        \node[image, right=of i34] (i35) {\includegraphics[trim={220px 0 220px 0},clip,width=0.11\textwidth]{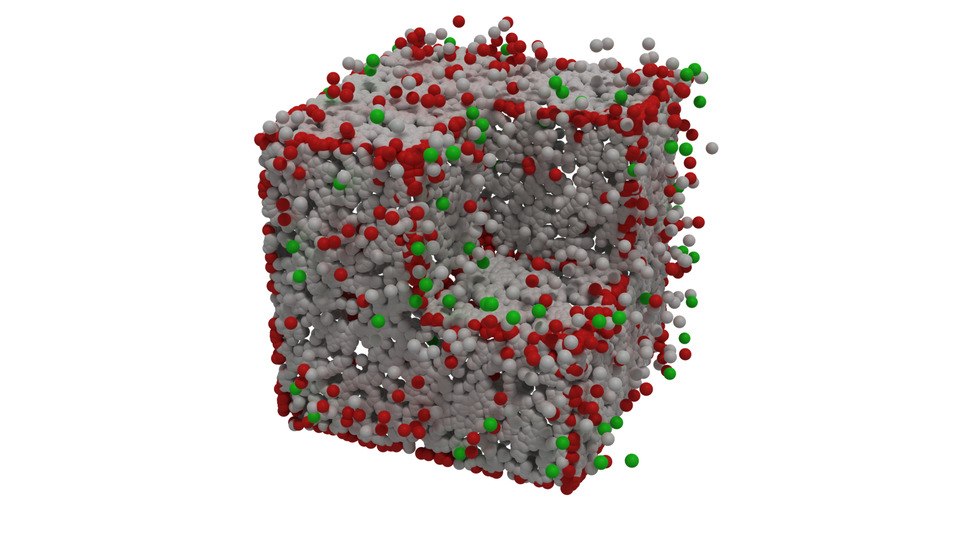}};
        \node[image, right=of i35] (i36) {\includegraphics[trim={220px 0 220px 0},clip,width=0.11\textwidth]{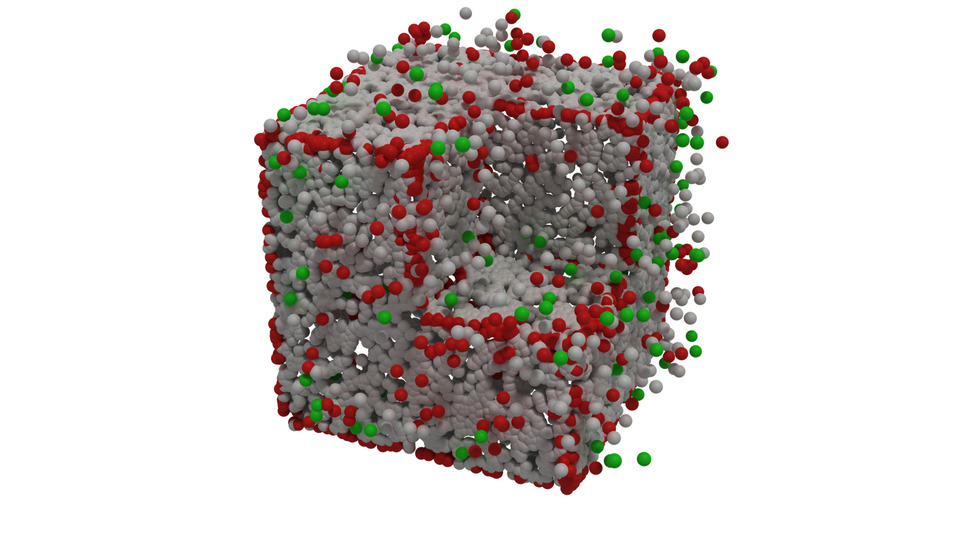}};
        \node[image, right=of i36] (i37) {\includegraphics[trim={220px 0 220px 0},clip,width=0.11\textwidth]{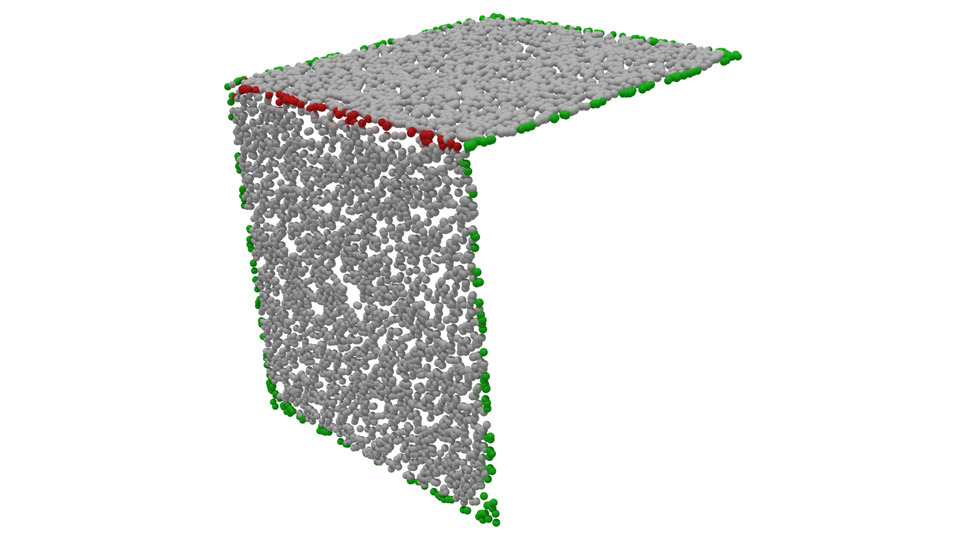}};
        \node[image, right=of i37] (i38) {\includegraphics[trim={220px 0 220px 0},clip,width=0.11\textwidth]{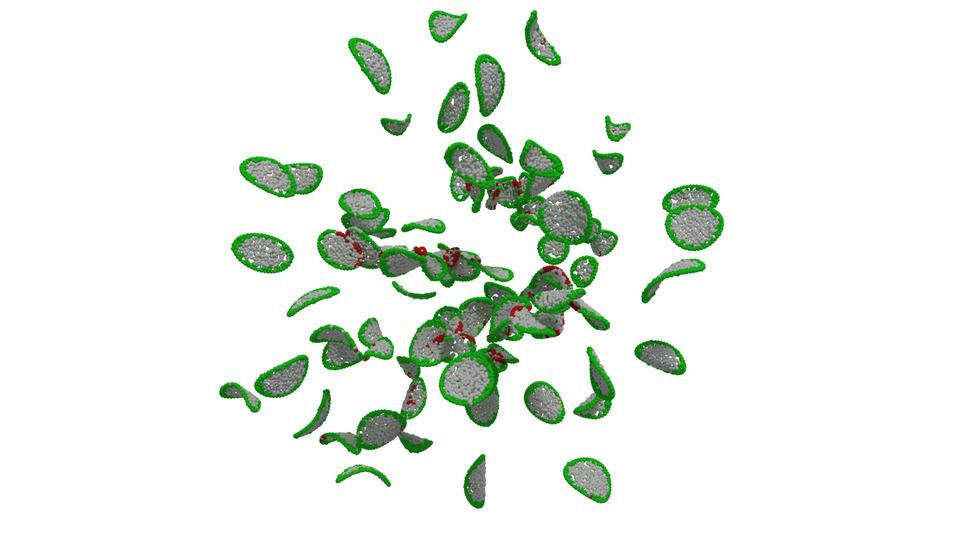}};

        \node[left=of i11, label_side] (ls1) {PCEDNet\\(\emph{Default})};
        \node[left=of i21, label_side] (ls2) {BoundED (Ours)\\(\emph{Default})};
        \node[left=of i31, label_side] (ls3) {BoundED (Ours)\\(\emph{Default++})};
    \end{tikzpicture}
    \caption{\label{fig:qual_eval_default}
        Comparison of the results on the \emph{Default++} evaluation set.
        The dataset used for training is reported in parentheses.
        As first and second row were trained on the \emph{Default} dataset, the respective approches are by design not able to detect boundaries.
    }
\end{figure*}

\begin{figure*}[!t]
    \centering
    \begin{tikzpicture}[image/.style = {inner sep=0mm, outer sep=0pt},
                        label_top/.style = {anchor=south},
                        label_side/.style = {rotate=90, anchor=south, align=center},
                        node distance = 5mm and 2mm]
        \node[image]               (i11) {\includegraphics[trim={200px 0 200px 0},clip,height=3.4cm]{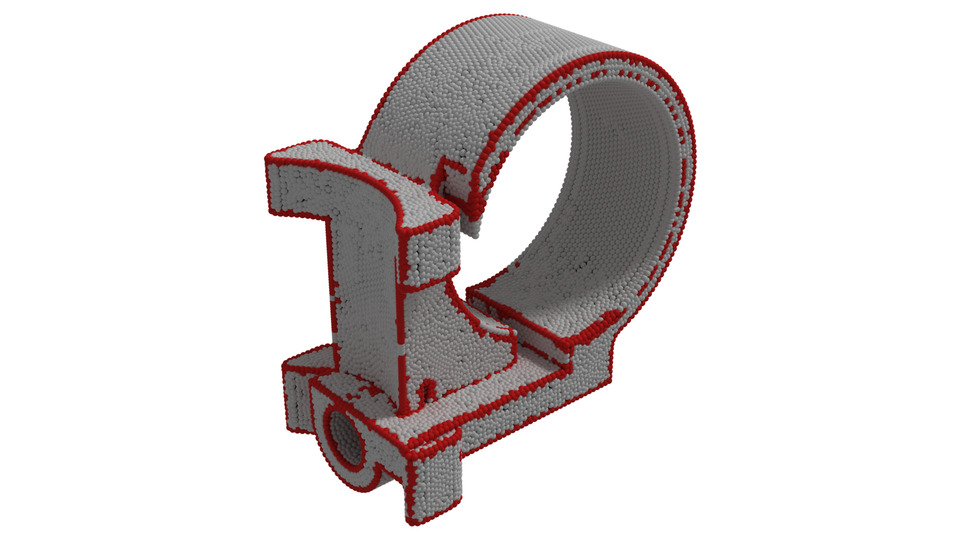}};
        \node[image, right=of i11] (i12) {\includegraphics[trim={200px 0 200px 0},clip,height=3.4cm]{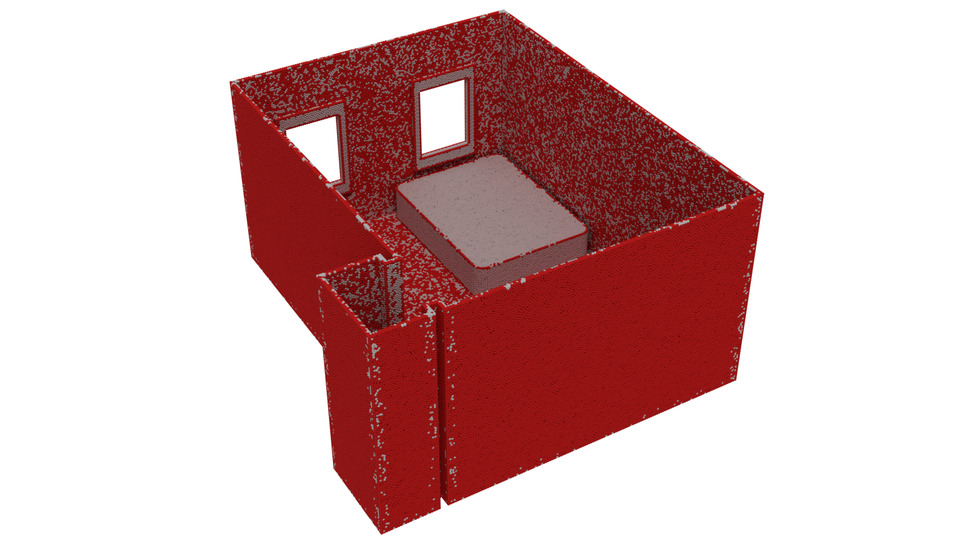}};
        \node[image, right=of i12] (i13) {\includegraphics[trim={250px 0 250px 0},clip,height=3.4cm]{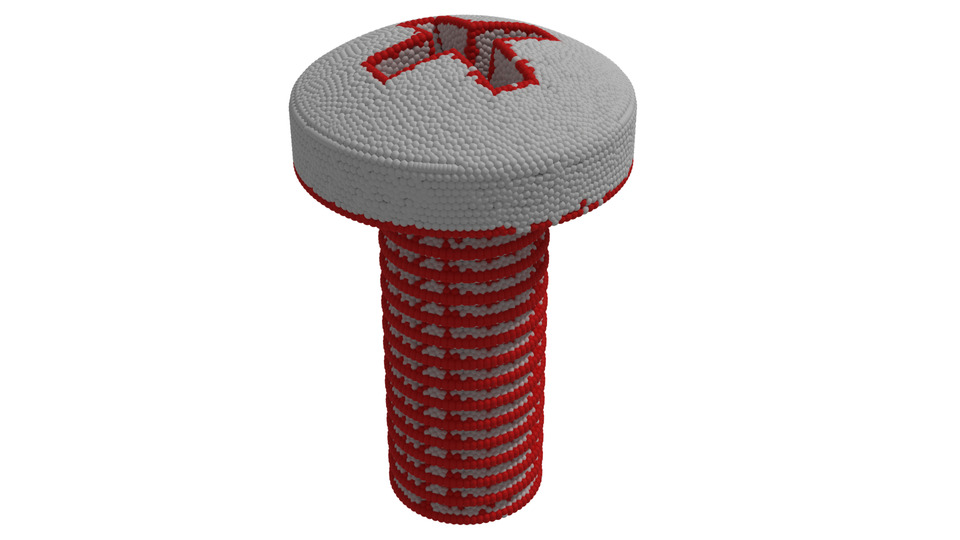}};
        \node[image, right=of i13] (i14) {\includegraphics[trim={100px 0 100px 0},clip,height=3.4cm]{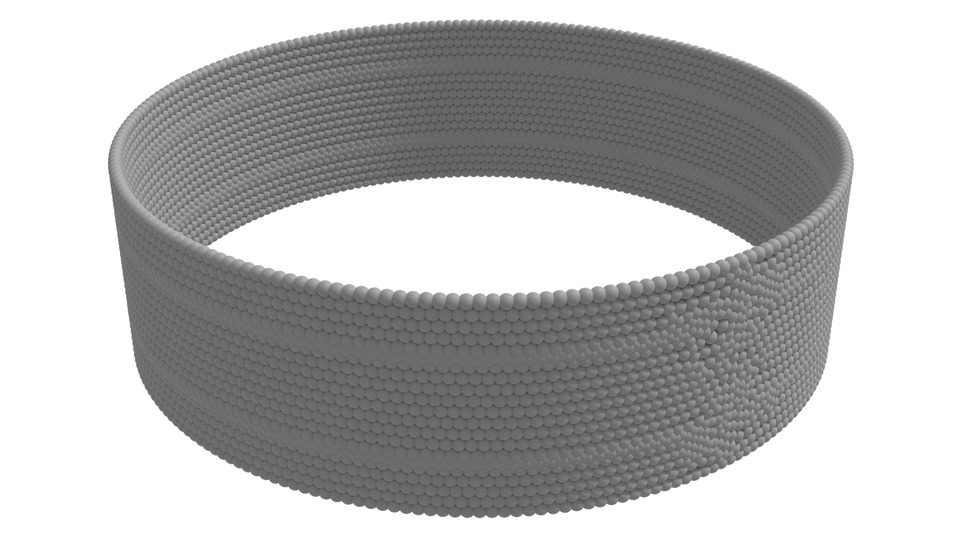}};
        \node[image, below=of i11] (i21) {\includegraphics[trim={200px 0 200px 0},clip,height=3.4cm]{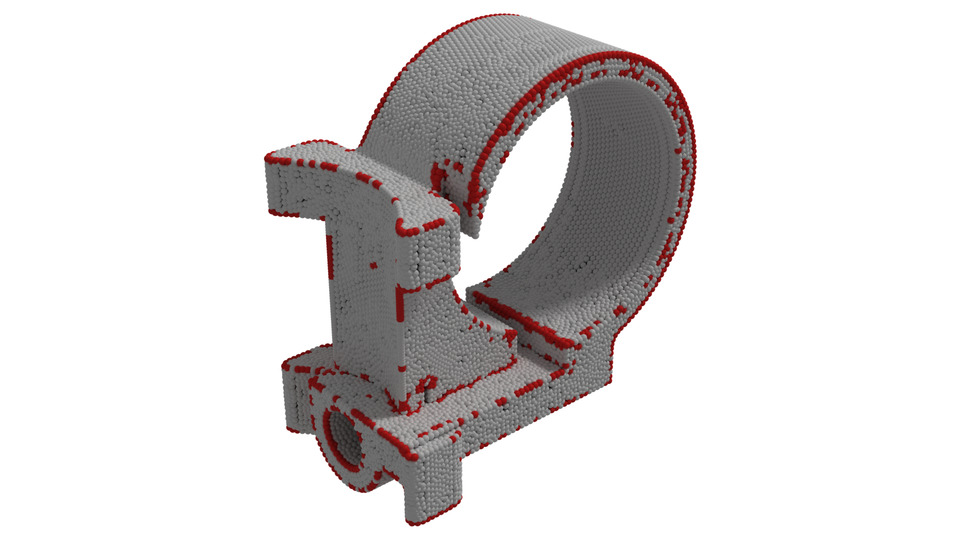}};
        \node[image, below=of i12] (i22) {\includegraphics[trim={200px 0 200px 0},clip,height=3.4cm]{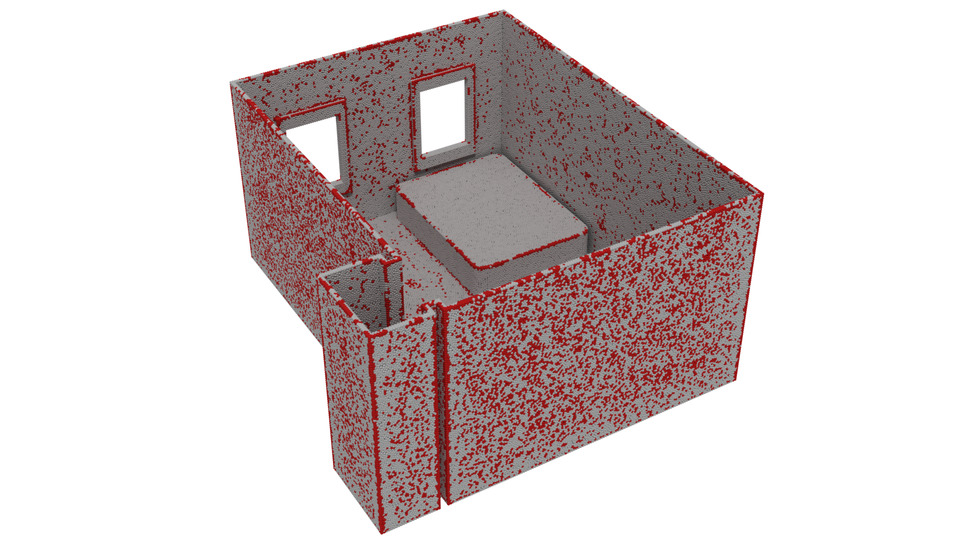}};
        \node[image, below=of i13] (i23) {\includegraphics[trim={250px 0 250px 0},clip,height=3.4cm]{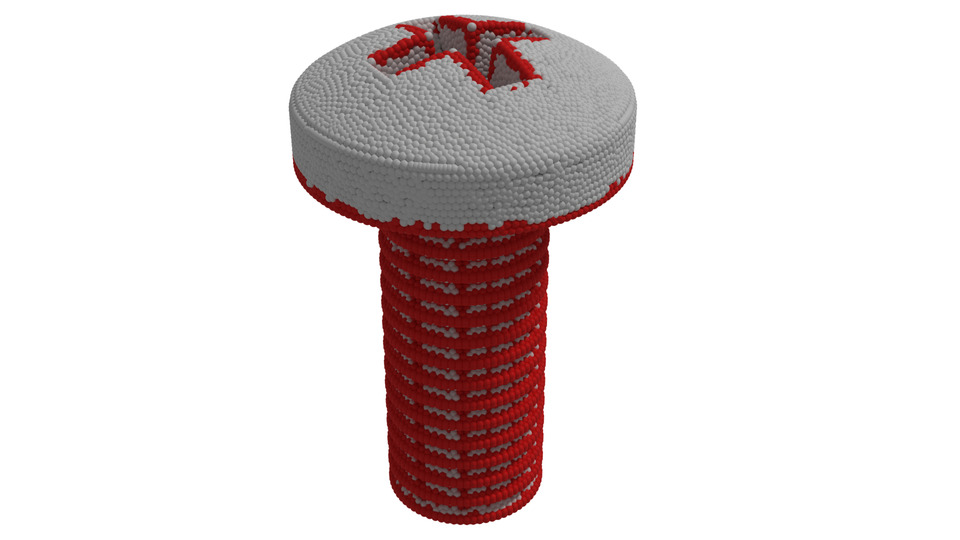}};
        \node[image, below=of i14] (i24) {\includegraphics[trim={100px 0 100px 0},clip,height=3.4cm]{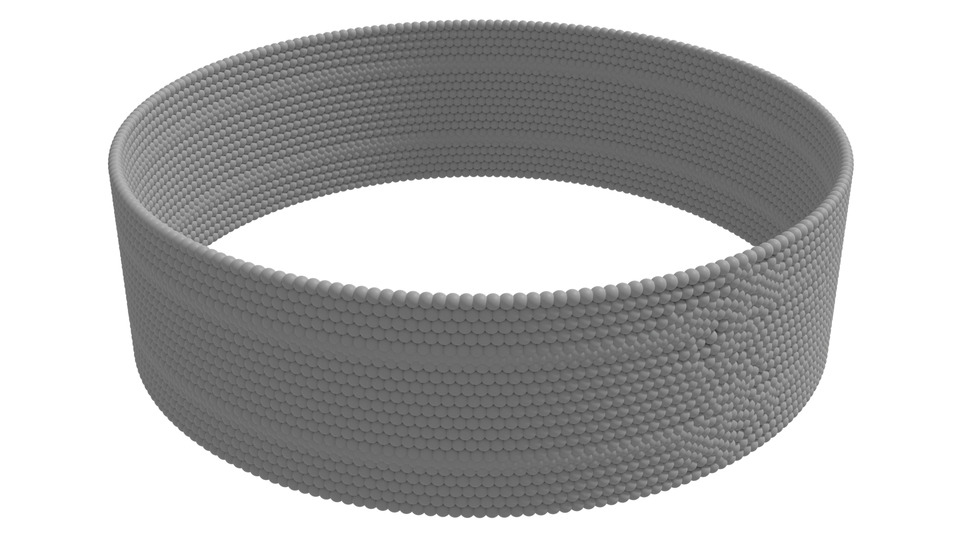}};
        \node[image, below=of i21] (i31) {\includegraphics[trim={200px 0 200px 0},clip,height=3.4cm]{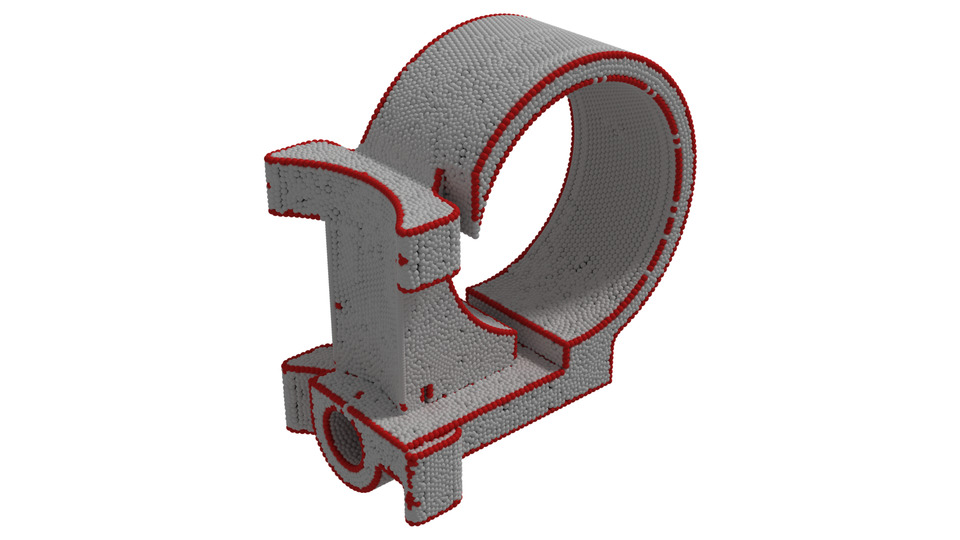}};
        \node[image, below=of i22] (i32) {\includegraphics[trim={200px 0 200px 0},clip,height=3.4cm]{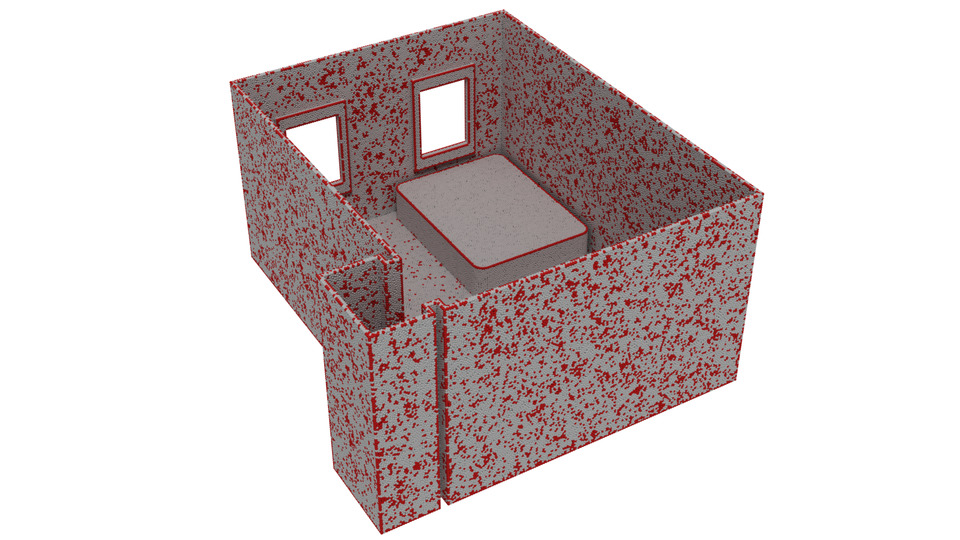}};
        \node[image, below=of i23] (i33) {\includegraphics[trim={250px 0 250px 0},clip,height=3.4cm]{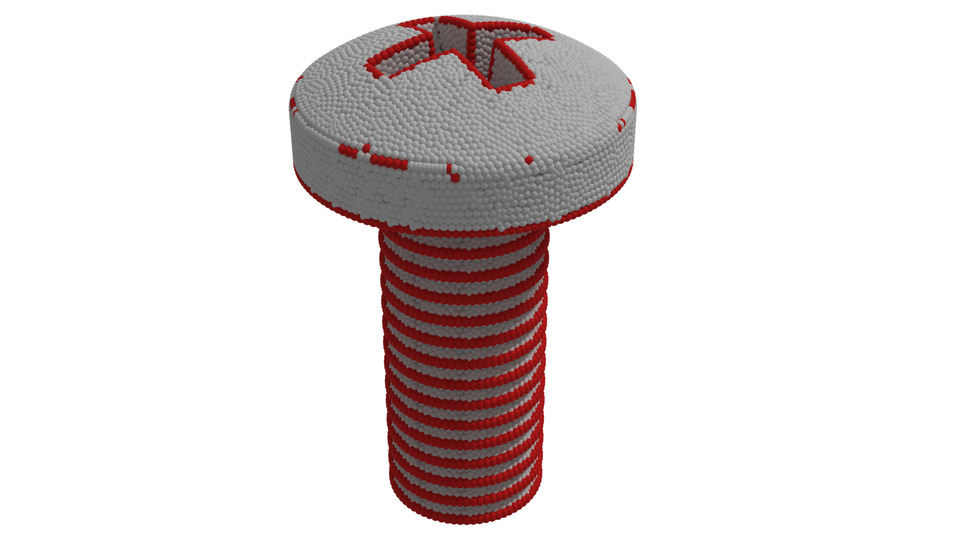}};
        \node[image, below=of i24] (i34) {\includegraphics[trim={100px 0 100px 0},clip,height=3.4cm]{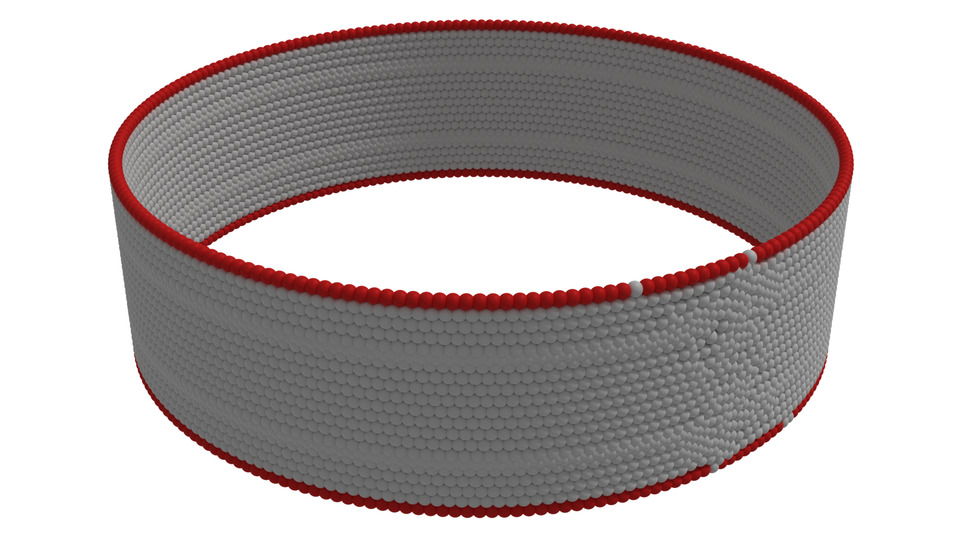}};
        \node[image, below=of i31] (i41) {\includegraphics[trim={200px 0 200px 0},clip,height=3.4cm]{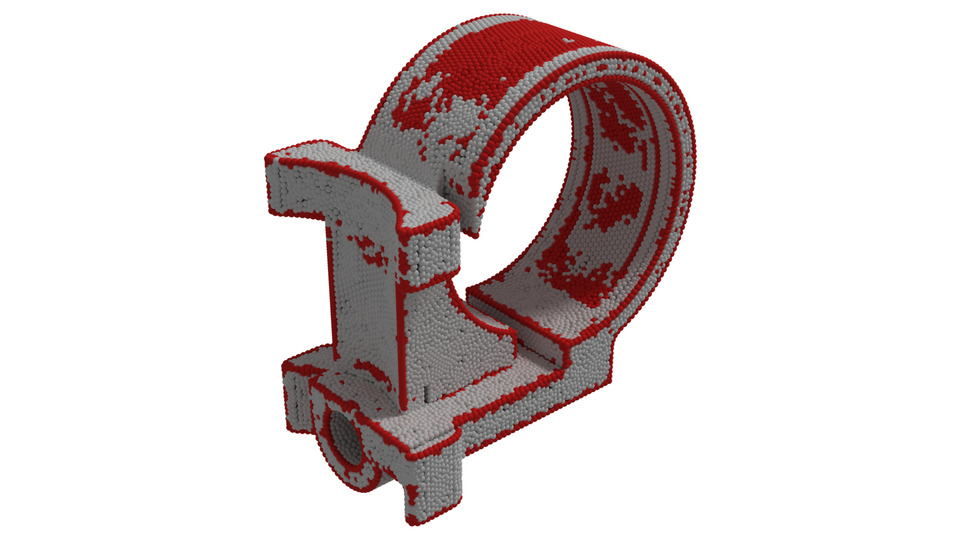}};
        \node[image, below=of i32] (i42) {\includegraphics[trim={200px 0 200px 0},clip,height=3.4cm]{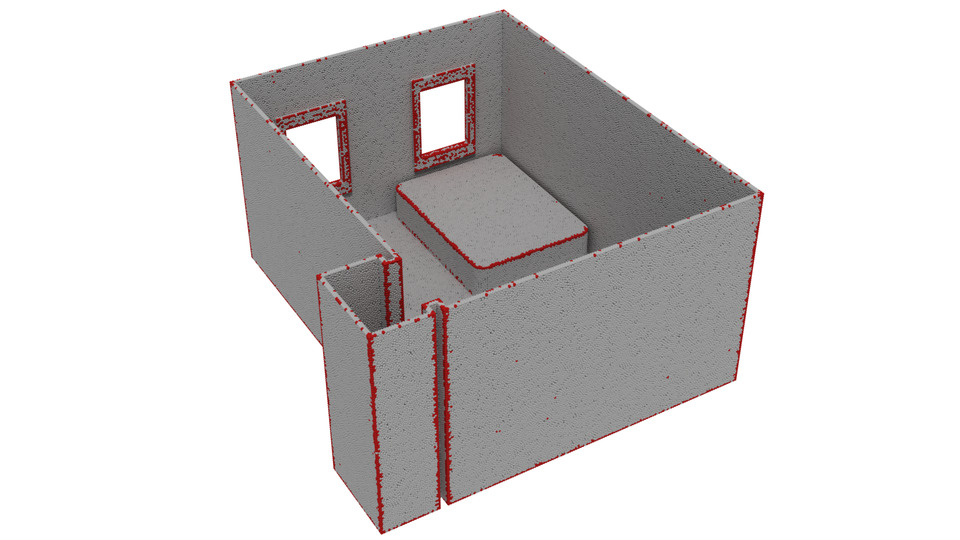}};
        \node[image, below=of i33] (i43) {\includegraphics[trim={250px 0 250px 0},clip,height=3.4cm]{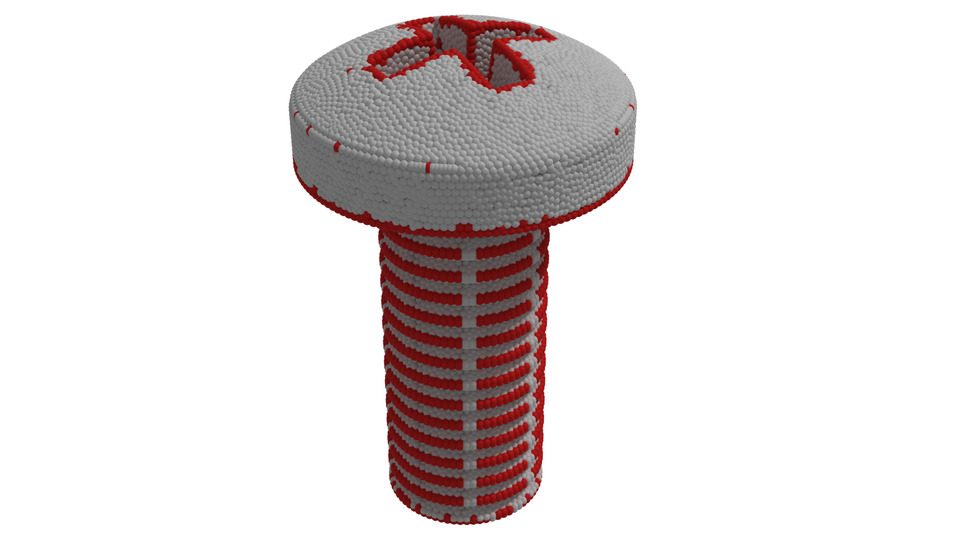}};
        \node[image, below=of i34] (i44) {\includegraphics[trim={100px 0 100px 0},clip,height=3.4cm]{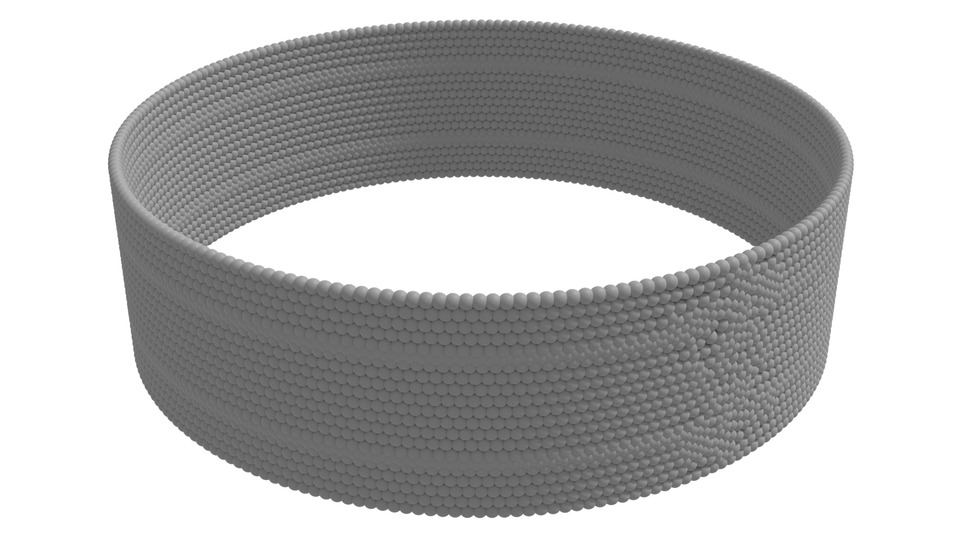}};
        \node[image, below=of i41] (i51) {\includegraphics[trim={200px 0 200px 0},clip,height=3.4cm]{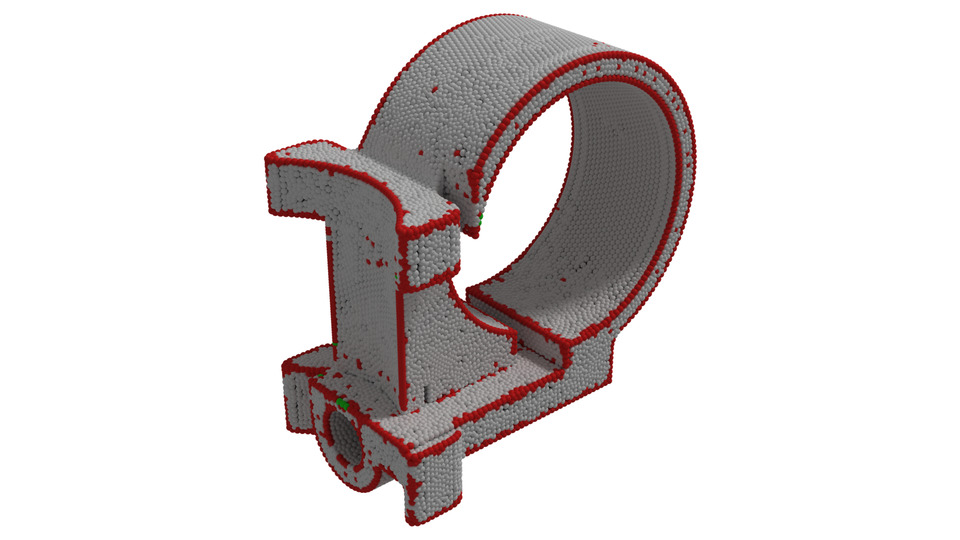}};
        \node[image, below=of i42] (i52) {\includegraphics[trim={200px 0 200px 0},clip,height=3.4cm]{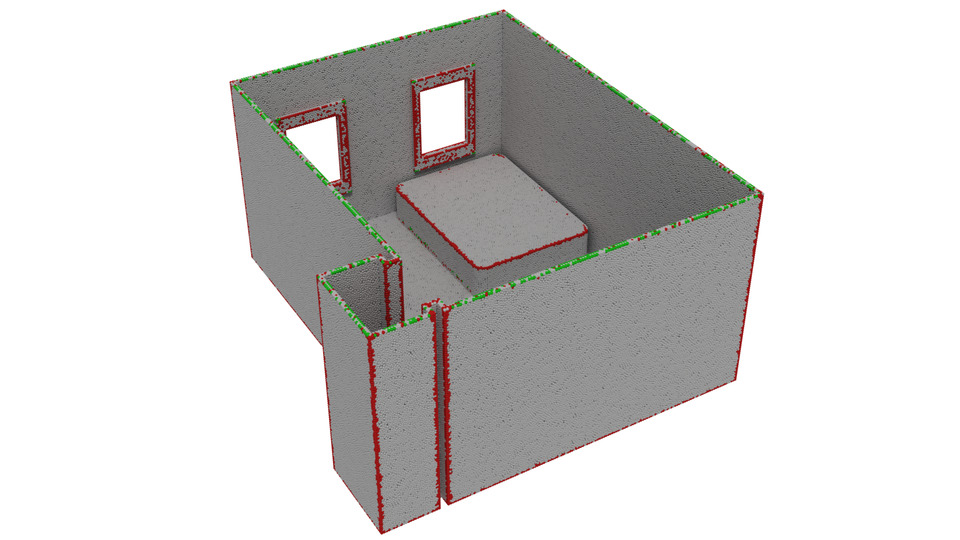}};
        \node[image, below=of i43] (i53) {\includegraphics[trim={250px 0 250px 0},clip,height=3.4cm]{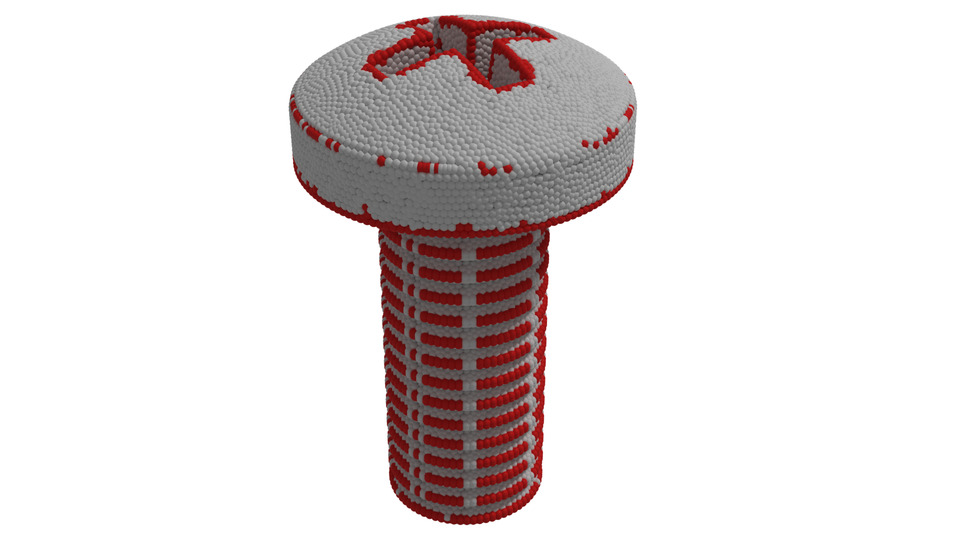}};
        \node[image, below=of i44] (i54) {\includegraphics[trim={100px 0 100px 0},clip,height=3.4cm]{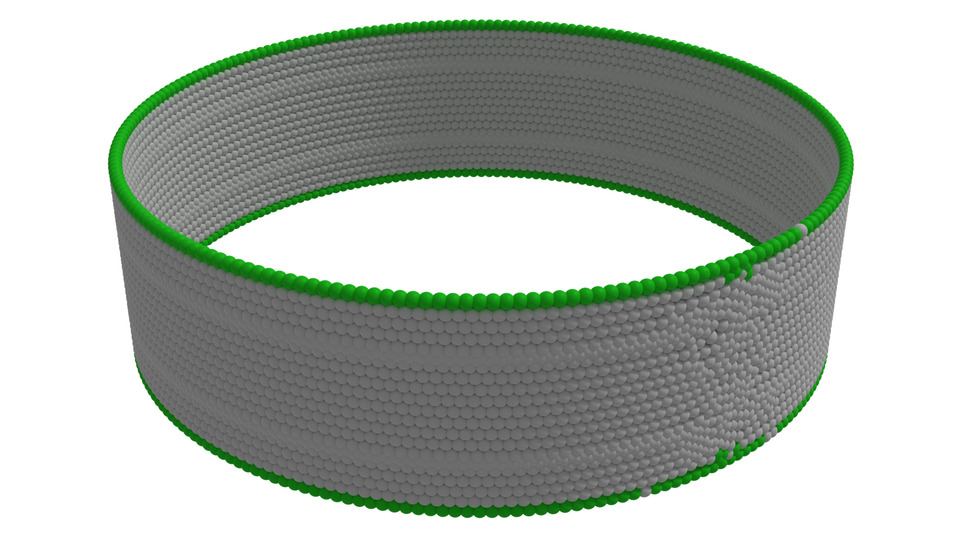}};

        \node[label_top, above=of i11] (lt1) {0027};
        \node[label_top, above=of i12] (lt2) {0059};
        \node[label_top, above=of i13] (lt3) {0117};
        \node[label_top, above=of i14] (lt4) {1222};
        \node[left=of i11, label_side] (ls1) {PCEDNet\\(\emph{ABC})};
        \node[left=of i21, label_side] (ls2) {PCEDNet\\(\emph{Default})};
        \node[left=of i31, label_side] (ls3) {BoundED (Ours)\\(\emph{ABC})};
        \node[left=of i41, label_side] (ls4) {BoundED (Ours)\\(\emph{Default})};
        \node[left=of i51, label_side] (ls5) {BoundED (Ours)\\(\emph{Default++})};

    \end{tikzpicture}
    \caption{\label{fig:qual_eval_abc}
        Comparison of PCEDNet and BoundED trained on three different datasets and evaluated on four different models from the \emph{ABC} evaluation dataset.
        The dataset used for training the respective approach is given in parentheses.
        Algorithms trained on \emph{Default} or \emph{ABC} are not able to detect boundaries by design.
    }
\end{figure*}
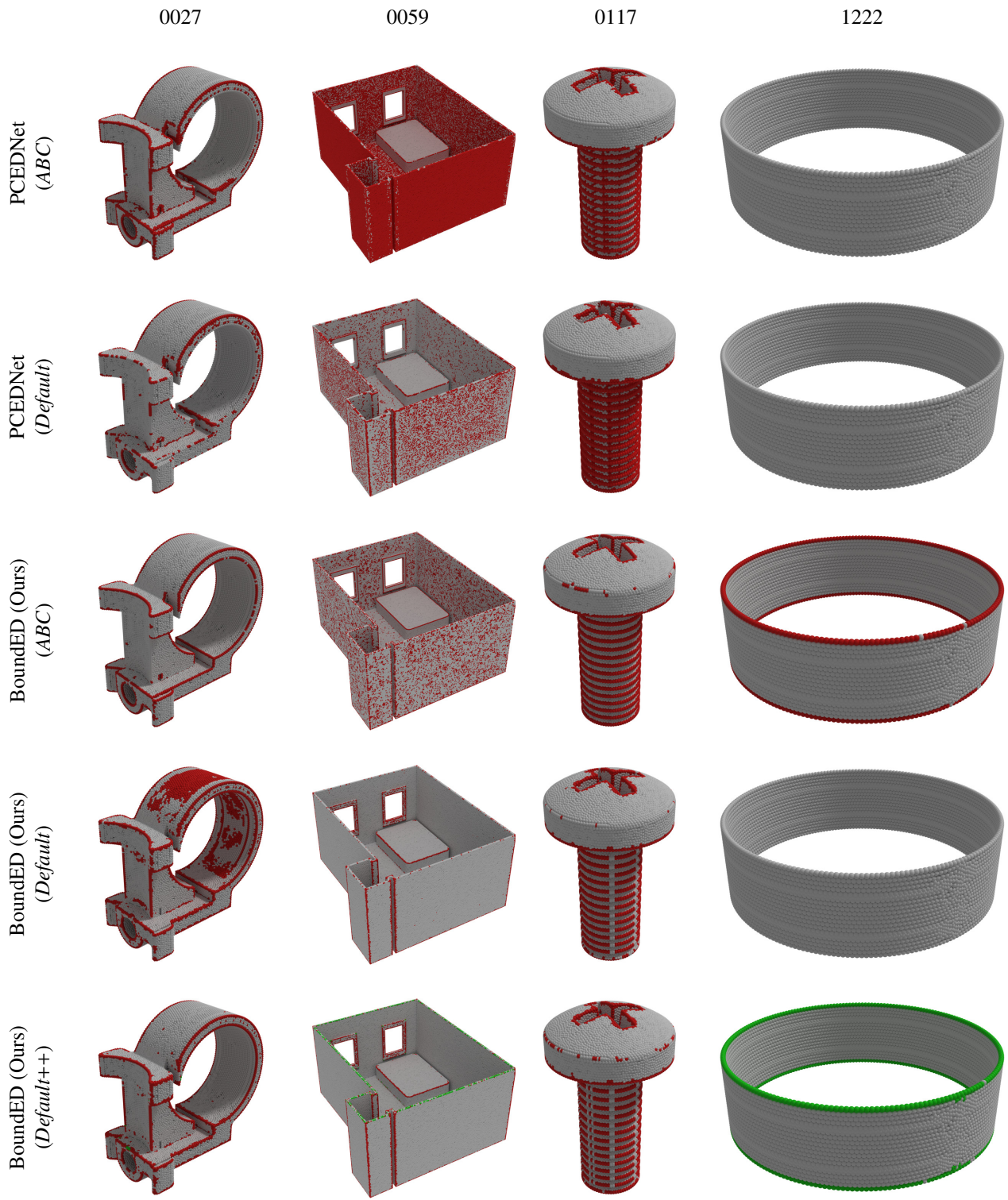

\begin{figure*}[!t]
    \centering
    \begin{tikzpicture}[overview/.style = {inner sep=0mm, outer sep=3mm},
                        zoom/.style = {inner sep=0mm, outer sep=0pt, clip, rounded corners={0.3\linewidth/2.400*0.2}},
                        frame/.style = {inner sep=0mm, outer sep=0pt, rectangle, line width=0.5mm, rounded corners={0.3\linewidth/2.400*0.2}},
                        marker/.style = {inner sep=0mm, outer sep=0pt, rectangle, line width=0.25mm, rounded corners={0.9\linewidth/21.200*0.2}, minimum width={0.9\linewidth/21.200*2.400}, minimum height={0.9\linewidth/21.200*1.600}},
                        label/.style = {rotate=90, anchor=south},
                        node distance = 2mm and 3mm]

        \node[overview] (overview1) at (0, 0) {\includegraphics[width=0.9\linewidth]{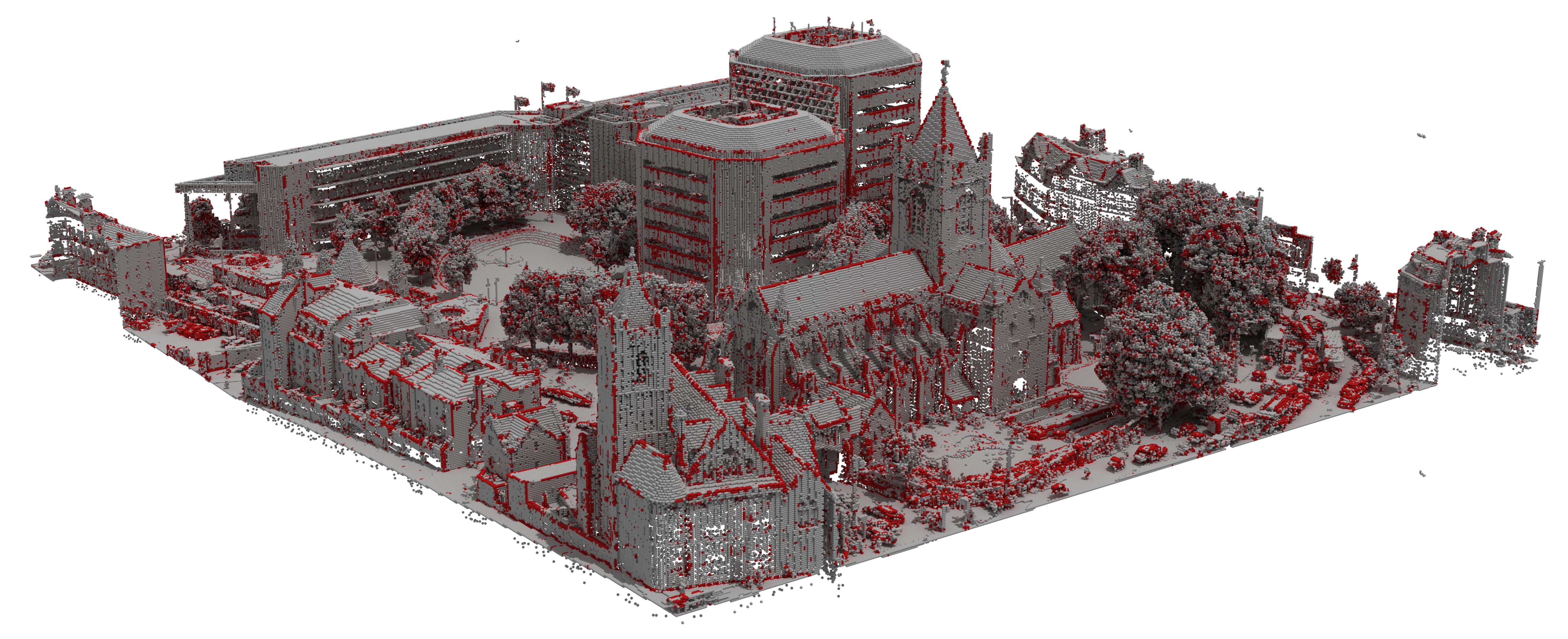}};
        \begin{scope}[shift=(overview1.south west),
                      x=(overview1.south east),
                      y=(overview1.north west)]
            \node[marker, draw=green] at ({9.425/21.200}, {2.812/8.640}) {};
            \node[marker, draw=red] at ({5.432/21.200}, {6.200/8.640}) {};
            \node[marker, draw=blue] at ({12.766/21.200}, {4.401/8.640}) {};
        \end{scope}
        \node[label] at (overview1.west) {PCEDNet (\emph{Default})};

        \node[zoom, below=of overview1] (1_zoom2) {\includegraphics[width=0.3\linewidth]{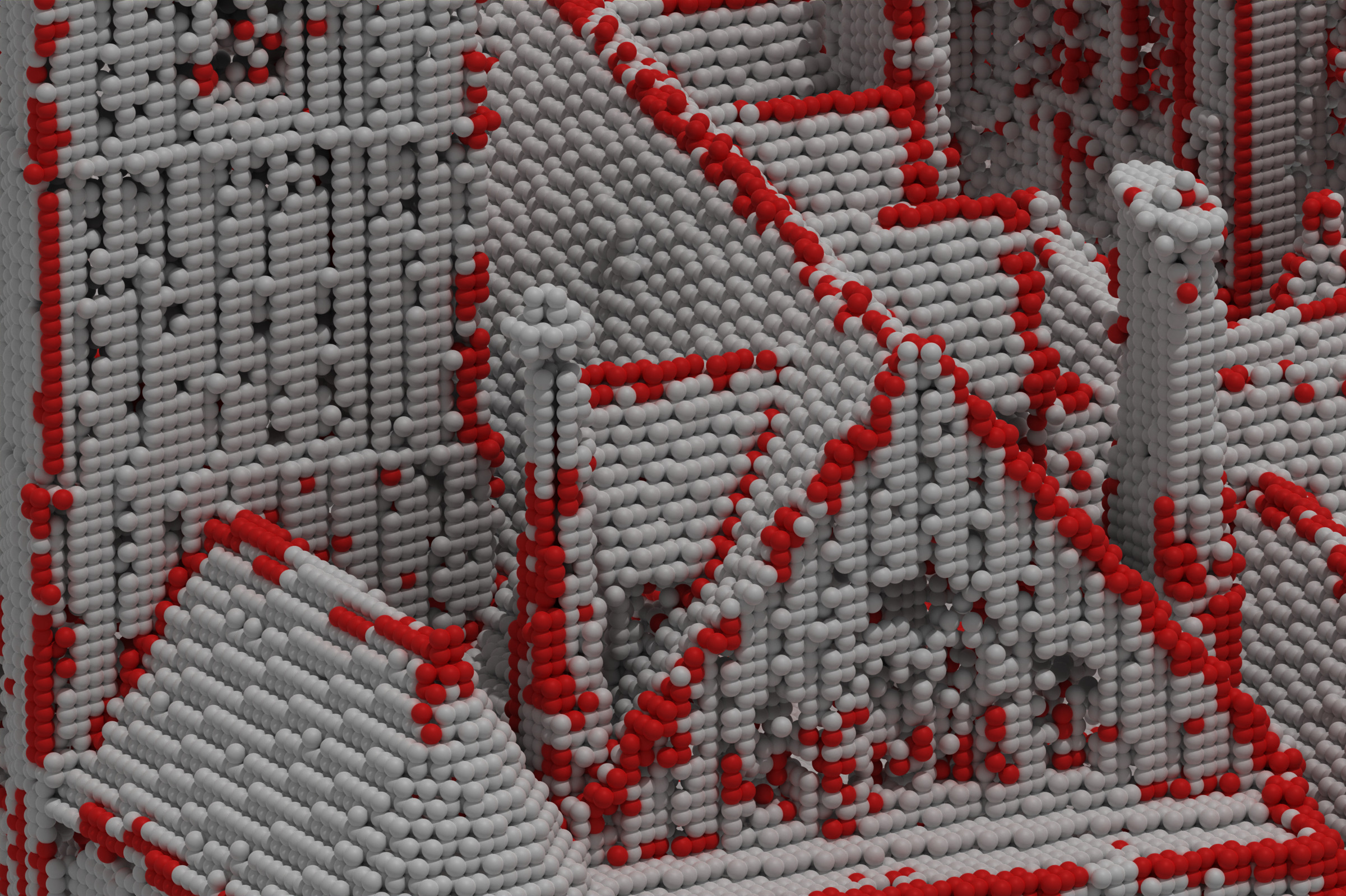}};
        \node[frame, draw=green, fit=(1_zoom2)] {};
        \node[zoom, left=of 1_zoom2] (1_zoom3) {\includegraphics[width=0.3\linewidth]{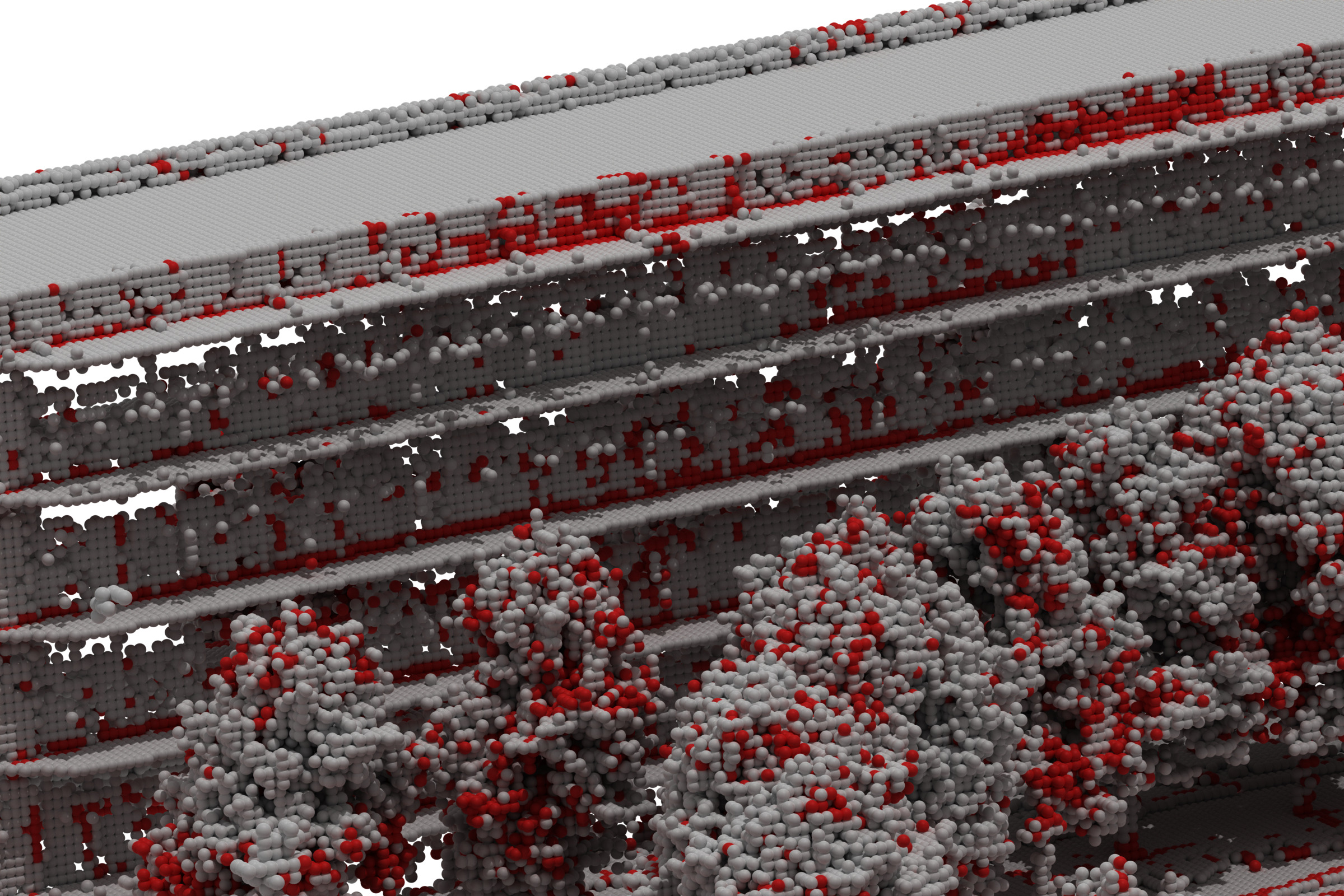}};
        \node[frame, draw=red, fit=(1_zoom3)] {};
        \node[zoom, right=of 1_zoom2] (1_zoom1) {\includegraphics[width=0.3\linewidth]{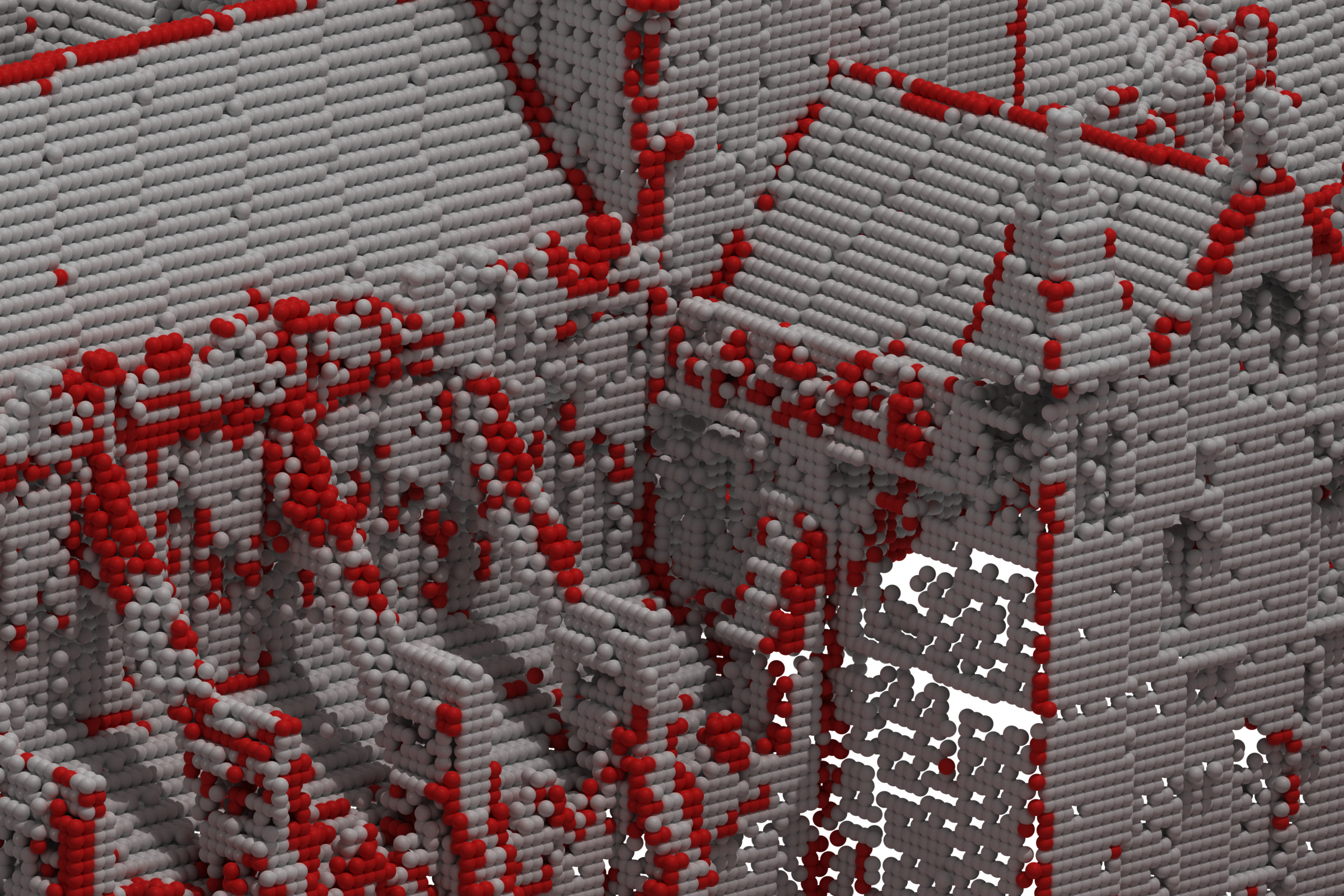}};
        \node[frame, draw=blue, fit=(1_zoom1)] {};

        \node[zoom, below=of 1_zoom2] (2_zoom2) {\includegraphics[width=0.3\linewidth]{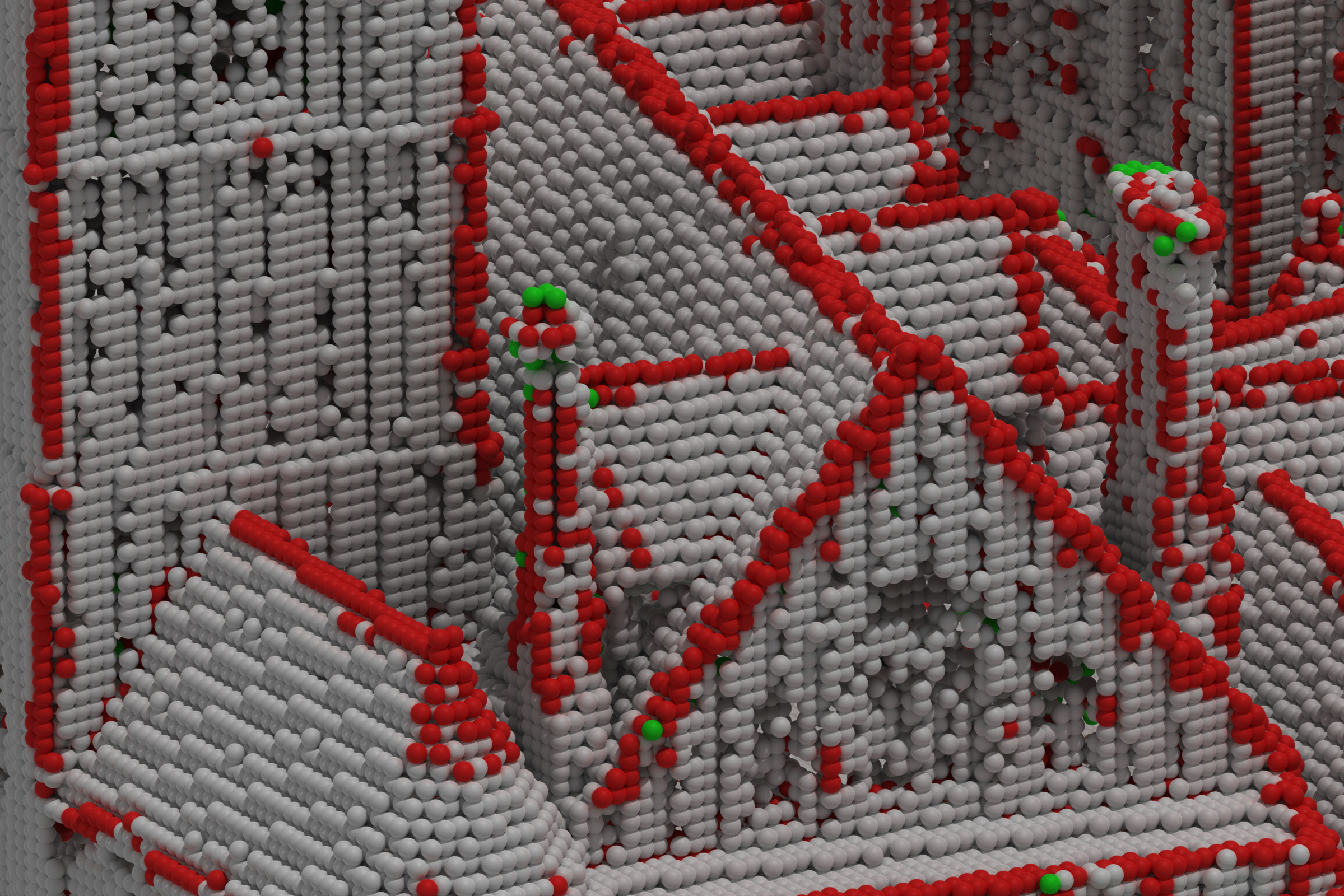}};
        \node[frame, draw=green, fit=(2_zoom2)] {};
        \node[zoom, left=of 2_zoom2] (2_zoom3) {\includegraphics[width=0.3\linewidth]{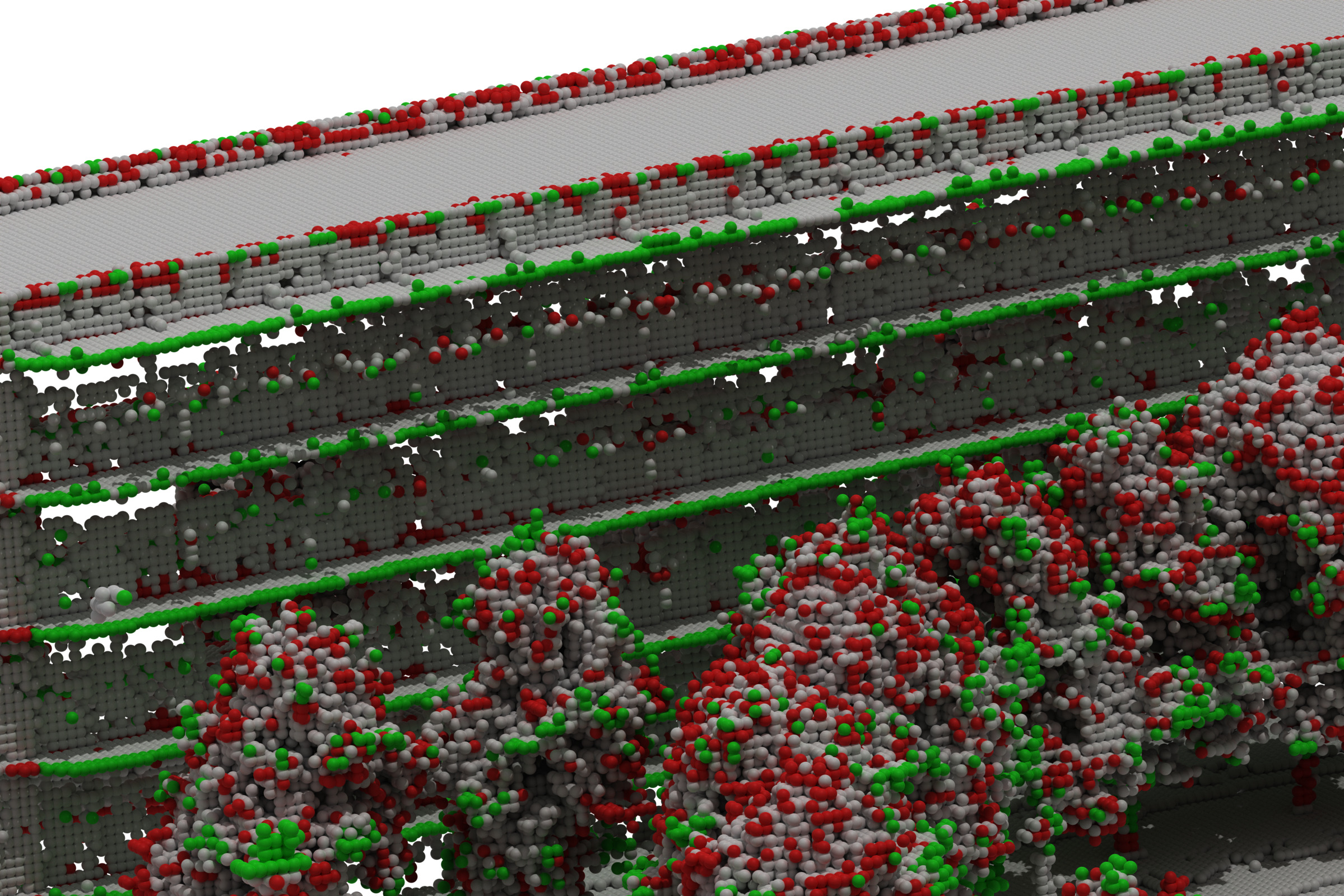}};
        \node[frame, draw=red, fit=(2_zoom3)] {};
        \node[zoom, right=of 2_zoom2] (2_zoom1) {\includegraphics[width=0.3\linewidth]{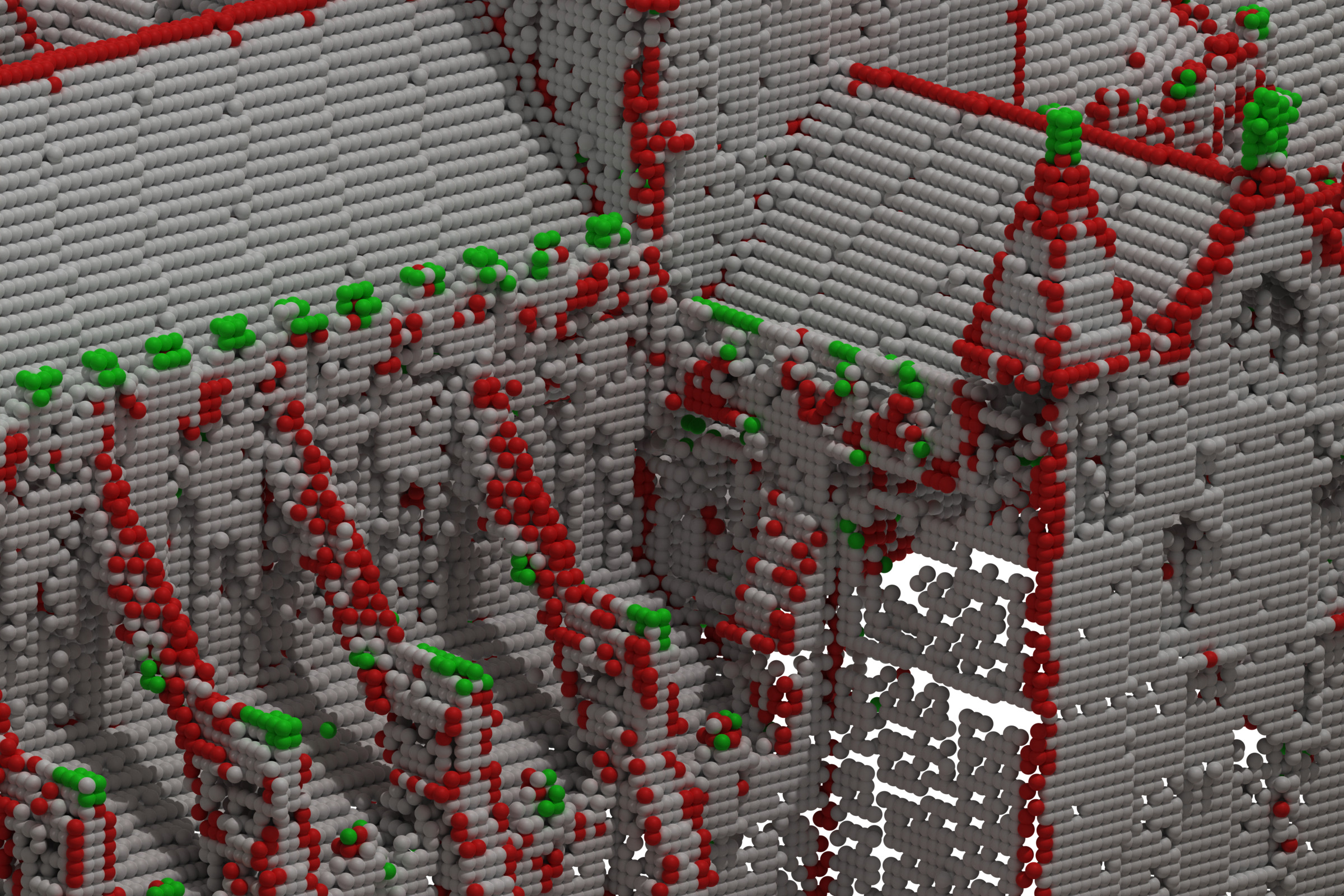}};
        \node[frame, draw=blue, fit=(2_zoom1)] {};

        \node[overview, below=of 2_zoom2] (overview2) {\includegraphics[width=0.9\linewidth]{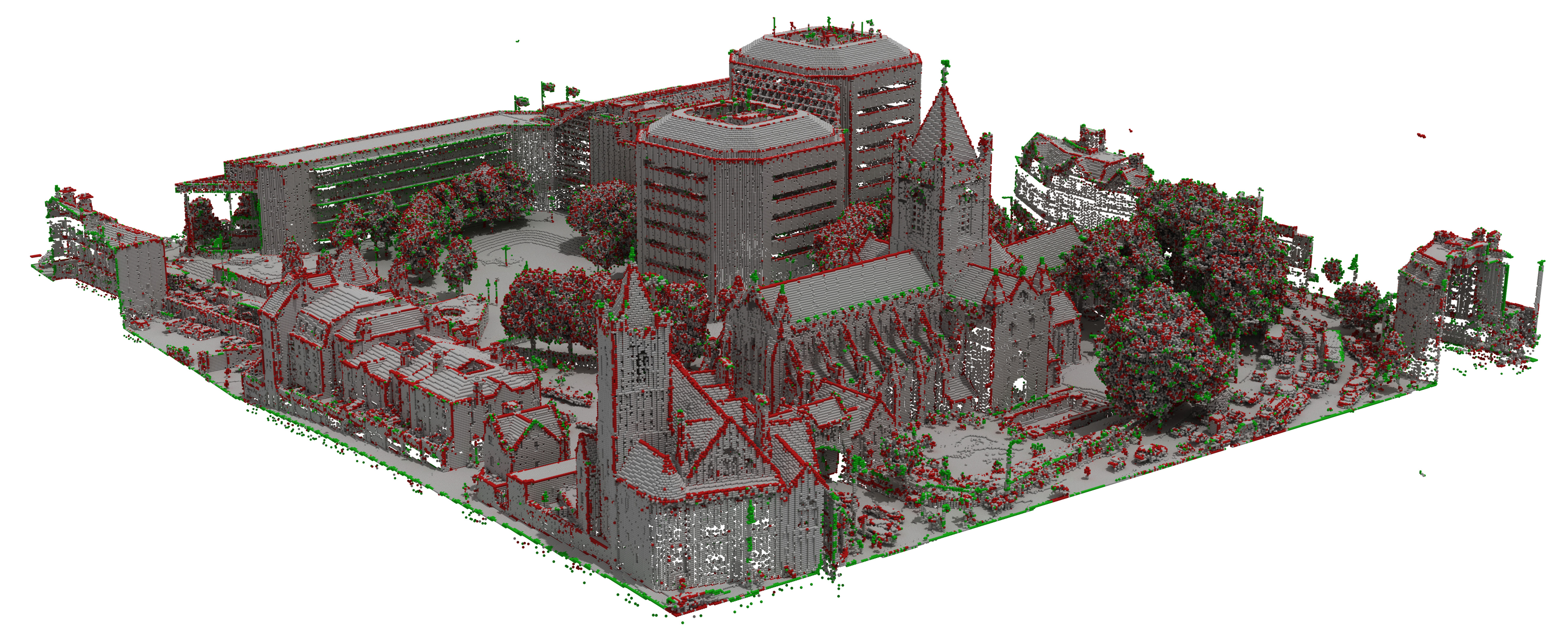}};
        \begin{scope}[shift=(overview2.south west),
                      x=(overview2.south east),
                      y=(overview2.north west)]
            \node[marker, draw=green] at ({9.425/21.200}, {2.812/8.640}) {};
            \node[marker, draw=red] at ({5.432/21.200}, {6.200/8.640}) {};
            \node[marker, draw=blue] at ({12.766/21.200}, {4.401/8.640}) {};
        \end{scope}
        \node[label] at (overview2.west) {BoundED (Ours) (\emph{Default++})};

    \end{tikzpicture}
    \caption{\label{fig:qual_eval_christ_church}
        Classification result of PCEDNet trained on \emph{Default} (top) and BoundED trained on \emph{Default++} (bottom) for the mid-sized (1.9 million points) scanned \emph{christ\_church} point cloud.
        Three different zoomed parts are depicted for direct comparison (middle).
    }
\end{figure*}
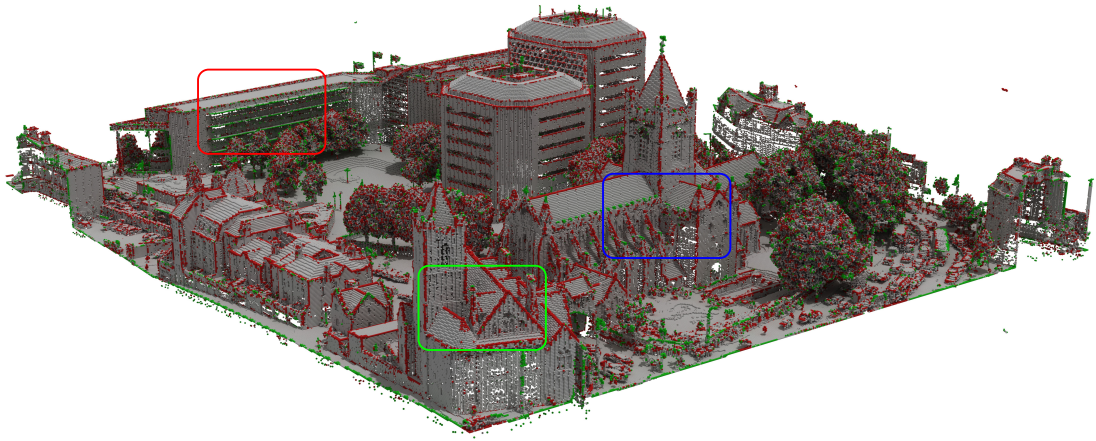

\begin{figure*}[!t]
    \centering
    \begin{tikzpicture}[image/.style = {inner sep=0mm, outer sep=0pt},
                        label/.style = {anchor=south},
                        node distance = 1mm and 2mm]
        \node[image] (i11) at (0, 0)     {\includegraphics[trim={0 150px 0 150px},clip,width=0.4\textwidth]{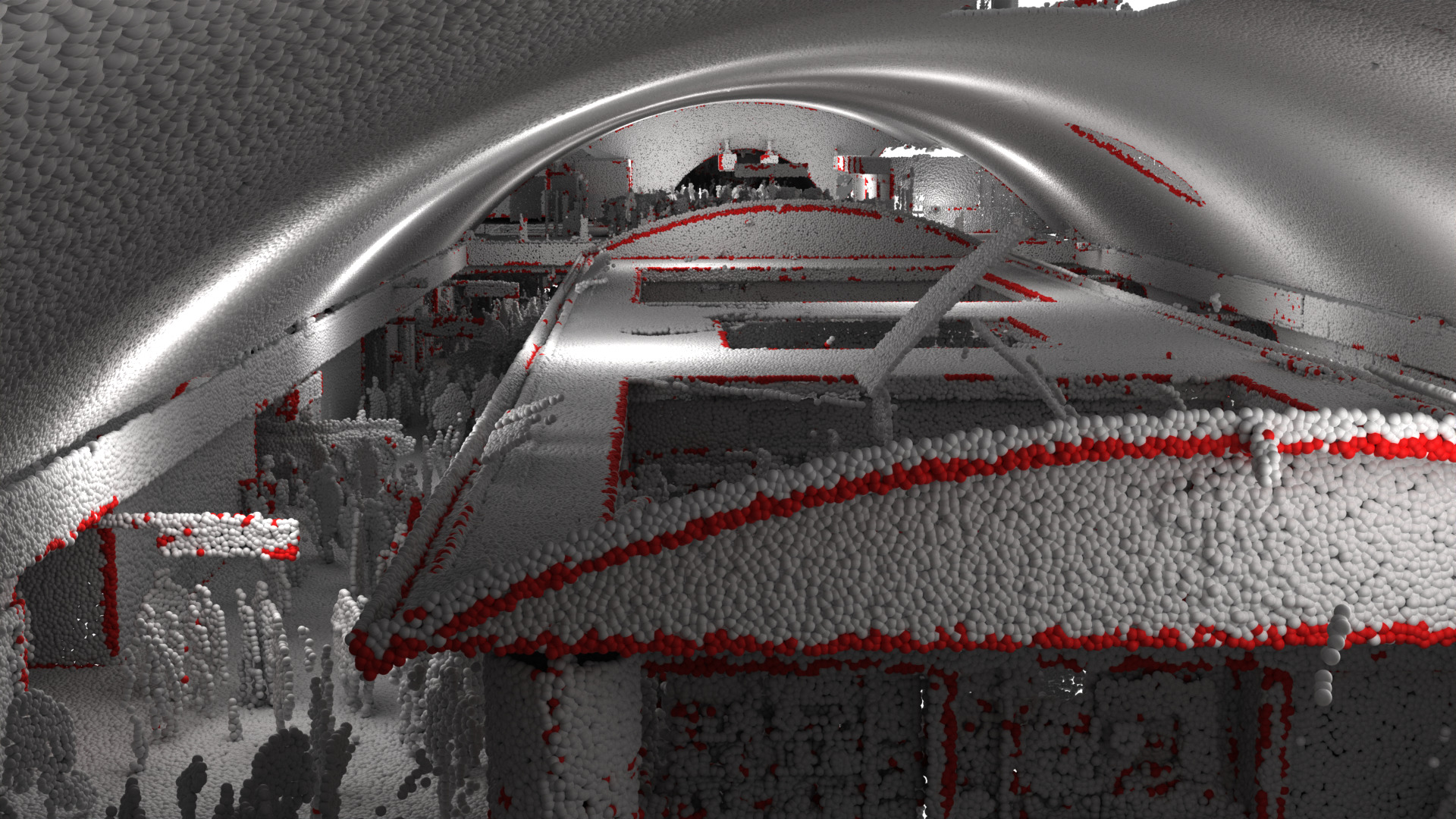}};
        \node[image, right=of i11] (i12) {\includegraphics[trim={0 150px 0 150px},clip,width=0.4\textwidth]{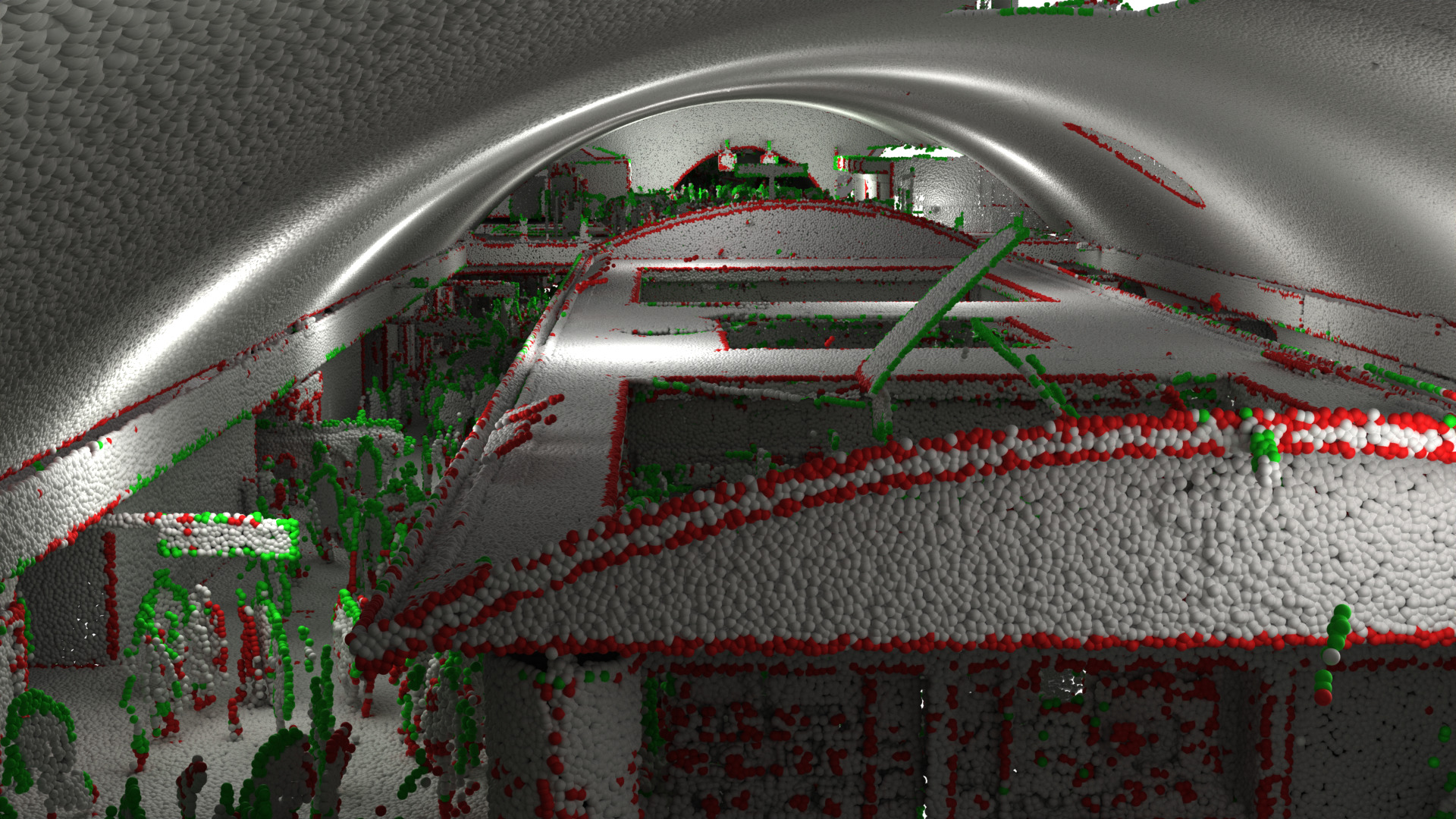}};
        \node[image, below=of i11] (i21) {\includegraphics[trim={0 150px 0 150px},clip,width=0.4\textwidth]{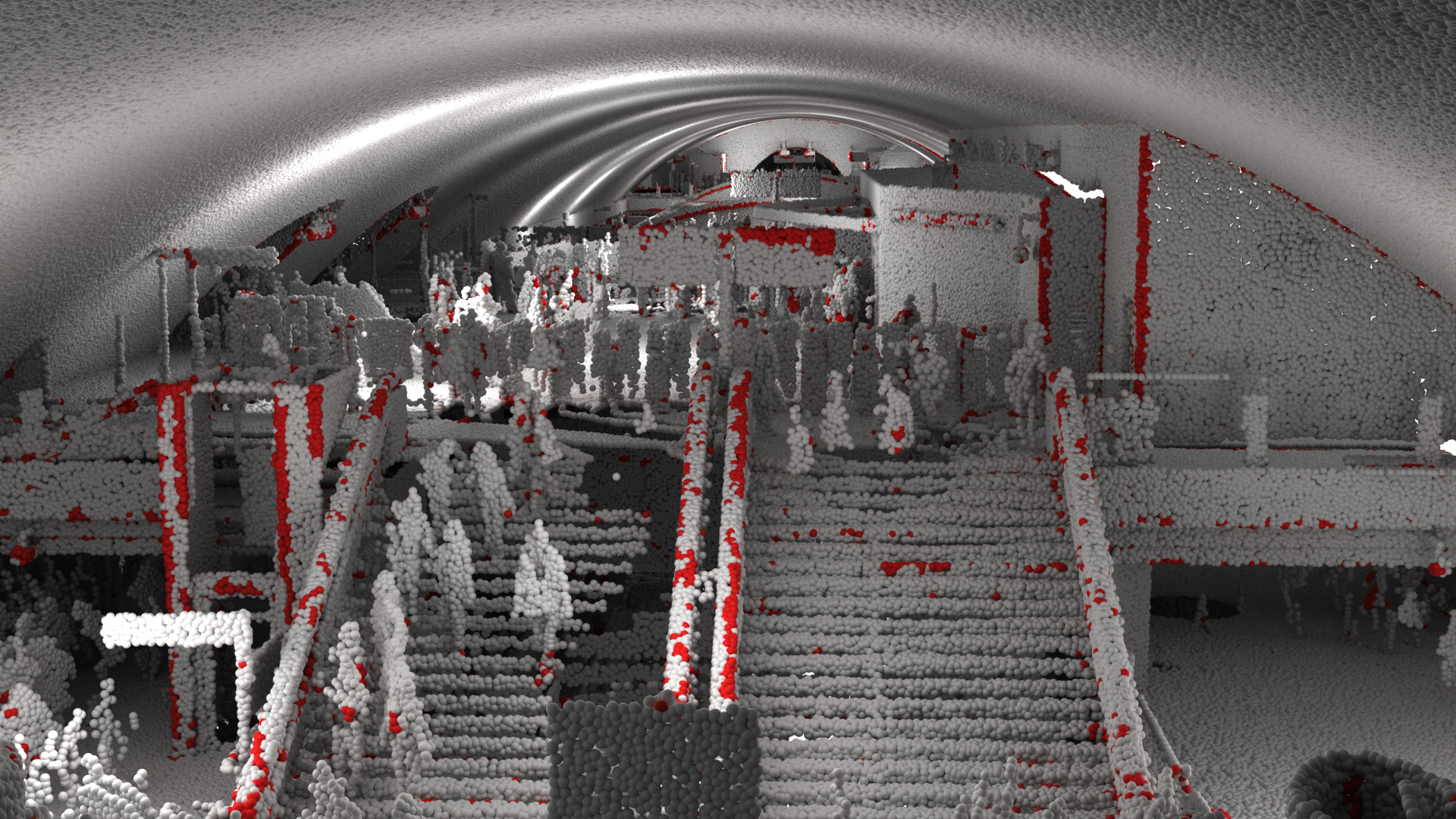}};
        \node[image, below=of i12] (i22) {\includegraphics[trim={0 150px 0 150px},clip,width=0.4\textwidth]{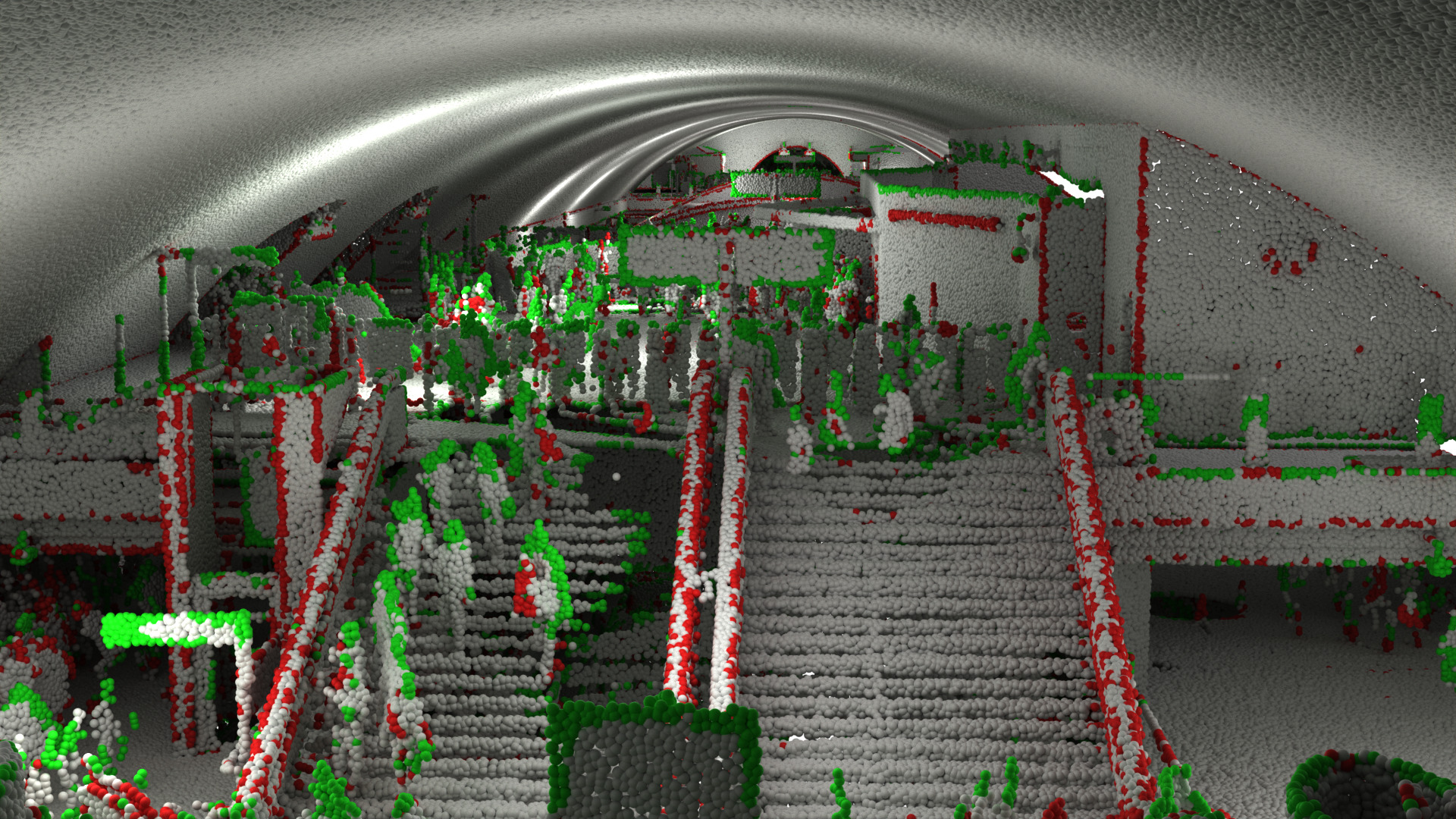}};
        \node[image, below=of i21] (i31) {\includegraphics[trim={0 150px 0 150px},clip,width=0.4\textwidth]{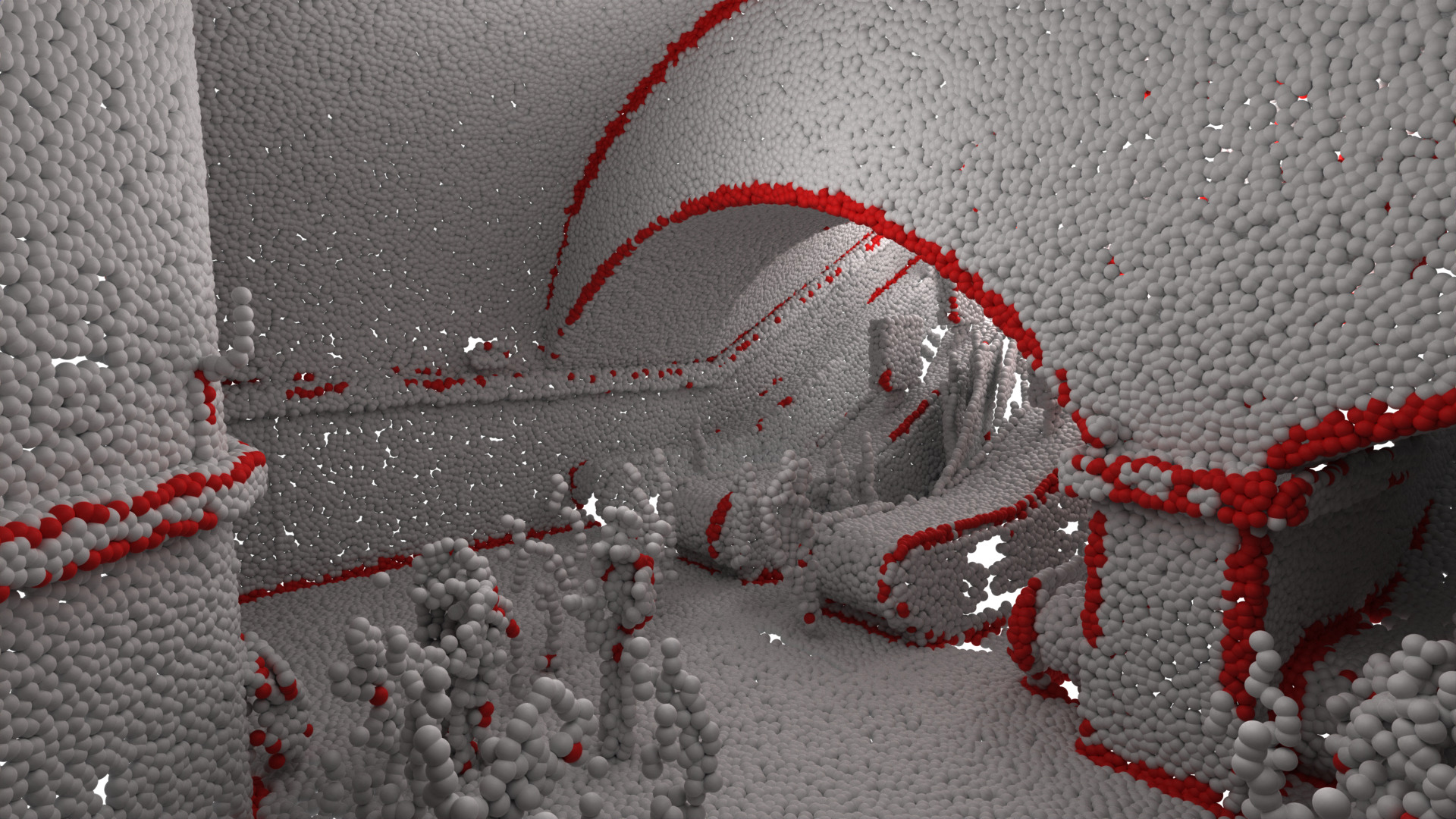}};
        \node[image, below=of i22] (i32) {\includegraphics[trim={0 150px 0 150px},clip,width=0.4\textwidth]{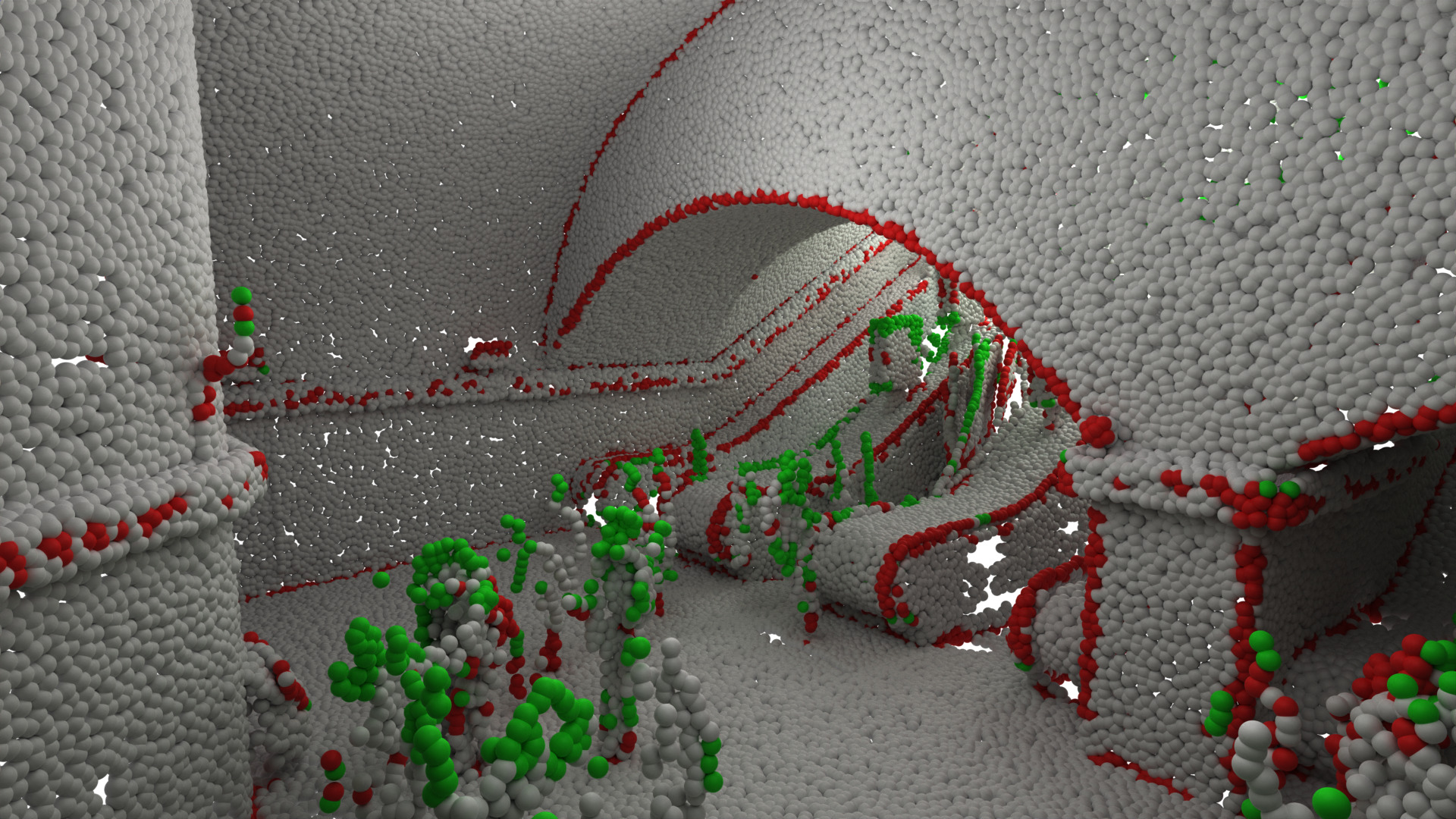}};

        \node[label, above= of i11] (l1) {PCEDNet (\emph{Default})};
        \node[label, above= of i12] (l2) {BoundED (Ours) (\emph{Default++})};
    \end{tikzpicture}
    \caption{\label{fig:qual_eval_station}
        Classification result of PCEDNet trained on \emph{Default} (left) and BoundED trained on \emph{Default++} (right) for the large (12.5 million points) scanned \emph{train\_station} point cloud.
    }
\end{figure*}

\begin{figure*}[!t]
    \centering
    \begin{tikzpicture}[mainimg/.style = {inner sep=0mm, outer sep=0pt},
                        zoom/.style = {inner sep=0mm, outer sep=0pt, clip, rounded corners={0.2\linewidth/2.000*0.185}},
                        frame/.style = {inner sep=0mm, outer sep=0pt, rectangle, line width=0.5mm, rounded corners={0.2\linewidth/2.000*0.185}},
                        marker1/.style = {inner sep=0mm, outer sep=0pt, rectangle, line width=0.25mm, rounded corners={0.65\linewidth/16.000*0.185}, minimum width={0.65\linewidth/16.000*2.000}, minimum height={0.65\linewidth/16.000*2.000}},
                        marker2/.style = {inner sep=0mm, outer sep=0pt, rectangle, line width=0.25mm, rounded corners={0.3\linewidth/8.000*0.185}, minimum width={0.3\linewidth/8.000*2.000}, minimum height={0.3\linewidth/8.000*2.000}},
                        label/.style = {rotate=90, anchor=south},
                        node distance = 2mm and 2mm]

        \node[mainimg] (p1) at (0, 0) {\includegraphics[width=0.58\linewidth]{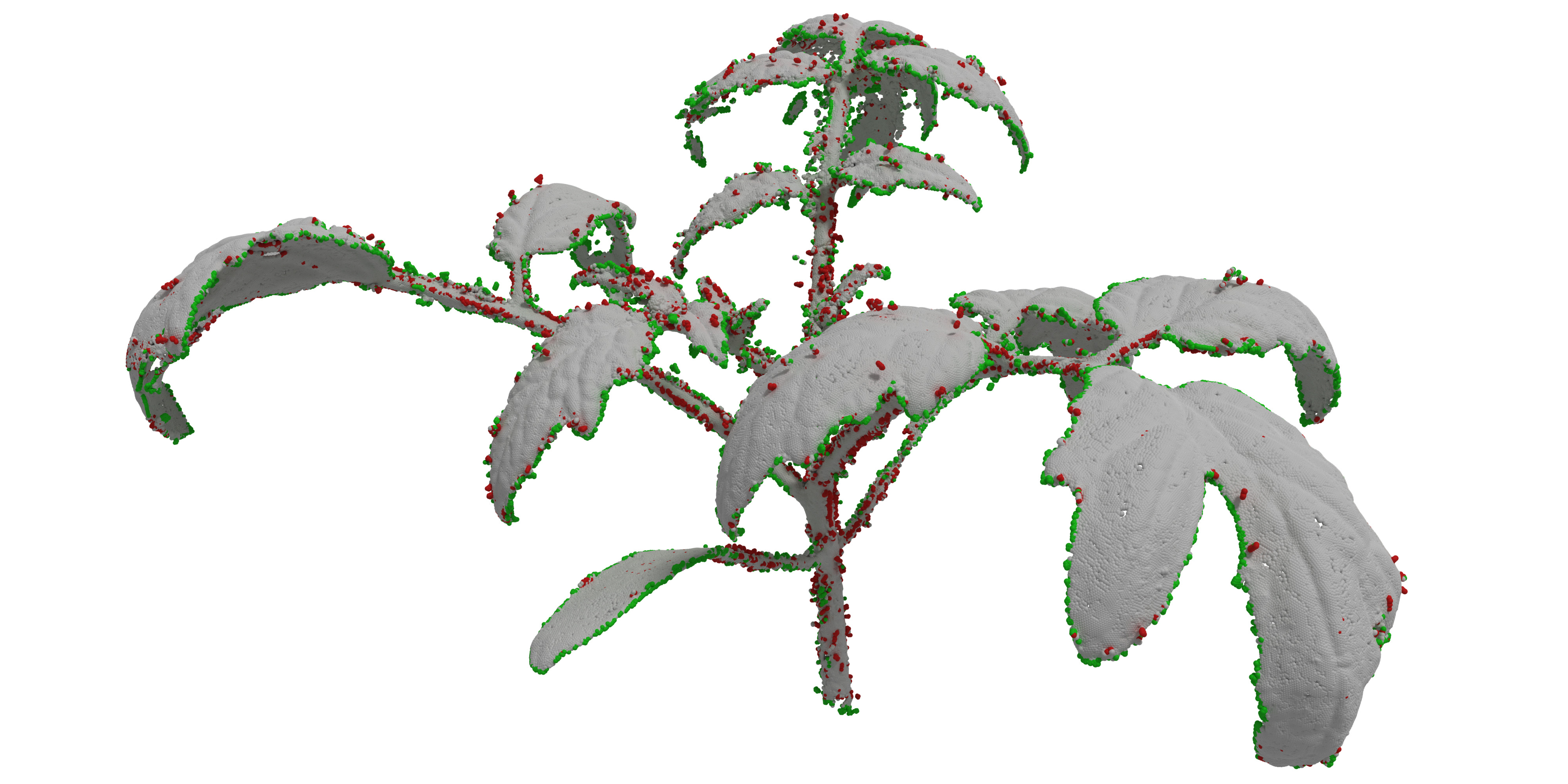}};
        \begin{scope}[shift=(p1.south west),
                      x=(p1.south east),
                      y=(p1.north west)]
            \node[marker1, draw=blue] at ({5.802/16.000}, {3.571/8.000}) {};
        \end{scope}

        \node[mainimg, anchor=south west] (p2) at (p1.north west) {\includegraphics[width=0.3\linewidth]{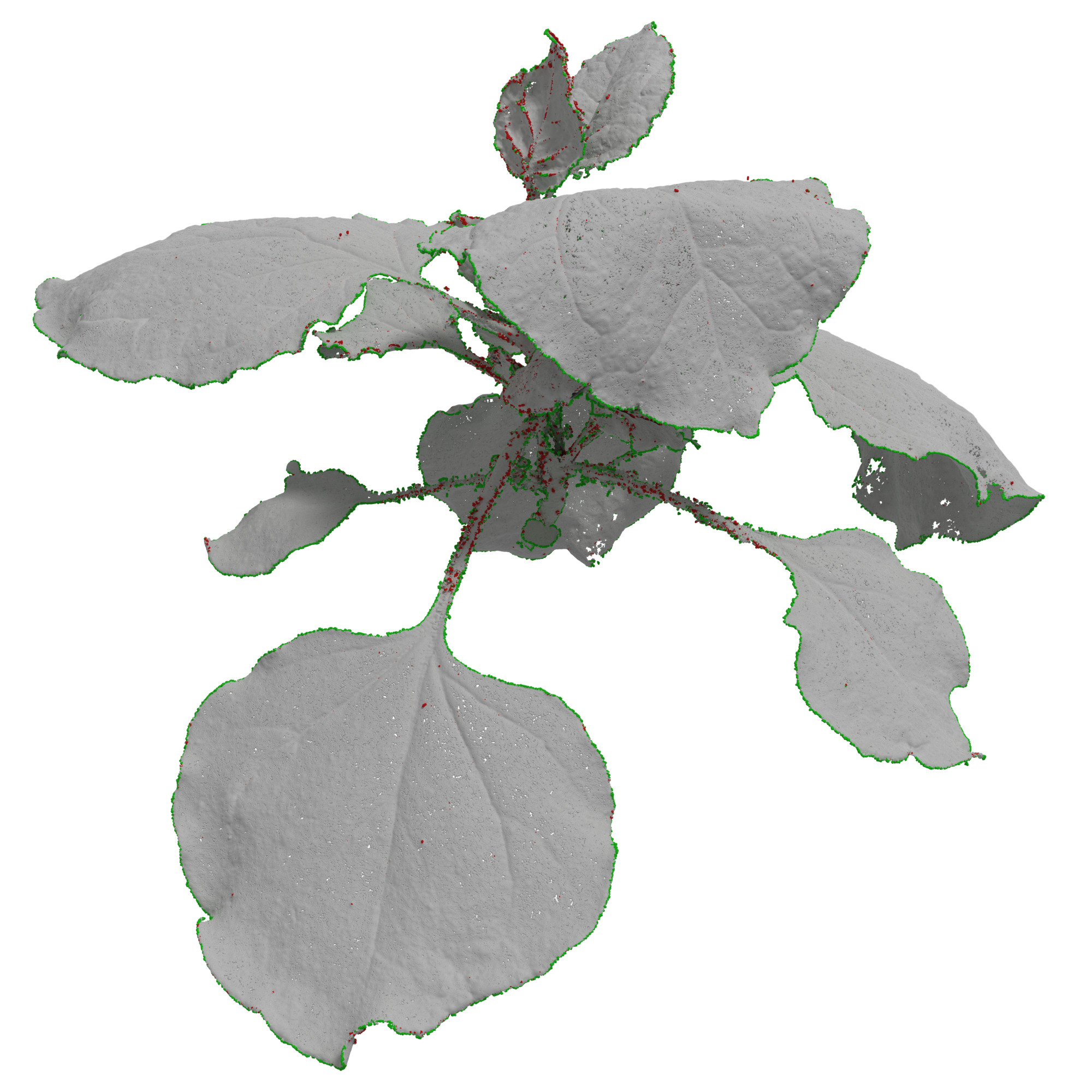}};
        \begin{scope}[shift=(p2.south west),
                      x=(p2.south east),
                      y=(p2.north west)]
            \node[marker2, draw=red] at ({3.073/8.000}, {5.864/8.000}) {};
        \end{scope}

        \node[mainimg, anchor=south east] (p3) at (p1.north east) {\includegraphics[width=0.3\linewidth]{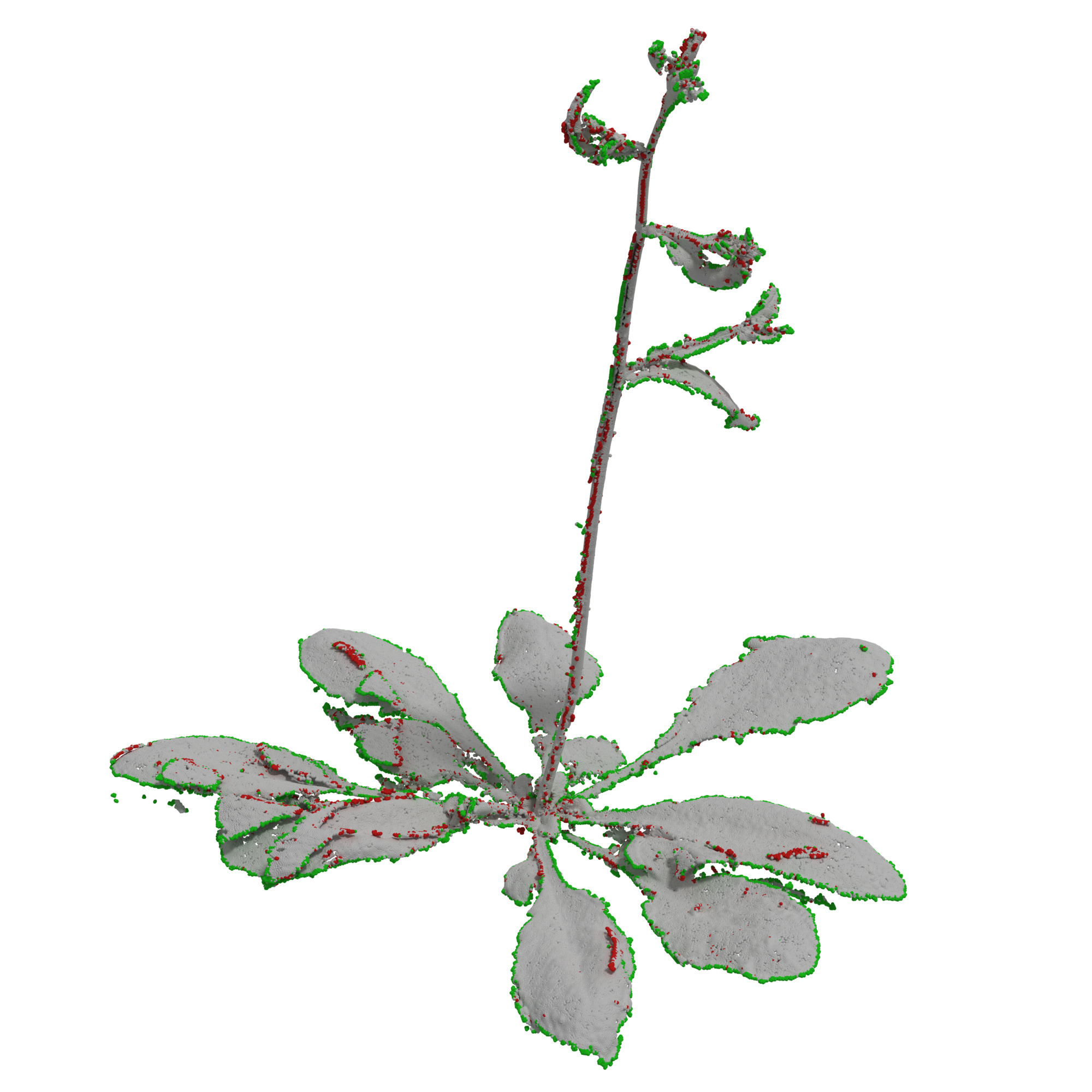}};
        \begin{scope}[shift=(p3.south west),
                      x=(p3.south east),
                      y=(p3.north west)]
            \node[marker2, draw=green] at ({4.277/8.000}, {1.131/8.000}) {};
        \end{scope}

        \node[zoom, anchor=west] (zoom2) at ($(p3.north east)!0.5!(p1.south east)$) {\includegraphics[width=0.18\linewidth]{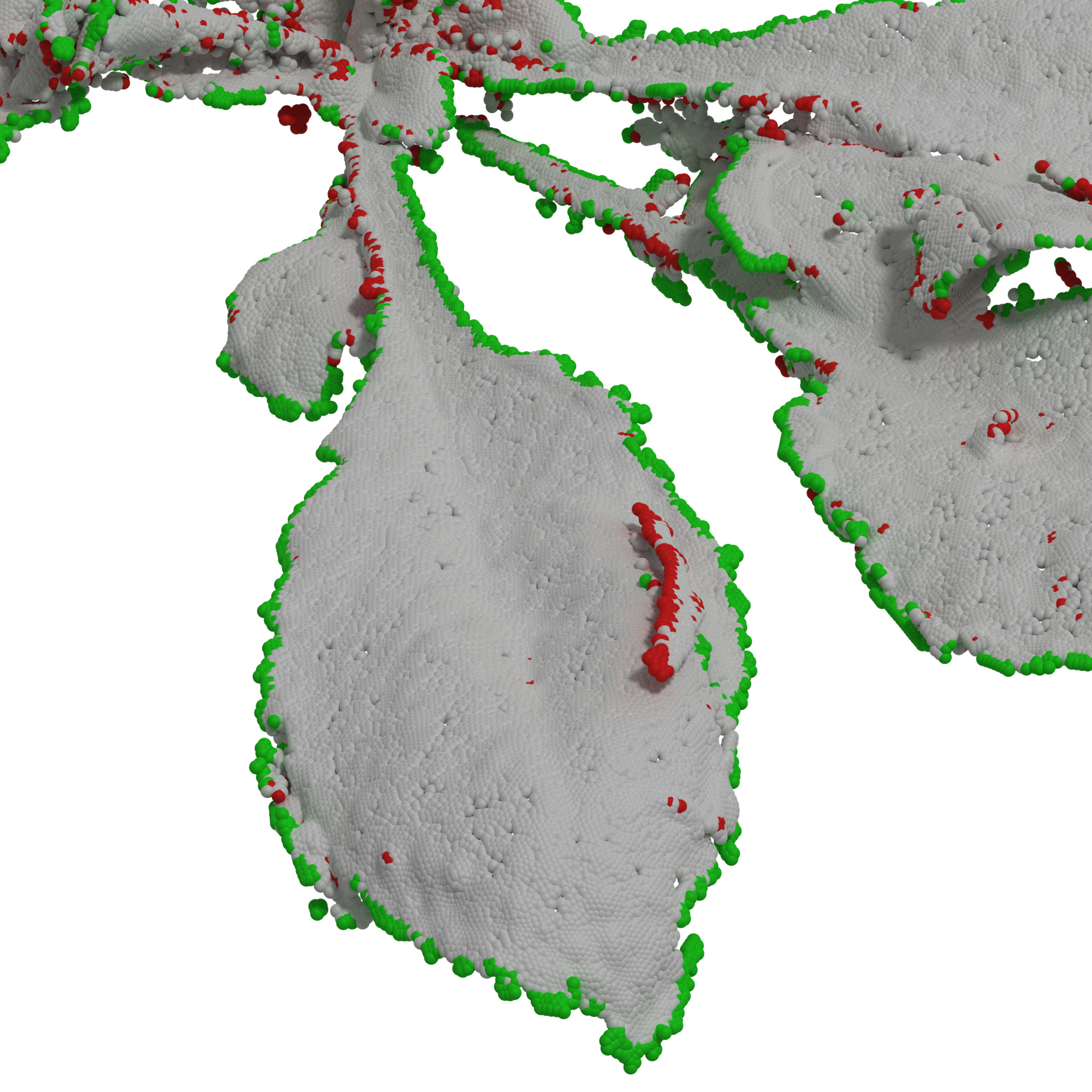}};
        \node[frame, draw=green, fit=(zoom2)] {};
        \node[zoom, above=of zoom2] (zoom1) {\includegraphics[width=0.18\linewidth]{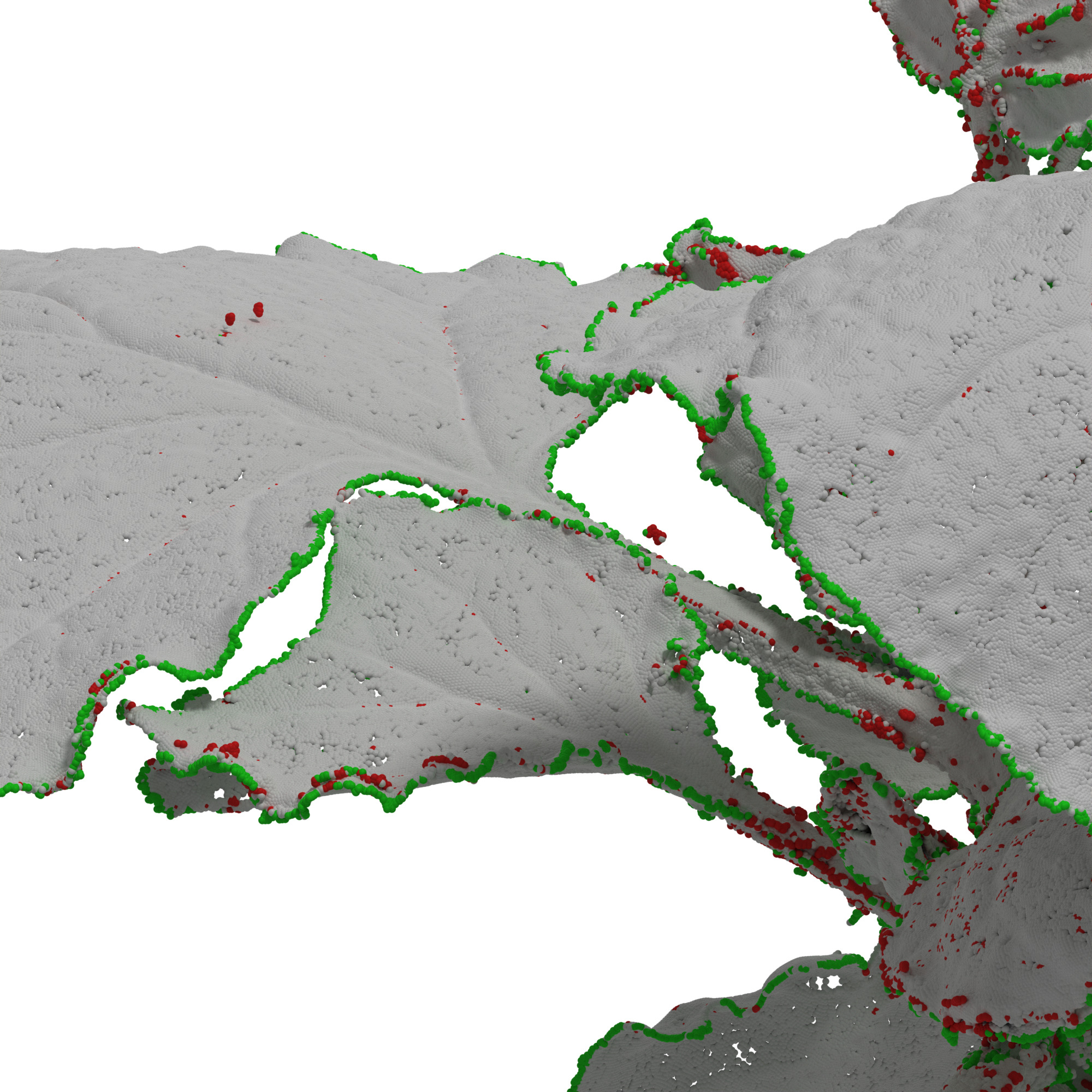}};
        \node[frame, draw=red, fit=(zoom1)] {};
        \node[zoom, below=of zoom2] (zoom3) {\includegraphics[width=0.18\linewidth]{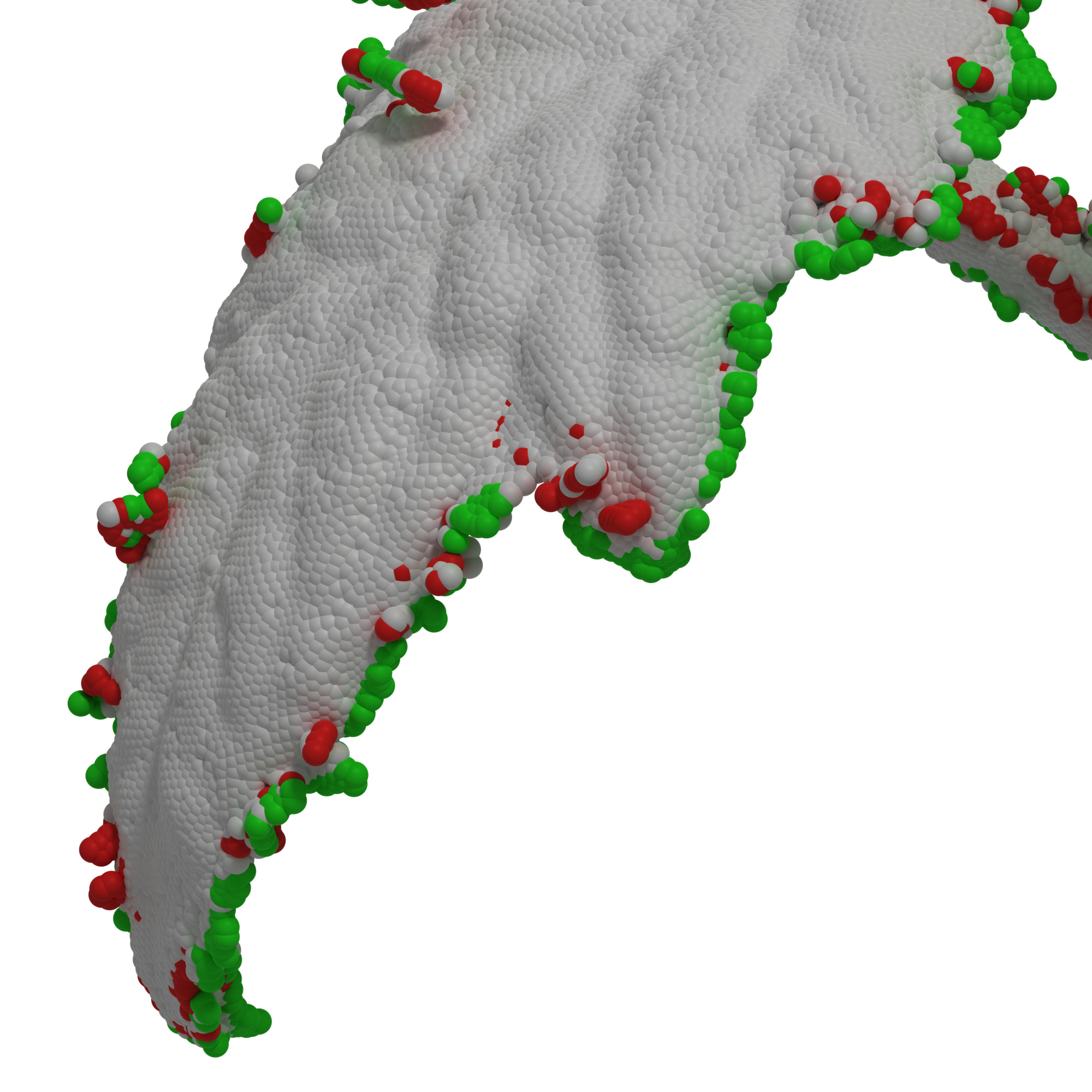}};
        \node[frame, draw=blue, fit=(zoom3)] {};

    \end{tikzpicture}
    \caption{\label{fig:qual_eval_plants}
        3D scanned plant point clouds classified using our approach BoundED trained on \emph{Default++}.
    }
\end{figure*}

\subsection{Qualitative Comparison}\label{sec:qual_eval}

If trained on \emph{Default++}, our algorithm learns to identify the sharp edges in the evaluation models well as can be seen in Figure~\ref{fig:qual_eval_default}.
The edge detection results seem to be even a bit more consistent than the ones of PCEDNet trained on \emph{Default}.
However, PCEDNet does a better job at classifying outliers as non-edge.
We suspect, that the GLS features are more robust regarding outliers and our network was not able to compensate for this as outliers are strongly underrepresented in the training data.
Results on some evaluation models of the \emph{ABC} dataset are depicted in Figure~\ref{fig:qual_eval_abc}.
PCEDNet exhibits mixed performance on the models 0027 and 0059.
Depending on the dataset used for training, the algorithm either tends to have problems with the thin wall of model 0059 or produces less consistent results on some parts of model 0027.
The most consistent results, however, are produced by our approach trained on the \emph{Default++} dataset.
It is the only configuration that produces an inner circular edge on model 0027 without holes while not massively overclassifying the cylindrical wall at the top of the model as sharp edge.
The classification results of points which are part of the screw thread in model 0117 are not as consistent as the detected sharp edges contain many holes.
The screw is classified best by our algorithm trained on \emph{ABC}, but this combination erroneously classifies the boundary of model 1222 as sharp edges.
BoundED also works well on actual 3D scanned real-world data as shown in Figures~\ref{fig:qual_eval_christ_church} and~\ref{fig:qual_eval_station}.
On the \emph{christ\_church} point cloud, it outperforms PCEDNet in classifying the sharp-edges of roofs (see green zoom-in) and also gives good results for the fine stone structures of the church (see blue zoom-in).
The results on the \emph{station} point cloud are similar.
Especially for the third row, our algorithm gives much more consistent results in the area of the escalator.

\begin{figure*}[!t]
    \centering
    \begin{tikzpicture}[image/.style = {inner sep=0mm, outer sep=0pt},
                        label_top/.style = {anchor=south, align=center},
                        label_side/.style = {rotate=90, anchor=south, align=center},
                        node distance = 2mm and 1mm]
        \node[image]               (i11) {\includegraphics[trim={75px 0 75px 0},clip,height=2.2cm]{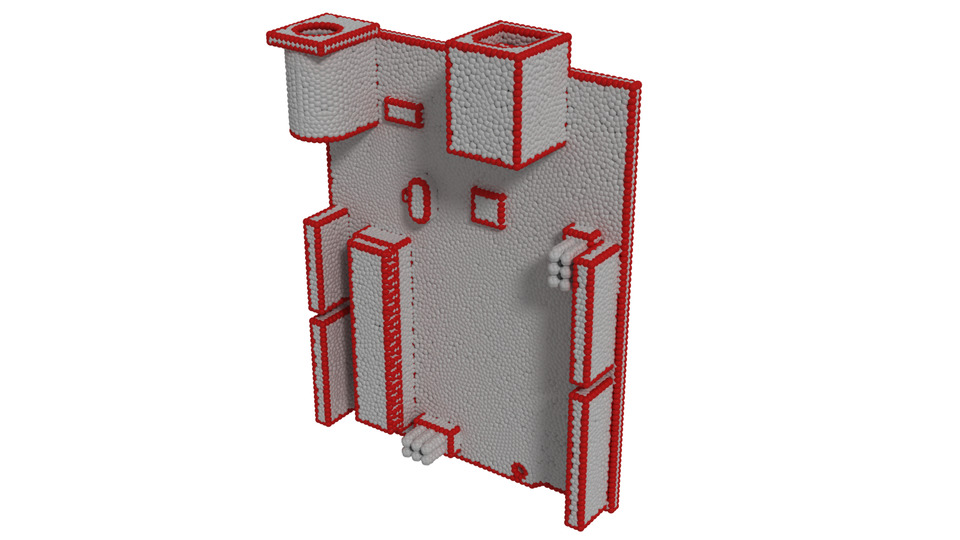}};
        \node[image, right=of i11] (i12) {\includegraphics[trim={75px 0 75px 0},clip,height=2.2cm]{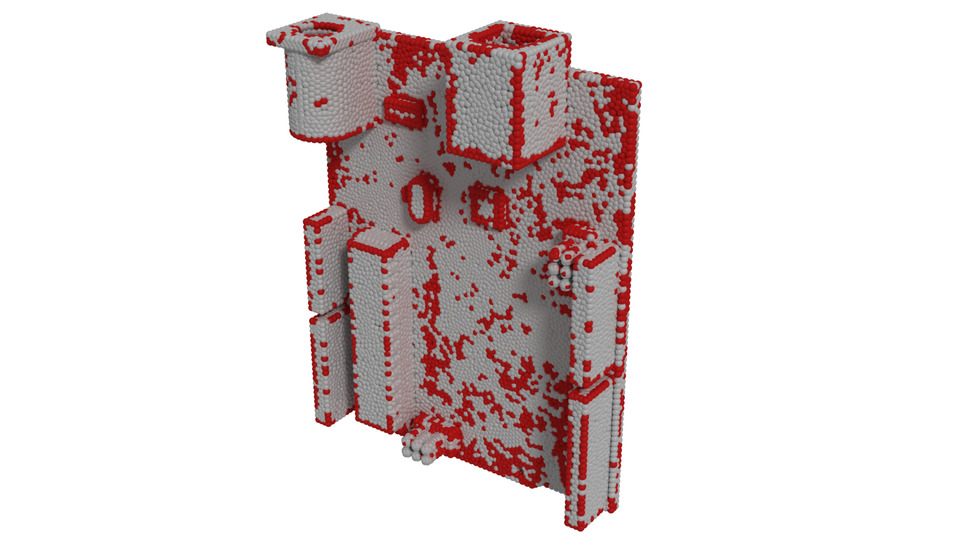}};
        \node[image, right=of i12] (i13) {\includegraphics[trim={75px 0 75px 0},clip,height=2.2cm]{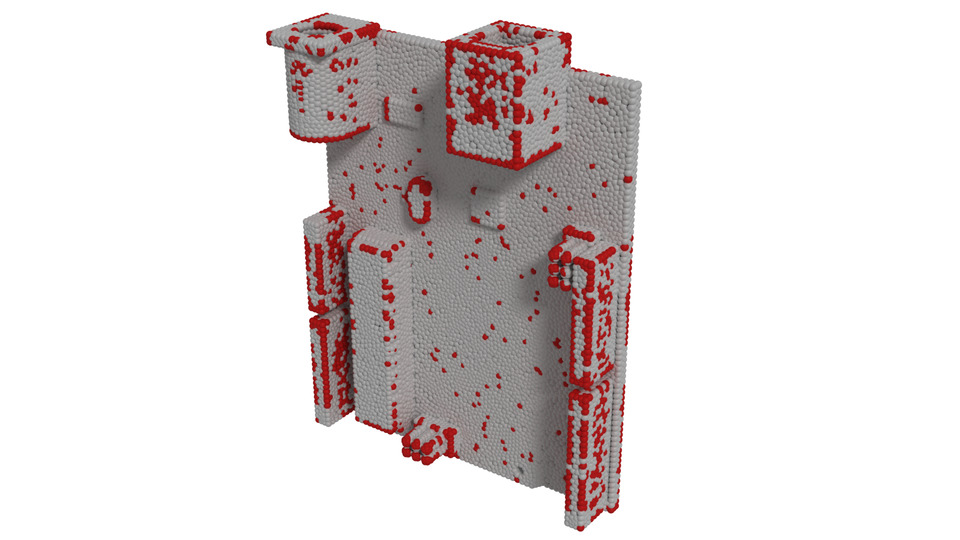}};
        \node[image, right=of i13] (i14) {\includegraphics[trim={75px 0 75px 0},clip,height=2.2cm]{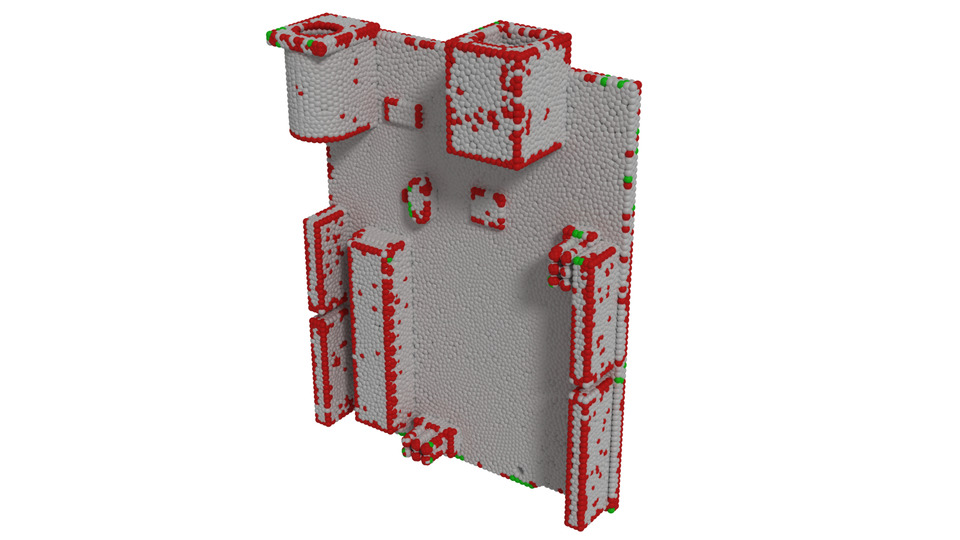}};
        \node[image, below=of i11] (i21) {\includegraphics[trim={75px 0 75px 0},clip,height=2.2cm]{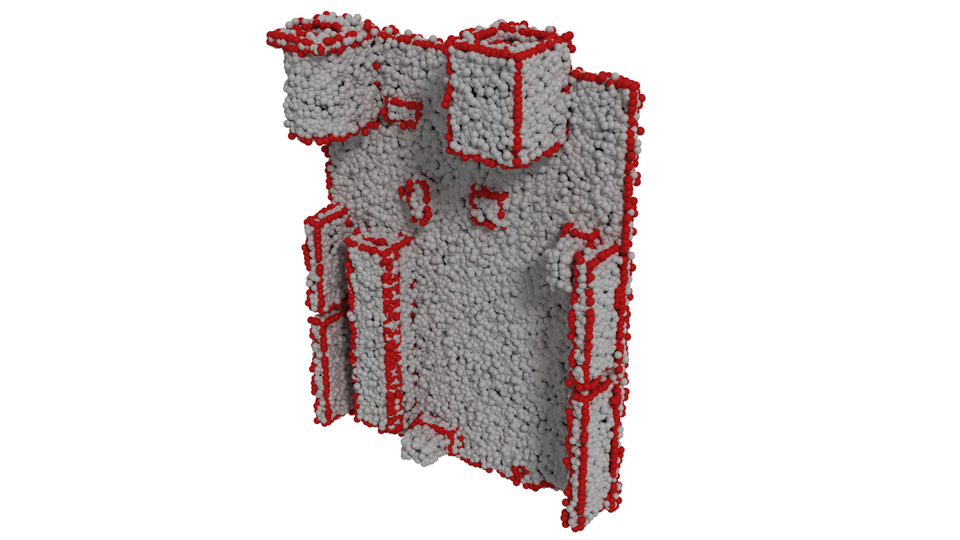}};
        \node[image, right=of i21] (i22) {\includegraphics[trim={75px 0 75px 0},clip,height=2.2cm]{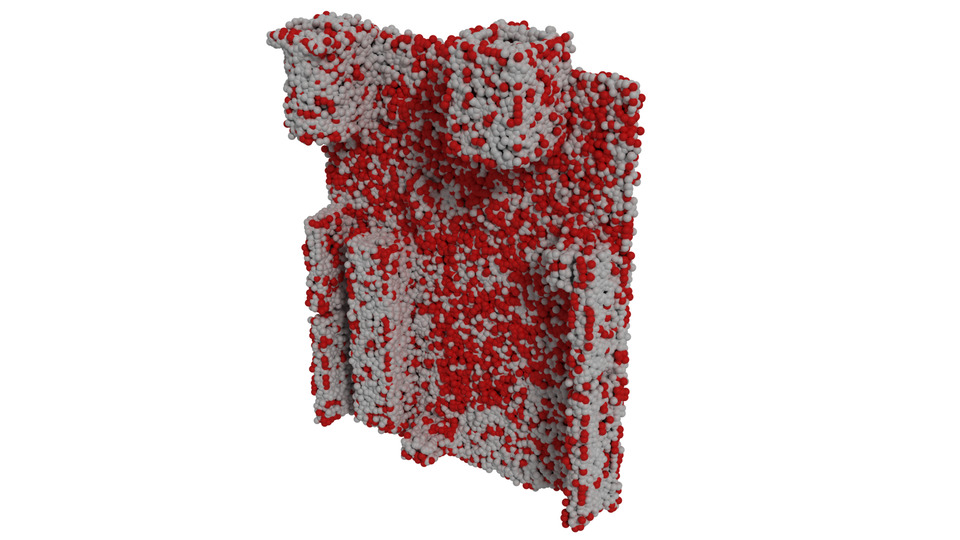}};
        \node[image, right=of i22] (i23) {\includegraphics[trim={75px 0 75px 0},clip,height=2.2cm]{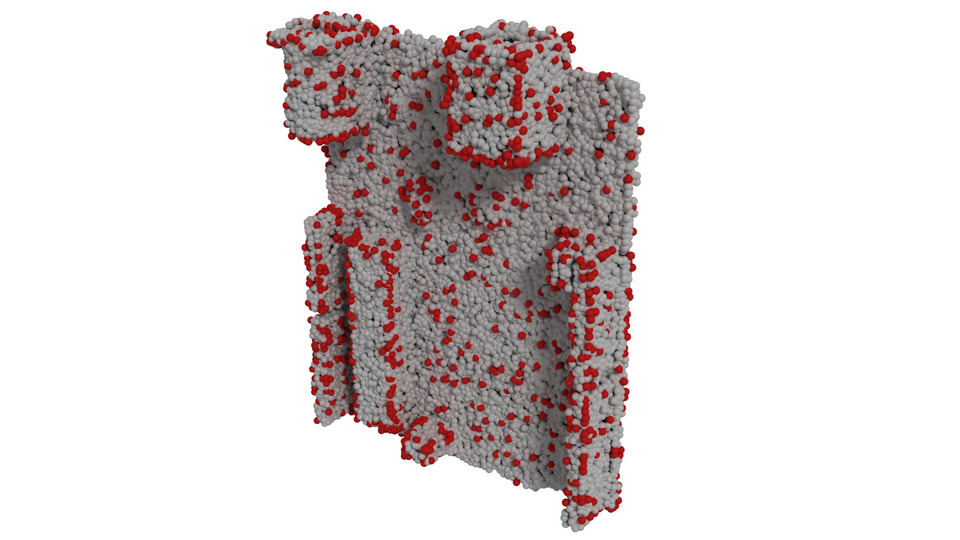}};
        \node[image, right=of i23] (i24) {\includegraphics[trim={75px 0 75px 0},clip,height=2.2cm]{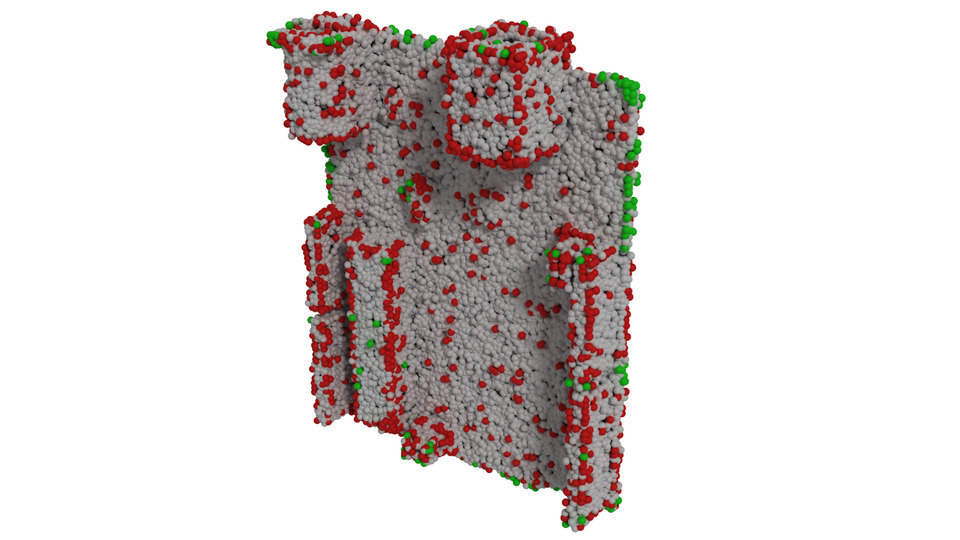}};
        \node[image, below=of i21] (i31) {\includegraphics[trim={75px 0 75px 0},clip,height=2.2cm]{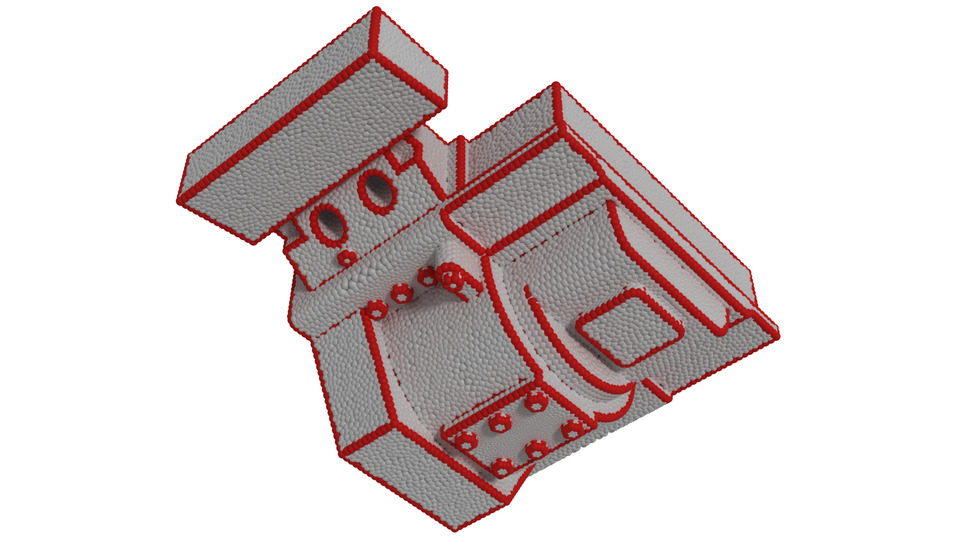}};
        \node[image, right=of i31] (i32) {\includegraphics[trim={75px 0 75px 0},clip,height=2.2cm]{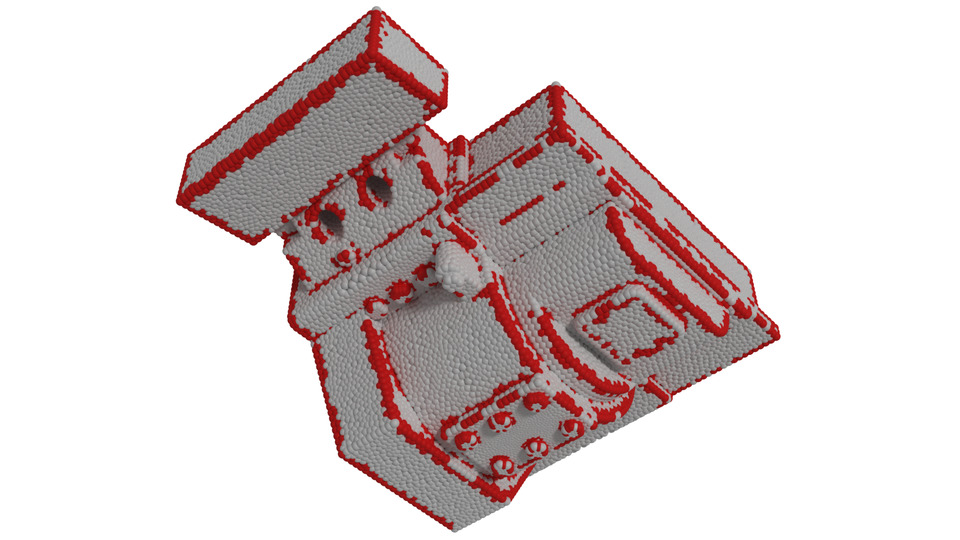}};
        \node[image, right=of i32] (i33) {\includegraphics[trim={75px 0 75px 0},clip,height=2.2cm]{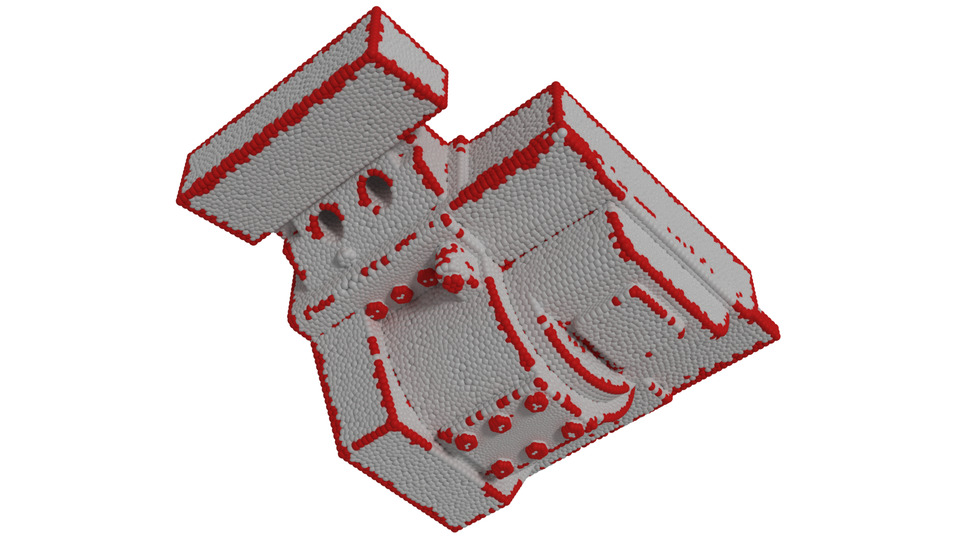}};
        \node[image, right=of i33] (i34) {\includegraphics[trim={75px 0 75px 0},clip,height=2.2cm]{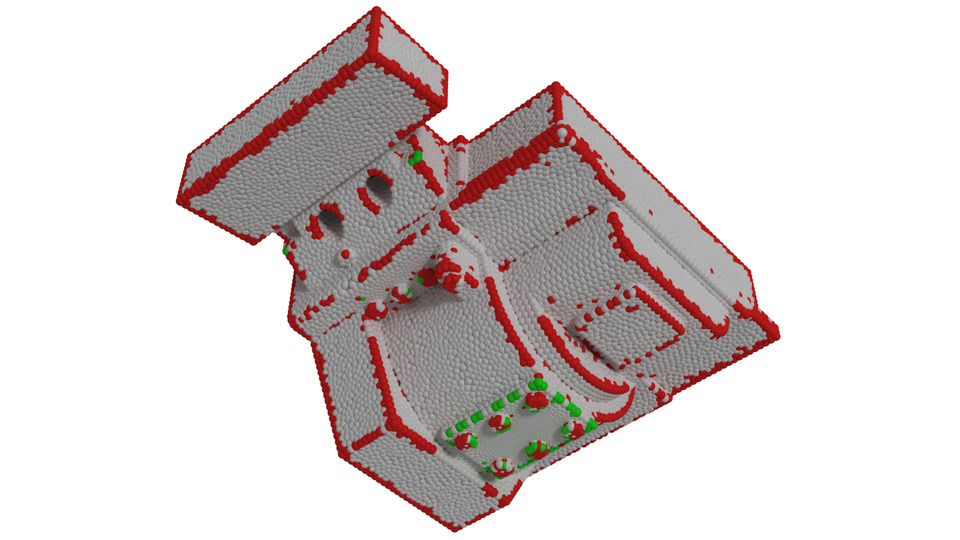}};
        \node[image, below=of i31] (i41) {\includegraphics[trim={75px 0 75px 0},clip,height=2.2cm]{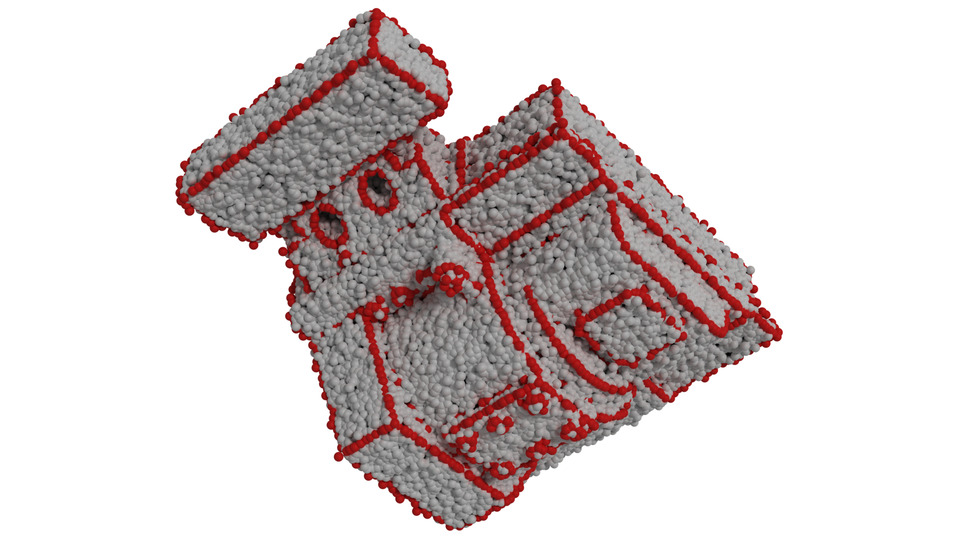}};
        \node[image, right=of i41] (i42) {\includegraphics[trim={75px 0 75px 0},clip,height=2.2cm]{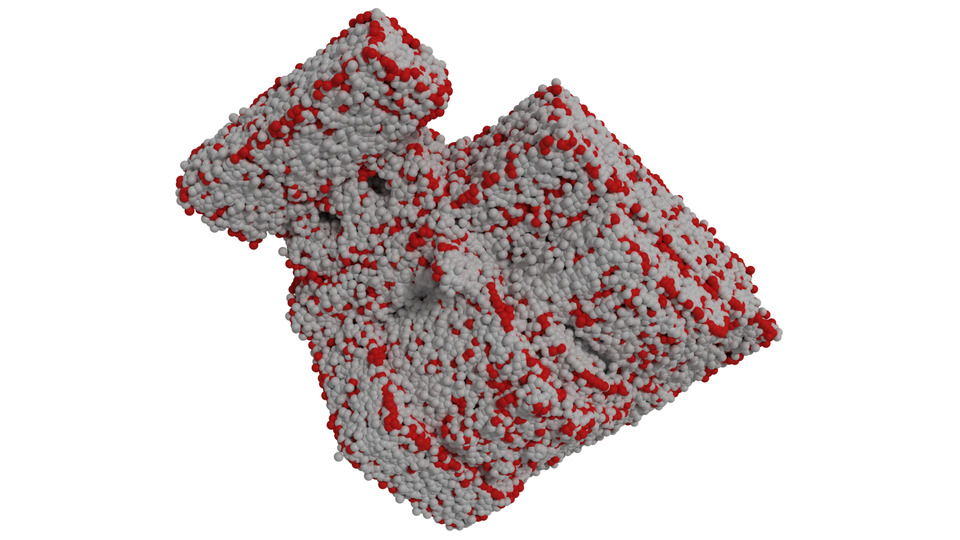}};
        \node[image, right=of i42] (i43) {\includegraphics[trim={75px 0 75px 0},clip,height=2.2cm]{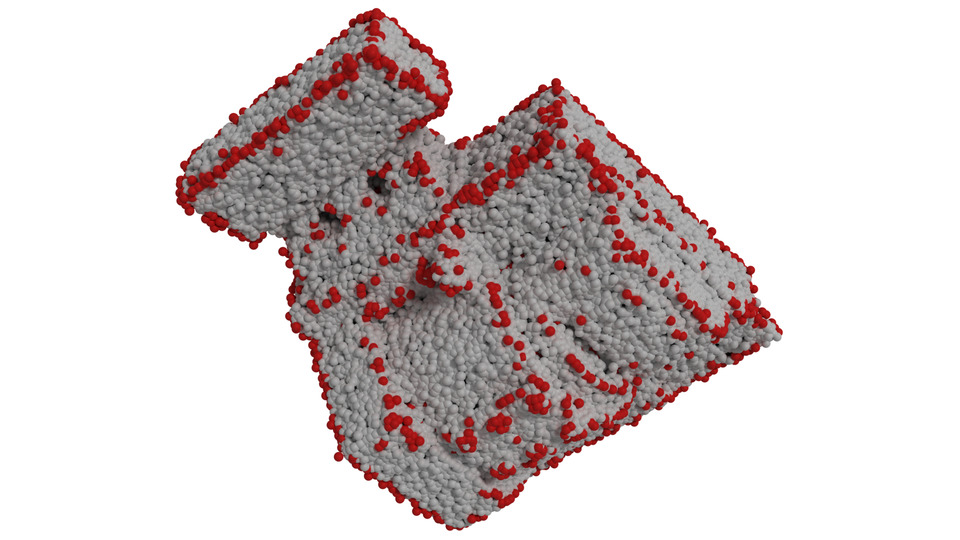}};
        \node[image, right=of i43] (i44) {\includegraphics[trim={75px 0 75px 0},clip,height=2.2cm]{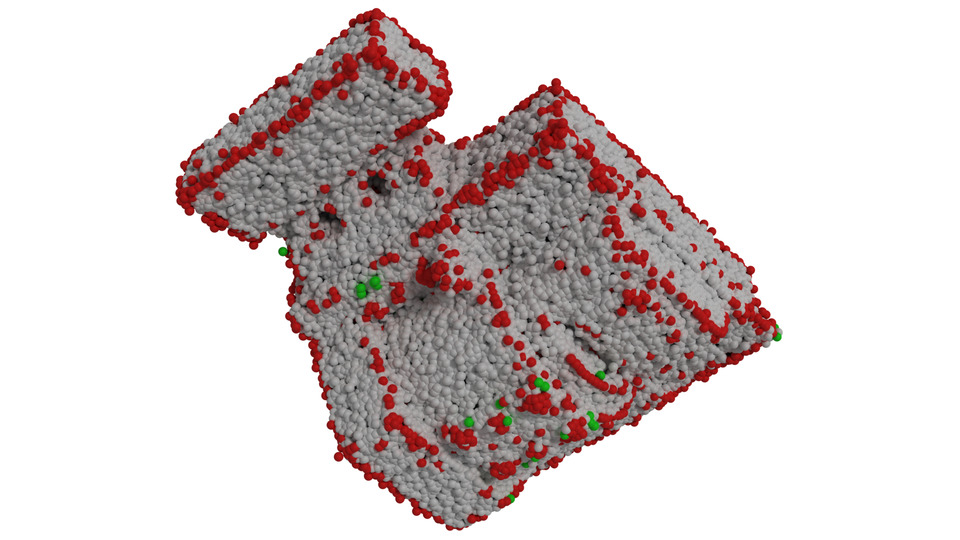}};
        \node[image, below=of i41] (i51) {\includegraphics[trim={75px 0 75px 0},clip,height=2.2cm]{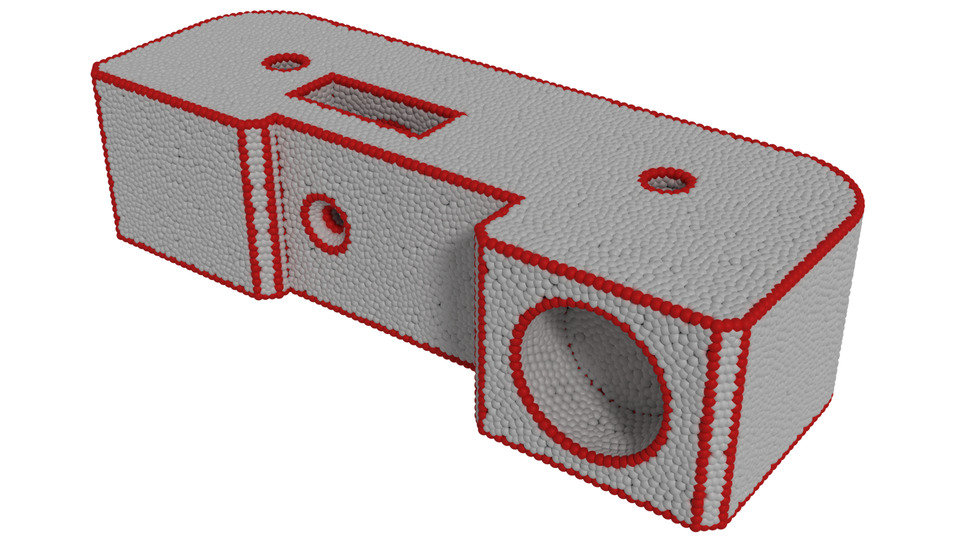}};
        \node[image, right=of i51] (i52) {\includegraphics[trim={75px 0 75px 0},clip,height=2.2cm]{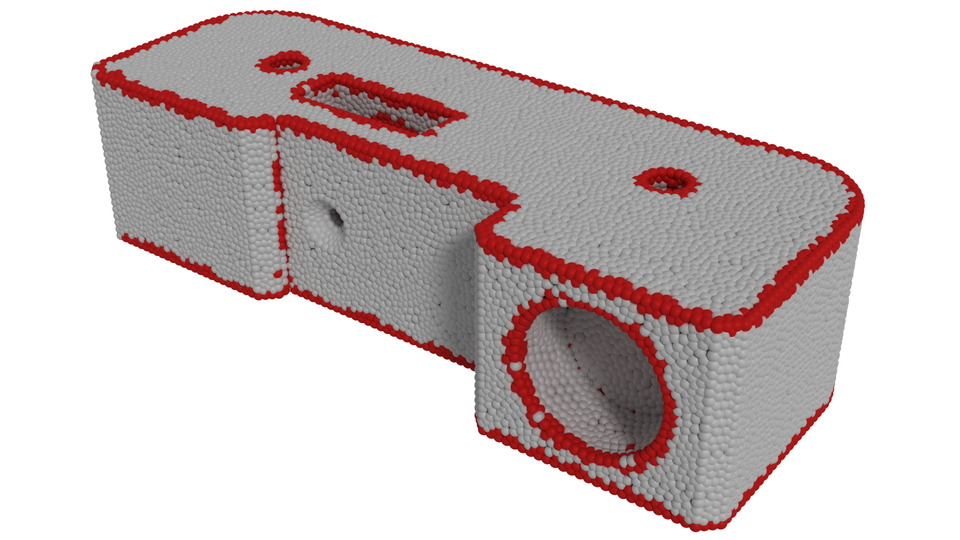}};
        \node[image, right=of i52] (i53) {\includegraphics[trim={75px 0 75px 0},clip,height=2.2cm]{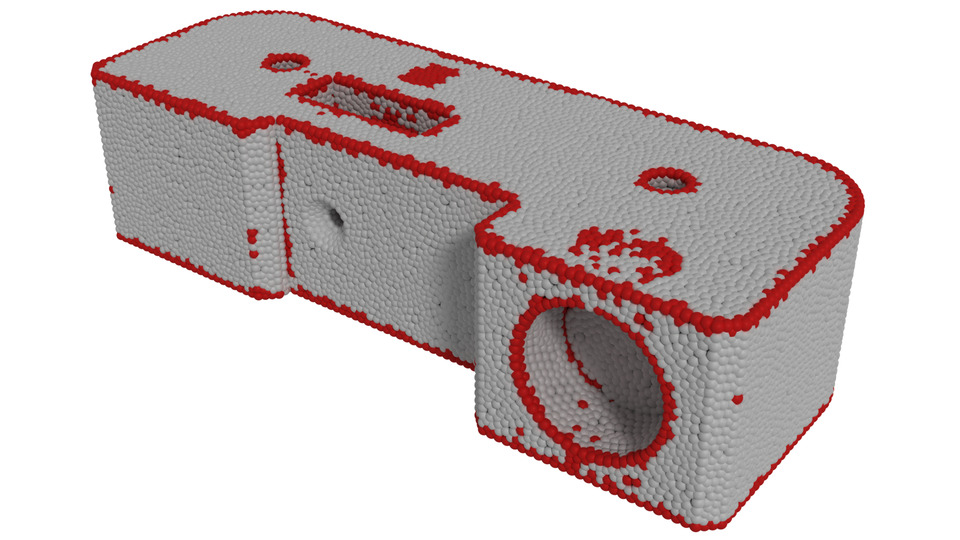}};
        \node[image, right=of i53] (i54) {\includegraphics[trim={75px 0 75px 0},clip,height=2.2cm]{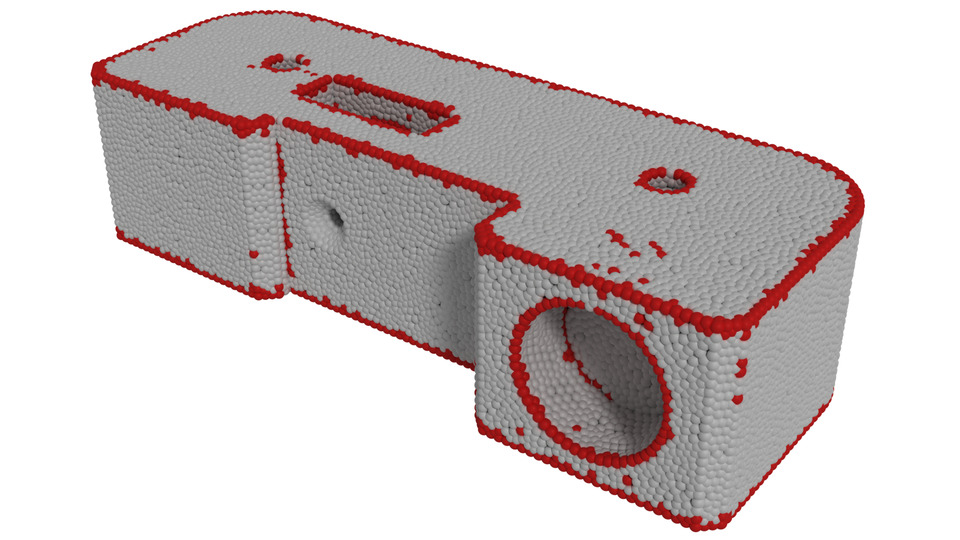}};
        \node[image, below=of i51] (i61) {\includegraphics[trim={75px 0 75px 0},clip,height=2.2cm]{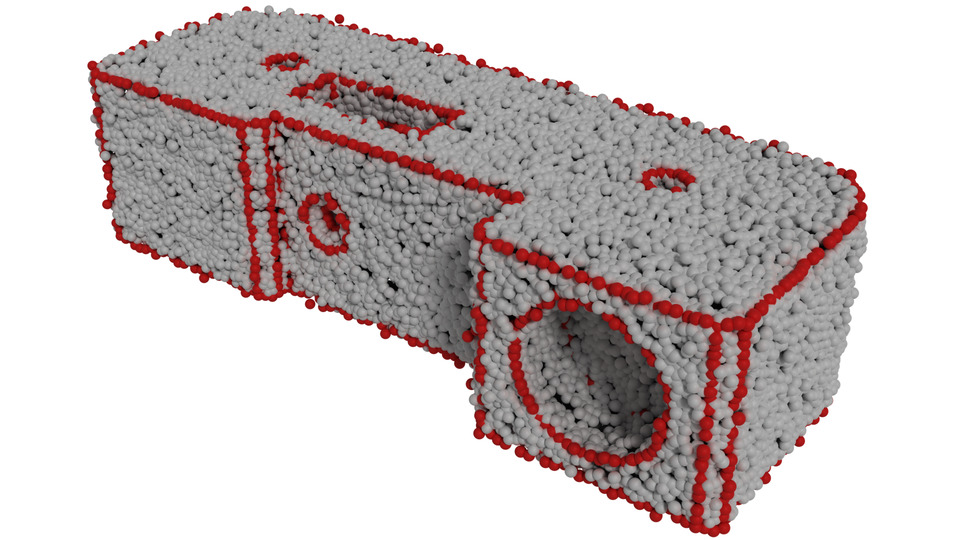}};
        \node[image, right=of i61] (i62) {\includegraphics[trim={75px 0 75px 0},clip,height=2.2cm]{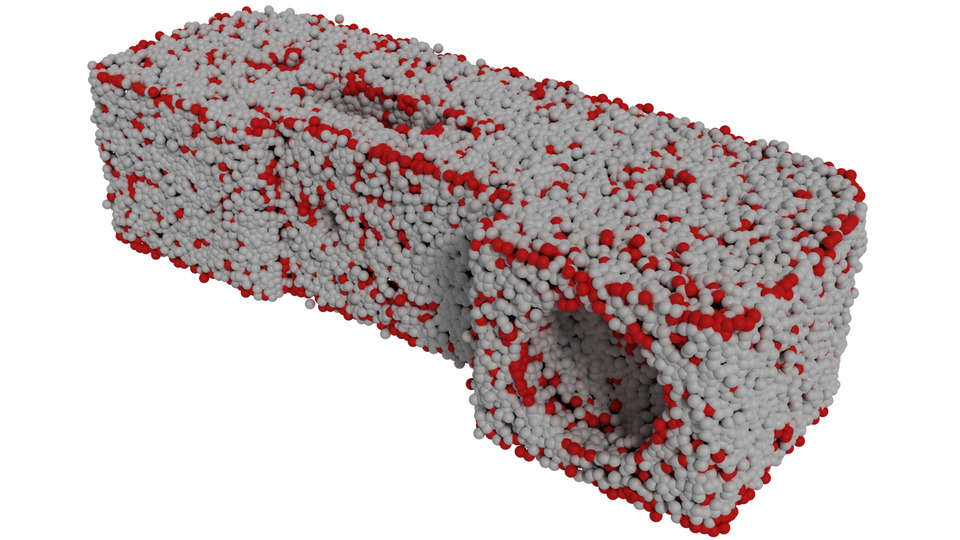}};
        \node[image, right=of i62] (i63) {\includegraphics[trim={75px 0 75px 0},clip,height=2.2cm]{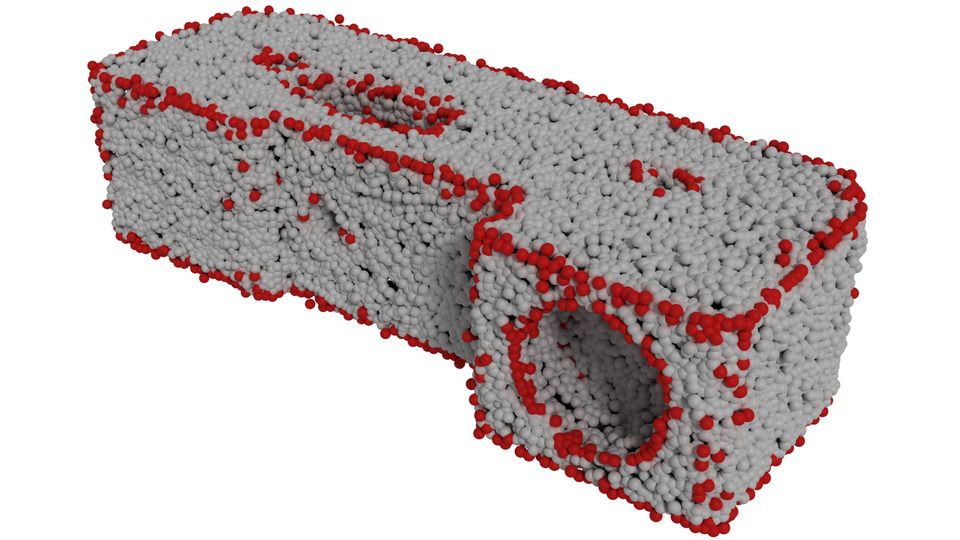}};
        \node[image, right=of i63] (i64) {\includegraphics[trim={75px 0 75px 0},clip,height=2.2cm]{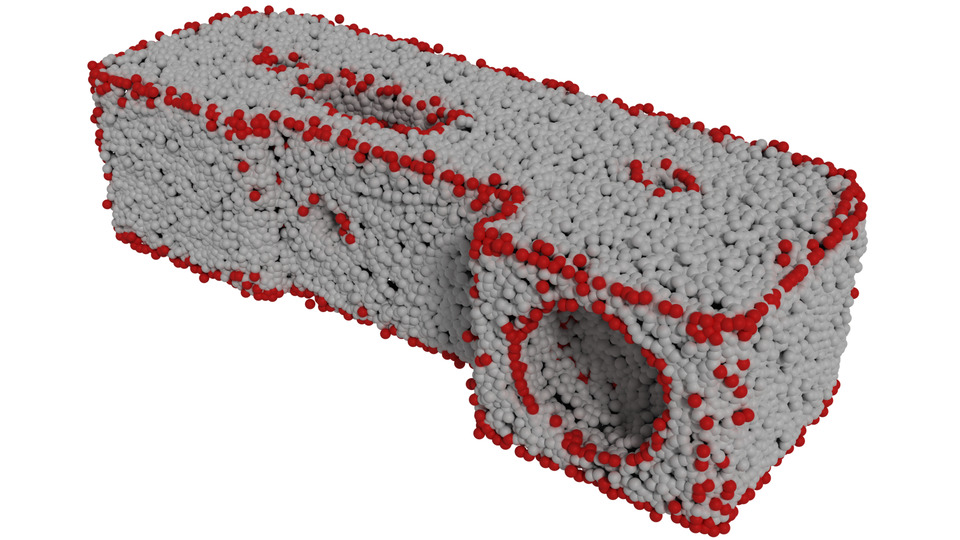}};
        \node[image, below=of i61] (i71) {\includegraphics[trim={75px 0 75px 0},clip,height=2.2cm]{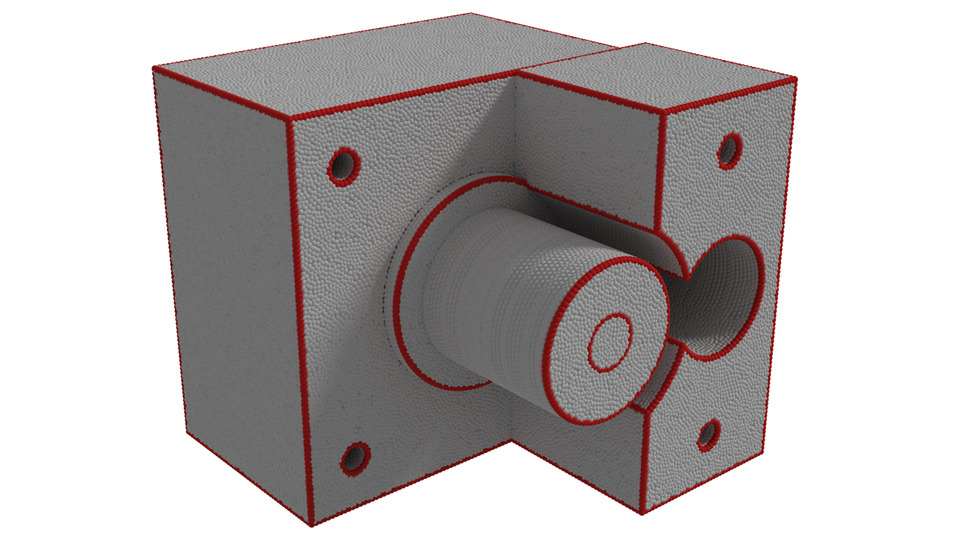}};
        \node[image, right=of i71] (i72) {\includegraphics[trim={75px 0 75px 0},clip,height=2.2cm]{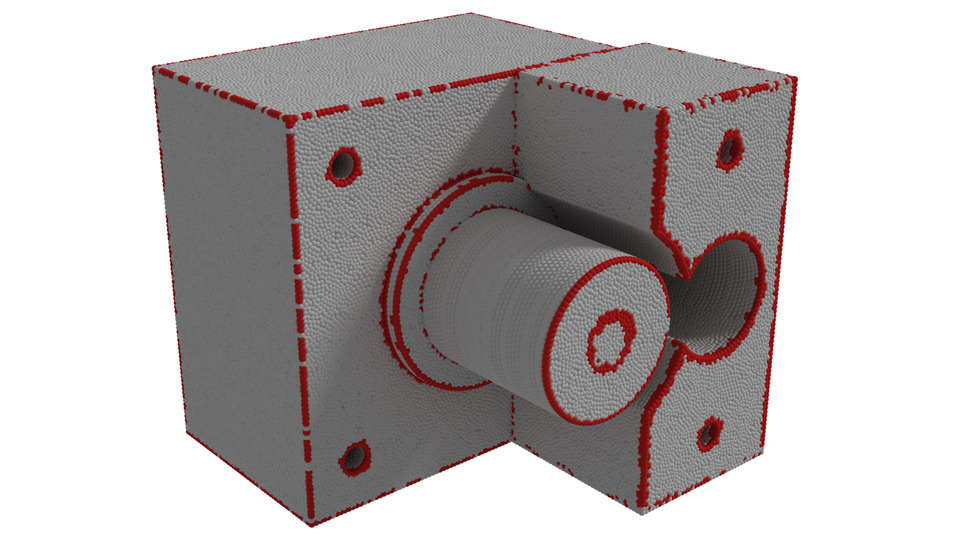}};
        \node[image, right=of i72] (i73) {\includegraphics[trim={75px 0 75px 0},clip,height=2.2cm]{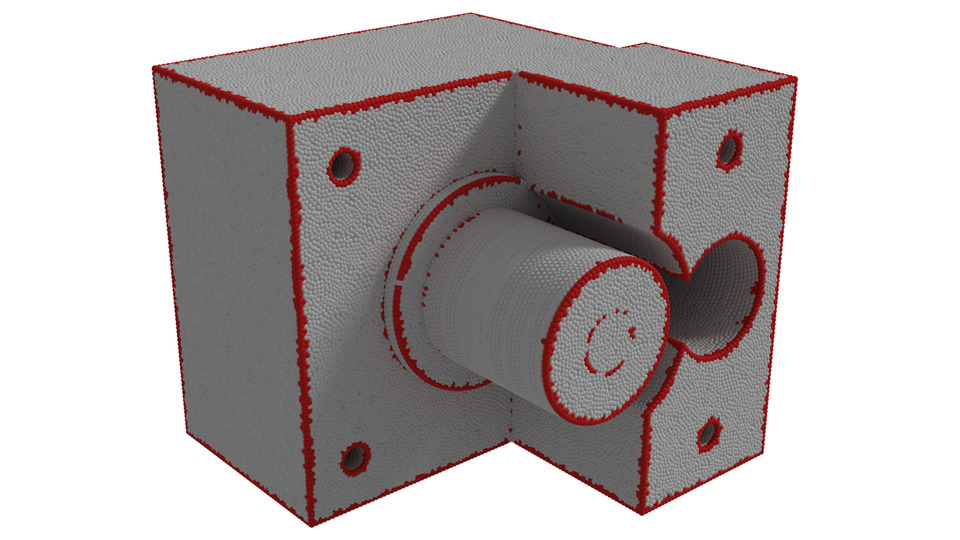}};
        \node[image, right=of i73] (i74) {\includegraphics[trim={75px 0 75px 0},clip,height=2.2cm]{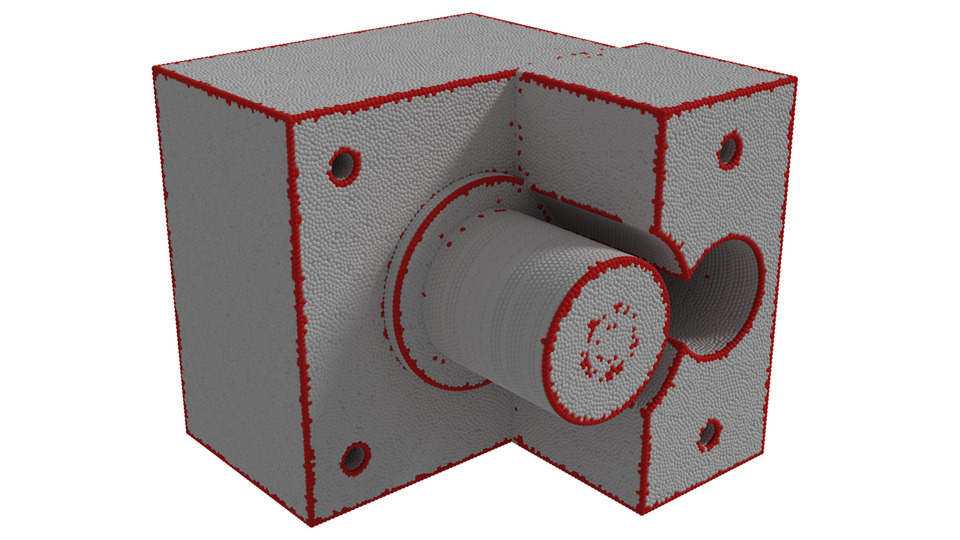}};
        \node[image, below=of i71] (i81) {\includegraphics[trim={75px 0 75px 0},clip,height=2.2cm]{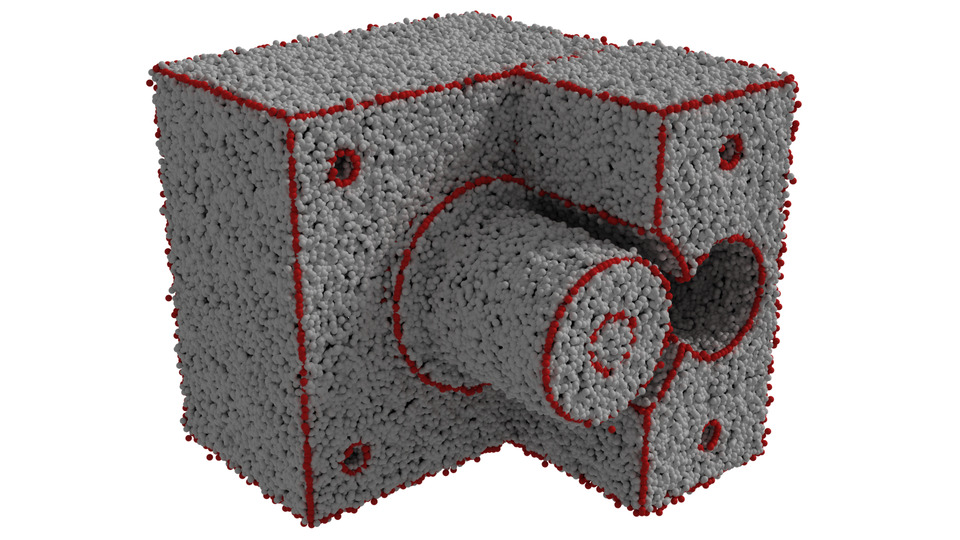}};
        \node[image, right=of i81] (i82) {\includegraphics[trim={75px 0 75px 0},clip,height=2.2cm]{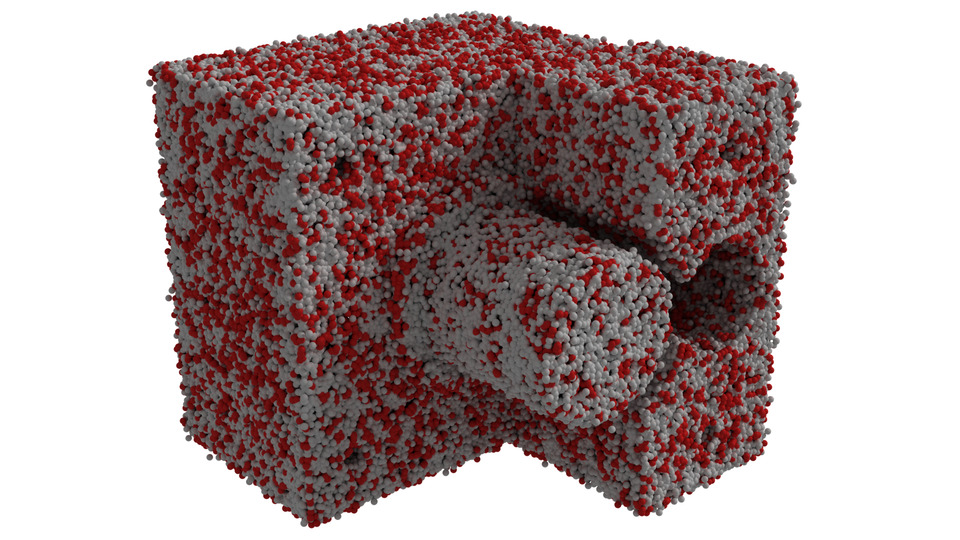}};
        \node[image, right=of i82] (i83) {\includegraphics[trim={75px 0 75px 0},clip,height=2.2cm]{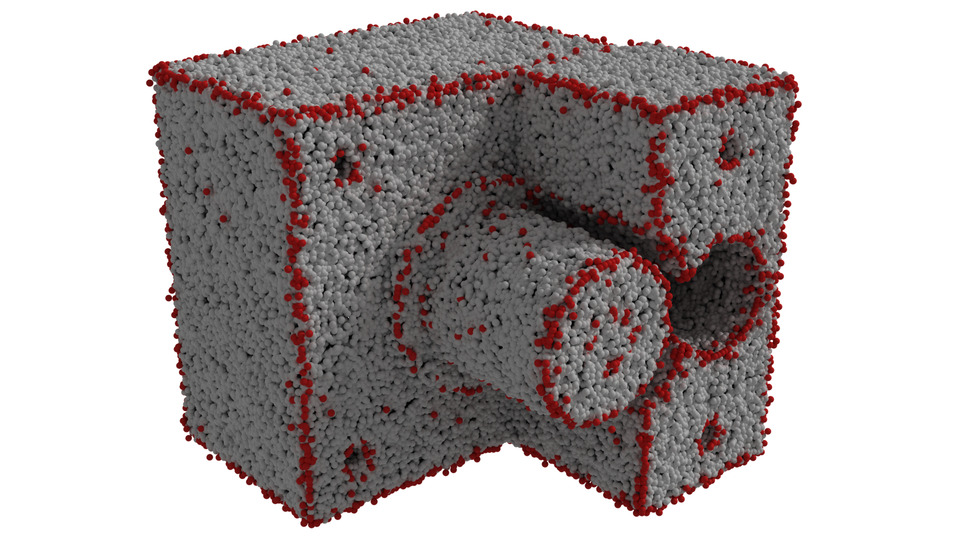}};
        \node[image, right=of i83] (i84) {\includegraphics[trim={75px 0 75px 0},clip,height=2.2cm]{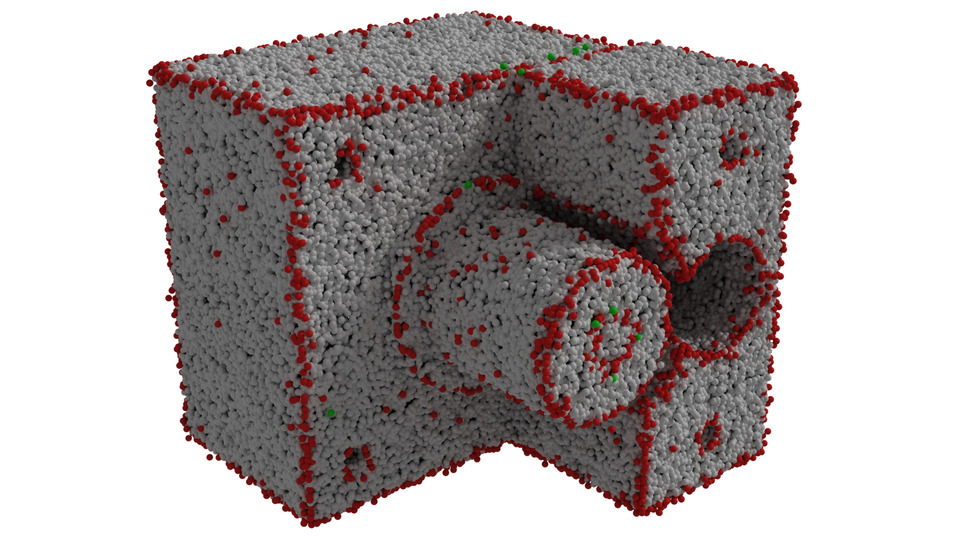}};

        \node[label_top, above=of i11] (lt1) {Ground\\Truth};
        \node[label_top, above=of i12] (lt2) {PCEDNet\\(\emph{Default})};
        \node[label_top, above=of i13] (lt3) {BoundED (Ours)\\(\emph{Default})};
        \node[label_top, above=of i14] (lt4) {BoundED (Ours)\\(\emph{Default++})};
        \node[left=of i11, label_side] (ls1) {Clean};
        \node[left=of i21, label_side] (ls2) {Noisy};
        \node[label_side] (ls12) at ($(ls1.north)!0.5!(ls2.north)$) {2207};
        \node[left=of i31, label_side] (ls3) {Clean};
        \node[left=of i41, label_side] (ls4) {Noisy};
        \node[label_side] (ls34) at ($(ls3.north)!0.5!(ls4.north)$) {3395};
        \node[left=of i51, label_side] (ls5) {Clean};
        \node[left=of i61, label_side] (ls6) {Noisy};
        \node[label_side] (ls56) at ($(ls5.north)!0.5!(ls6.north)$) {4986};
        \node[left=of i71, label_side] (ls7) {Clean};
        \node[left=of i81, label_side] (ls8) {Noisy};
        \node[label_side] (ls78) at ($(ls7.north)!0.5!(ls8.north)$) {7487};

    \end{tikzpicture}
    \caption{\label{fig:noise}
        Comparison of PCEDNet and BoundED regarding bahavior on noisy data.
        The dataset used for training the respective approach is given in parentheses.
    }
\end{figure*}

\begin{table*}[t]
    \scriptsize
    \begin{center}
        \begin{tabular}{lcccccc}\toprule
                                                                                                                                     & Precision($\uparrow$) & Recall($\uparrow$) & MCC($\uparrow$) & F1($\uparrow$) & Accuracy($\uparrow$) & IoU($\uparrow$) \\ \midrule
            $\bm{x}_{i,k} = (\bm{d}_{i,k}, \bm{s}_{i,k}, \bm{c}_{i,k})^T$                                                                                           & 0.308      & 0.680      & 0.413      & 0.427      & 0.888      & 0.272      \\
            $\bm{x}_{i,k} = (\bm{\sigma}_{i,k})^T$                                                                                                                  & 0.331      & 0.379      & 0.310      & 0.354      & 0.915      & 0.215      \\
            $\bm{x}_{i,k} = (\bm{\sigma}_{i,k,\mathrm{upper}}, \bm{\sigma}_{i,k,\mathrm{lower}})^T$                                                                 & 0.380      & 0.113      & 0.172      & 0.172      & \bf{0.934} & 0.094      \\
            $\bm{x}_{i,k} = (\bm{\sigma}_{i,k,\mathrm{upper}}, \bm{\sigma}_{i,k,\mathrm{lower}}, \bm{s}_{i,k})^T$                                                   & 0.361      & 0.743      & 0.464      & 0.474      & 0.903      & 0.311      \\
            $\bm{x}_{i,k} = (\bm{\sigma}_{i,k,\mathrm{upper}}, \bm{\sigma}_{i,k,\mathrm{lower}}, \bm{d}_{i,k}, \bm{s}_{i,k})^T$                                     & 0.423      & 0.711      & 0.499      & 0.518      & 0.923      & 0.349      \\
            $\bm{x}_{i,k} = (\bm{\sigma}_{i,k,\mathrm{upper}}, \bm{\sigma}_{i,k,\mathrm{lower}}, \bm{d}_{i,k}, \bm{c}_{i,k})^T$                                     & 0.416      & 0.248      & 0.296      & 0.326      & 0.933      & 0.195      \\
            $\bm{x}_{i,k} = (\bm{\sigma}_{i,k,\mathrm{upper}}, \bm{\sigma}_{i,k,\mathrm{lower}}, \bm{s}_{i,k}, \bm{c}_{i,k})^T$                                     & 0.327      & 0.760      & 0.449      & 0.454      & 0.889      & 0.293      \\
            $\bm{x}_{i,k} = (\bm{\sigma}_{i,k}, \bm{d}_{i,k}, \bm{s}_{i,k}, \bm{c}_{i,k})^T$                                                                        & 0.359      & 0.723      & 0.472      & 0.482      & 0.903      & 0.318      \\
            $\bm{x}_{i,k} = (\bm{\sigma}_{i,k,\mathrm{upper}}, \bm{\sigma}_{i,k,\mathrm{lower}}, \bm{d}_{i,k}, \bm{s}_{i,k}, \bm{c}_{i,k})^T$                       & \bf{0.436} & 0.709      & 0.522      & \bf{0.542} & 0.927      & \bf{0.371} \\
            $\bm{x}_{i,k} = (\bm{\sigma}_{i,k}, \bm{\sigma}_{i,k,\mathrm{upper}}, \bm{\sigma}_{i,k,\mathrm{lower}}, \bm{d}_{i,k}, \bm{s}_{i,k}, \bm{c}_{i,k})^T$    & 0.373      & \bf{0.866} & \bf{0.529} & 0.521      & 0.903      & 0.352      \\
            \bottomrule
        \end{tabular}
        \caption{\label{tab:abl_features}
            Ablation study regarding the choice of features used as input for the network.
            The table lists median scores for various classification metrics.
            \emph{Default++} dataset is used for training as well as evaluation.
            The individual features are defined in Sections~\ref{sec:features} and~\ref{sec:network}.
        }
    \end{center}
\end{table*}

\begin{table*}[t]
    \scriptsize
    \begin{center}
        \begin{tabular}{lcccccc}\toprule
                                                                                                                                     & Precision($\uparrow$) & Recall($\uparrow$) & MCC($\uparrow$) & F1($\uparrow$) & Accuracy($\uparrow$) & IoU($\uparrow$) \\ \midrule
             2 scales (128, 32)                                                                                                      & 0.313      & \bf{0.857} & 0.470      & 0.456      & 0.876      & 0.295      \\
             4 scales (128, 64, 32, 16)                                                                                              & 0.436      & 0.709      & \bf{0.522} & 0.542      & 0.927      & 0.371      \\
             8 scales (128, 91, 64, 45, 32, 23, 16, 11)                                                                              & \bf{0.477} & 0.639      & 0.515      & 0.542      & \bf{0.935} & 0.372      \\
            16 scales (128, 108, 91, 76, 64, 54, 45, 38, 32, 27, 23, 19, 16, 13, 11, 10)                                             & 0.463      & 0.662      & 0.520      & \bf{0.545} & 0.932      & \bf{0.375} \\
            \bottomrule
        \end{tabular}
        \caption{\label{tab:abl_scales}
            Ablation study regarding the number of scales used by our network.
            The table lists median scores for various classification metrics.
            The \emph{Default++} dataset is used for training as well as evaluation.
        }
    \end{center}
\end{table*}

\subsection{Boundary Detection}

As already mentioned in Section~\ref{sec:introduction}, the processing of point clouds often requires the detection of boundaries in addition to sharp edges due to potentially very fine structures as well as finite resolution.
This is especially important if the scanned object has many fine structures like leafs on plants or fine fins on buildings.
Due to the GLS~\citep{mellado2012growing} features used in PCEDNet~\citep{himeur2021pcednet}, which rely on point normals estimated using a small neighborhood of points, PCEDNet is by design not able to detect boundaries in point clouds.
In contrast, using our proposed set of features and the extended \emph{Default++} dataset makes our approach capable of detecting boundaries in addition to sharp edges.
Figure~\ref{fig:qual_eval_default} shows successfully detected boundaries for the two rightmost models, i.e. the only ones containing actual boundary points.
For model 1222 of the \emph{ABC} datasets evaluation data (see Figure~\ref{fig:qual_eval_abc}), the boundary is found almost perfectly as well.
Despite being actually 3D structures and therefore not boundaries in the strict sense, the top of the walls of model 0059 are detected as a boundary as well.
Due to the low thickness of the walls, this is a reasonable behavior depending on the exact use-case for the extracted boundary data.
Very thin structures being identified correctly as boundary can also be seen in the red zoom-in of Figure~\ref{fig:qual_eval_christ_church}.
In the \emph{station} point cloud (see Figure~\ref{fig:qual_eval_station}), mostly points of thin signs and humans are identified as boundary points.
Note, that humans in this point cloud are mostly two-dimensional due to the scanning procedure and rather low resolution.
Finally, results on plants are depicted in Figure~\ref{fig:qual_eval_plants}.
All leaves are nicely separated by boundaries.
Some stems contain sharp-edge points due to scanning artifacts.

\subsection{Behavior on Noisy Data}\label{sec:noise}

In addition to the results on clean point clouds in Figure~\ref{fig:qual_eval_abc}, Figure~\ref{fig:noise} shows a direct comparison on clean as well as noisy data taken from the \emph{ABC} dataset of our algorithm BoundED and PCEDNet~\citep{himeur2021pcednet}.
The respective noisy models are taken from~\citet{himeur2021pcednet}\footnote{Available at: \url{https://storm-irit.github.io/pcednet-supp/abc_noise_0.04.html}, accessed on 10/19/2022.}.
Note, that we assume only the point positions to be given.
Thus, the normals needed by PCEDNet were calculated according to the authors instructions via meshlab~\citep{meshlab}.
BoundED outperforms PCEDNet on noisy data if both are trained on \emph{Default} as it is significantly less prone to predict false positives in originally flat regions.
The difference is particularly noticeable in the eighth row on model 7487.
Furthermore, in the sixth row on model 4986, PCEDNet has difficulties in detecting the prominent sharp edges at the top and botom of the object.
As the network architectures of both approaches are very similar, we expect that the main reason for our approach to perform better in the presence of noise is the additional robustness of our features due to the underlying statistics.

\subsection{Ablation Study}

In the scope of an additional ablation study, we validate the chosen features as well as the selected number of scales.
Table~\ref{tab:abl_features} shows median scores of various classification metrics for results of our approach trained and evaluated on the \emph{Default++} dataset.
While all chosen features seem to contribute positively to the classification result, experiments suggest, that $\bm{s}_{i,k}$ is the most important feature.
We suspect the reason for this to be the high importance of its tangential component for the detection of boundaries while additionally the normal component can be utilized by the network to classify sharp-edge points.
Also note, that the partitioning of the neighborhood according to the estimated normal degrades classification quality if the singular values of the respective covariance matrices are given to the network in isolation.
However, if being combined with the other proposed features, the partioning improves the results.
We reason, that the partitioning provides the network with additional cues for finetuning the results, but is too dependent on correct normal estimation to be suited for robust sharp-edge and boundary point detection without further information.
Passing the singular values $\bm{\sigma}_{i,k}$ of the unpartitioned neighborhood's covariance matrix to the network in addition to $\bm{\sigma}_{i,k,\mathrm{upper}}$ and $\bm{\sigma}_{i,k,\mathrm{lower}}$ does not seem to improve the results significantly.
The impact of the number of scales is shown in Table~\ref{tab:abl_scales}.
For the experiments, we chose to use $2^i$ neighbors per scale where $i$ is distributed evenly-spaced over the interval $(3, 7]$.
While the performance of our algorithm using 4, 8, or 16 scales is very similar, using 2 scales performs much worse.
As a trade-off between performance and computational burden, we use four scales.

\subsection{Limitations}

Despite yielding great results in most cases, the feature extraction step can fail in various scenarios.
If e.g. the smallest singular value of a neighborhood's covariance matrix does not correspond to the true surface normal, the points are partitioned in an unexpected way leading to very unpredictable results.
Note, that, by estimating the normal per-scale and passing all respective singular values to the network, it is able to extract additional information about the neighborhood.
Using only a single-scale noraml e.g. estimated during the scanning process might therefore lead to less accurate classifications.
To some degree, this can be compensated by the classification network given enough training data.
Nonetheless, the results would surely improve if the feature extraction step can already tackle such edge cases on its own by e.g. using a per-scale global normal smoothing step.
Furthermore, the \emph{Default} dataset was designed to yield good results if GLS features are used for classification, but it does not cover all relevant edge cases for our features.
Outliers seem to be one such example.
As they are heavily underrepresented in the training data, the classification network fails to distinguish those properly from points on the actual surface.
Designing a new point cloud dataset with our features in mind or even generating a dataset based on the feature values directly instead of going the detour over generating point clouds could solve this problem.
Finally, depending on the point cloud size, a large part of the time needed for the feature extraction step is spent on finding the $k$ neighbors of each point.
A custom tailored solution for this neighborhood search could probably improve the performance of the feature extraction significantly.
Due to the simplicity and compactness of the network, the same holds for the implementation of the classification network as general frameworks like PyTorch introduce a significant overhead in this situation.

%% file: tex/conclusion.tex
\section{Conclusion and Future Work}
\label{sec:conclusion}

In this work we introduced a novel set of per-point features to facilitate the detection of sharp edges and boundaries via a simple and compact neural classification network.
Due to the small network and an efficient GPU implementation for the feature extraction, the algorithm is faster than previous state-of-the-art methods while at the same time achieving more consistent classification results.
This could make the proposed BoundED algorithm a good choice for situations in which interactive classification is required.
The two-level covariance analysis conducted on the neighborhood of a point has, even in the simple form deployed in this work, proven to be a valuable tool to describe the local geometry. 
In the future, our novel features could be utilized to estimate the curvature of curved surfaces as well.
We expect the inclusion of higher order moments to further improve the results and enable us to also learn the estimation of distances to edges and boundaries.

%% file: tex/acknowledgements.tex
\section*{Acknowledgements}

Funding: This work was supported by the DFG-Project KL 1142/9-2.
%